\documentclass[review]{elsarticle}
\usepackage[utf8]{inputenc}

\usepackage{lineno,hyperref}
\modulolinenumbers[5]

\newtheorem{definition}{Definition}
\usepackage{graphicx}
\usepackage{todonotes}
\usepackage{mathtools}
\usepackage{amsfonts}
\usepackage{amssymb}
\usepackage{wrapfig}
\usepackage{times}
\usepackage{url}
\usepackage{makecell}
\usepackage{color, colortbl}
\usepackage{caption}
\usepackage{subcaption}
\usepackage[normalem]{ulem}
\definecolor{LightCyan}{rgb}{0.9,1,1}
\definecolor{LightGray}{rgb}{0.95,0.95,0.95}

\newcommand{\ME}{\mathcal{E}}
\newcommand{\MV}{\mathcal{A}}
\newcommand{\MW}{\mathcal{W}}
\newcommand{\MT}{\mathcal{T}}
\newcommand{\ML}{L}
\newcommand{\BB}{\mathbb{B}}
\newcommand{\R}{\mathbb{R}}
\newcommand{\NN}{\mathbb{N}}

\begin{document}

\begin{frontmatter}
\title{An Explainable Decision Support System for Predictive Process Analytics}

\author[1,2]{Riccardo Galanti\corref{cor1}}
\ead{riccardo.galanti@ibm.com}

\author[2]{Massimiliano de Leoni\corref{cor1}}
\ead{deleoni@math.unipd.it}

\author[2]{Merylin Monaro\corref{cor1}}
\ead{merylin.monaro@unipd.it}

\author[2]{Nicolò Navarin\corref{cor1}}
\ead{nnavarin@math.unipd.it}

\author[1]{Alan Marazzi\corref{cor1}}
\ead{alan.marazzi@ibm.com}

\author[2]{Brigida Di Stasi\corref{cor1}}

\author[2]{Stéphanie Maldera\corref{cor1}}

\address[1]{IBM, Bologna, Italy}
\address[2]{University of Padua, Padua, Italy}

\begin{abstract}
Predictive Process Analytics is becoming an essential aid for organizations, providing online operational support of their processes. 
However, process stakeholders need to be provided with an explanation of the reasons why a given process execution is predicted to behave in a certain way. Otherwise, they will be unlikely to trust the predictive monitoring technology and, hence, adopt it.
This paper proposes a predictive analytics framework that is also equipped with explanation capabilities based on the game theory of Shapley Values. The framework has been implemented in the IBM Process Mining suite and commercialized for business users. The framework has been tested on real-life event data to assess the quality of the predictions and the corresponding evaluations. In particular, a user evaluation has been performed in order to understand if the explanations provided by the system were intelligible to process stakeholders.
\end{abstract}


\end{frontmatter}


\section{Introduction}
\label{sec:intro}

Process predictive analytics aims to monitor the running instances of a given process and to alert on those that risk to not meet the desired outcome, such as taking too long, costing too much, not sufficiently satisfying customers. 

Process predictive analytics techniques are typically based on Machine- or Deep-Learning models that are trained over process' event data (cf.\ Section~\ref{sec:Related_works} and the survey by M{\'{a}}rquez et al.~\cite{Marquez-Chamorro18}).
However, the majority of these techniques are based on black-box models (e.g.\ Neural networks), which are proven to be more accurate, compared to those based on explicit rules (e.g.\ classification/regression trees), which tend to be significantly less accurate.

While the priority remains on giving accurate predictions, users need to be provided with an explanation of the reasons why a given process execution is predicted to behave in a certain way. Black-box models fail to achieve this goal. Previous studies have shown that a necessary condition to build trust is to explain the reason of the provided predictions; otherwise, users would not trust the predictive-monitoring technology, and hence, adopt it~\cite{10.1007/s11257-017-9195-0,doshivelez2017rigorous}. 

Therefore, it has become more and more evident that conventional performance measures of predictive models, such as accuracy or F-score, are insufficient, and model interpretability/explainability needs to be incorporated into this assessment~\cite{DBLP:conf/icsoc/SindhgattaOM20}.
Stierle et al. report on the repertoire of techniques that have been developed to address the problem of interpreting/explaining the process predictive model~\cite{DBLP:conf/ecis/StierleBWZM021}.
However, they claim that \textit{''future works should assess the techniques in real-life settings with end-users''}, and that \textit{''two works presented a case study applying a real-life data set, yet, without deploying the technique in the companies. So far, no work exists with an experimental evaluation method.''}

In line with the claims above, this paper does not only propose an explainable predictive process analytics framework, but also reports on a user study when several real and potential process analysts, from academy and industry, are confronted with the implementation of the framework. In particular, the implementation is not limited to be an academic proof-of-concept prototype, but has been incorporated into a commercial tool, the IBM Process Mining suite, and made available to the customers. The user study using a commercial tool provides further strength to validation of the explanation undestandability by process analysts.

Experiments were conducted on different benchmarks and on several real-life datasets, with the aim of predicting different outcome indicators (Key Performance Indicators) and explaining the predictions. Both gradient boosting and LSTM models were used as predictor models. Results show that the prediction-model quality of the two solutions are comparable with a slight preference for gradient boosting models, with the latter ones that can be trained significantly faster and are thus preferable. The explainable framework returns explanations at global level, aiming to discover the principal factors driving the predictive model, but also at local level of each running process instance. The global-level and local-level explanations were provided to 20 users to fulfill 18 tasks related to process predictive analytics: the user study shows that the explanation framework can be understood by users and provides valuable insights into the factors that affect predictions. 

The paper is organized as follows.
Section~\ref{sec:Preliminary} states the problem addressed in this paper.
Section~\ref{sec:Related_works} summarizes the most relevant work related to process predictive monitoring and Explainable AI. Section~\ref{sec:predictive} illustrates the state of the art on using LSTM and Catboost models for predictive monitoring, on which we build to provide explanations.
Section~\ref{sec:Explainable} reports on our framework for explainable predictive process monitoring.
Section~\ref{sec:experiments} illustrates the case studies conducted on several publicly-available datasets. Section~\ref{sec:user_evaluation} provides an overview of the integration of the explainable predictive process monitoring framework with the IBM process mining suite and reports on the user evaluation conducted with process analysts; finally, Section~\ref{sec:conclusions} concludes the paper.

\section{Preliminary}
\label{sec:Preliminary}

The starting point for a prediction system is an \textit{event log}. 
An event log is a multiset of \emph{traces}. Each trace describes the life-cycle of a particular \emph{process instance} (i.e., a \emph{case}) in terms of the \emph{activities} executed and the process \emph{attributes} that are manipulated.
\begin{definition}[Events]
Let $\MV$ be the set of process attributes. Let $\MW_\MV$ be a function that assigns a domain $\MW_\MV(a)$ to each process attribute $a\in \MV$.
Let be $\overline{\MW} = \cup_{a \in \MV} \MW_\MV(a)$.
An \emph{event} $e \in \ME$ is a partial function $\MV \not\rightarrow \overline{\MW}$ assigning values to process attributes, with $e(a) \in \MW_\MV(a)$.
\end{definition}
A trace is a sequence of events. 
Note that the same event can potentially occur in different traces, namely attributes are given the same assignment in different traces. 
This means that potentially the entire same trace can appear multiple times. This motivates why an event log is to be defined as a multiset of traces:\footnote{Given a set $X$, $\BB(X)$ indicates the set of all multisets with the elements in $X$, and $X^*$ indicates the universe of all sequences over elements in $X$ (Kleene's Star).} 
\begin{definition}[Traces \& Event Logs]
Let $\ME$ be the universe of events.
A trace $\sigma$ is a sequence of events, i.e.\ $\sigma \in \ME^*$.
An event-log $\ML$ is a multiset of traces, i.e.\ $\ML \subset \BB(\ME^*)$.
\end{definition}
The predictive monitoring aims to estimate the future KPI values of the running cases. Here, we aim to be generic, meaning that KPIs can be of any nature:\footnote{Given a sequence $X$, $|X|$ indicates the length of $X$.}
\begin{definition}[KPI]
\label{def:KPI}
Let $\ME$ be the universe of events defined over a set $\MV$ of attributes. Let $\MW_K$ be the domain of the KPI values.
A KPI is a function $\MT: \ME^* \times \NN \not\rightarrow \MW_K$ such that, given a trace $\sigma \in \ME^*$ and an integer index $i \leq |\sigma|$, $\MT(\sigma,i)$ returns the KPI value of $\sigma$ after the occurrence of the first $i$ events. 
\end{definition}
Note that our KPI definition assumes it to be computed a posteriori, when the execution is completed and leaves a complete trail as a certain trace $\sigma$. In many cases, the KPI value is updated after each activity execution, which is recorded as next event in trace; however, other times, this is only known after the completion. We aim to be generic and account for all relevant cases. Given a trace  $\sigma = \langle e_1,\ldots,e_n \rangle$ that records a complete process execution, the following are three potential KPI definitions:
\begin{description}
\item [Remaining Time.]  $\MT_{remaining}(\sigma,i)$ is equal to the difference between the timestamp of $e_n$ and that of $e_i$. 
\item [Activity Occurrence.] It measures whether a certain activity is going to eventually occur in the future, such as an activity \emph{Open Loan} in a loan-application process. The corresponding KPI definition for the occurrence of an activity $A$ is $\MT_{occur\_A}(\sigma,i)$, which is equal to true if activity $A$ occurs in $\langle e_{i+1},\ldots,e_n \rangle$ and $i<n$; otherwise false.
\item [Customer Satisfaction.] This is typical KPI to analyze the customer journey~\cite{DBLP:conf/sac/TerragniH19}. Let us assume, without losing generality, to have a trace $\sigma = \langle e_1,\ldots,e_n \rangle$ where the satisfaction is known at the end, e.g.\ through a questionnaire. Assuming the satisfaction level is recorded with the last event - say $e_n(sat)$ - then, \linebreak $\MT_{cust\_satisf}(\sigma,i)=e_n(sat)$. 
\end{description}
The following definition states the prediction problem:
\begin{definition}[The Prediction Problem]
Let $\ML$ be an event log that records the execution of a given process, for which a KPI $\MT$ is defined. Let $\sigma = \langle e_1,\ldots,e_k \rangle$ be the  trace of a running case, which eventually will complete as  \mbox{$\sigma_T=\langle e_1,\ldots,e_k, e_{k+1}\ldots,e_n \rangle$}.
The prediction problem can be formulated as forecasting the value of $\MT(\sigma_T,i)$ for all $k < i \leq n$.
\end{definition}
As indicated in Section~\ref{sec:intro}, we aim to provide an explanation for the predictions. 

In particular, for each running case, we aim to return the set of attributes that influence its prediction the most,
with the corresponding magnitude and with the indication whether the attributes increase or decrease the predicted KPI's value.
Note that increasing the KPI value does not necessarily have a positive connotation because higher values do not necessarily mean better values. For instance, if the KPI is the process-instance cost, one wants to reduce the value. 

In the light of the above, for each trace $\sigma$, the problem can be stated as finding a (possibly different) function $\mathcal{K}_{(\sigma,\MT)}$  that, for each log attribute $a$, is defined for a subset of values of the domain of $a$, namely those that influence the prediction. 
In particular, for each pair $(a,v)$ in the domain of function $\mathcal{K}_{(\sigma,\MT)}$,  the absolute value of $\mathcal{K}_{(\sigma,\MT)}(a,v)$ indicates how much the fact that $a=v$
influences the KPI prediction.
The positive or negative sign of $\mathcal{K}(a,v)$ indicates whether the influence is towards increasing or decreasing the KPI value.
\begin{definition}[The Prediction-Explanation Problem]
\label{def:pred-expl}
Let $\ML$ be an event log over a set $\MV$ of attributes, with domains $\MW_\MV$.
Let $\sigma = \langle e_1,\ldots,e_k \rangle$ be a running case with a KPI prediction $\MT$. Let be $\overline{\MW} = \cup_{a \in \MV}~\MW_\MV(a)$.
Explaining the prediction is the problem of computing a function $\mathcal{K}_{(\sigma,\MT)}: \MV \times \overline{\MW} \not\rightarrow \R$ defined over a subset of attributes of $\MV$ that affects the prediction, with $\mathcal{K}_{(\sigma,\MT)}(a,v)$ defined only if $v \in \MW_\MV(a)$.
\end{definition}

\section{Related Works}
\label{sec:Related_works}

This section compares our framework with the literature. In particular, Section~\ref{sec:related-work-prediction} reports on the state of the art of Predictive Process Monitoring techniques, Section~\ref{sec:related-work-explanation-ml} discusses explainability approaches of Machine Learning models, while Section~\ref{sec:related-work-explanation-BPM} focuses on the application of explainability techniques in the Business Process Monitoring (BPM) field. 

\subsection{Prediction of Process-Related KPIs}
\label{sec:related-work-prediction}

The predictive-monitoring survey of M{\'{a}}rquez et al.~\cite{Marquez-Chamorro18} reports on the large repertoire of techniques and tools that were developed to address this problem.
However, the authors claim that \textit{``little attention has been given to [$\ldots$] explaining the prediction values to the users so that they can determine the best way to act upon''}, and that \textit{``it is necessary to develop tools that help users to query these models in order to get information that is relevant for them''}. 
These are in fact the problems tackled in this paper, so as to ensure that the predictive-monitoring system is trusted, and thus used. 

Predictive monitoring has been built on different machine and deep-learning techniques, and also on their ensemble~\cite{Marquez-Chamorro18}. Different research works have recently illustrated that Long Short-Term Memory networks (LSTMs) generally outperform other methods (see, e.g., ~\cite{Park19,TaxVRD17,LSTM_time}).  
However, since in production environments also the time is an important constraint, we wanted to leverage predictive models that could be trained in a shorter amount of time. In particular, we decided to rely on Catboost, which is a high-performance open source framework for gradient boosting on decision trees~\cite{Catboost} that showed competitive performances; a comparison between Catboost and LSTM predictive performances will be assessed in section~\ref{sec:performance_evaluation}.

Therefore, while our explanation framework is independent of the machine- or deep-learning technique that is employed, we operationalize it with Catboost. 
Section~\ref{sec:LSTM} provides further details on LSTMs, and details on how they are employed for business-process predictive monitoring, while Section~\ref{sec:Catboost} provides further details on Catboost algorithm.

\subsection{Explanation of Machine-Learning Models}
\label{sec:related-work-explanation-ml}

Some approaches exist in the literature to explain machine learning models, arisen from the need to understand complex black-box algorithms like ensembles of Decision Trees and Deep Learning~\cite{lime,shrikumar2017learning,shap,Shu:2019:DEF:3292500.3330935,lundberg2018explainable,DBLP:conf/iclr/2015,Meacham2019Explainable,gradients}. 
Conversely, other approaches exist in the literature which are naturally explainable~\cite{verenich_19_white_box,bohmer_2020_logo,brunk_2020_cause_vs_effect}, but these works do not really discuss the relevance and the quality of these explanations.   

The survey of Jagatheesaperumal et al.~\cite{Jagatheesaperumal_21} provides an overview of applications and challenges related to Big Data and AI in Industry 4.0, highlighting explainability as one of the main challenges in order to increase ML models adoption in industry.
The adoption of explanatory methods in industry is at an early stage; one of the most relevant works is by Rehse et al.~\cite{Rehse2019}, which also aims at providing a dashboard to process participants with predictions and their explanation. However, the paper does not provide sufficient details on the actual usage of the explainable-AI literature, and the very preliminary evaluation is based on one single artificial process that consists of a sequence of five activities. 
Sachan et al.~\cite{sachan20_explainable} apply a rule-based method in order to explain the chain of events leading to a decision for a loan application, but the approach requires the support of an expert's domain knowledge and the evaluation is based on a single business case study.
In \cite{Shu:2019:DEF:3292500.3330935} an approach of fake news detection grounded in explainability is introduced.

A significant amount of work in literature is focused on healthcare applications.
We highlight the work of Lundberg et al. \cite{lundberg2018explainable} that proposes an implementation of the Shapley Values in healthcare where the explanatory method is used to prevent hypoxaemia during surgery, and the work of Meacham et al. \cite{inbook}, where explainability is used for analysis of patience re-admittance.
In Coma{-}Puig and Carmona \cite{DBLP:journals/ml/Coma-PuigC22}, Shapley Values are leveraged in order to validate the correctness of the predictive approach in a utility company.
The SHAP implementation of the Shapley values has the strong theoretical foundation of the original game theory approach, with the advantage of providing offline explanations that are consistent with the online explanations. Moreover, SHAP avoids the problems in consistency seen in other explanatory approaches (e.g. the lack of robustness seen in the online surrogate models
\cite{alvarez2018robustness}).

\subsection{Explanations in the BPM field}
\label{sec:related-work-explanation-BPM}

The need to incorporate interpretability in addition to conventional performance measures (such as accuracy) when evaluating predictive models has been strongly emerging~\cite{DBLP:conf/icsoc/SindhgattaOM20} and several works have been focusing in comparing and evaluating explanations produced by different frameworks~\cite{Elkhawaga_22_XAI,Elkhawaga_22_issues,Stevens_22,10.1007/978-3-030-91431-8_4,10.1007/978-3-030-79108-7_8}.
The explainable survey of Stierle et al.~\cite{DBLP:conf/ecis/StierleBWZM021} reports on the repertoire of techniques that were developed to address this problem.
However, the authors claim that \textit{''future works should assess the techniques in real-life settings with end-users'', and that ''two works presented a case study applying a real-life data set, yet, without deploying the technique in the companies. So far, no work exists with an experimental evaluation method.''}
Rizzi et al.~\cite{rizzi_explainable_22} are among the first to investigate whether users actually understand the explanation plots returned by XAI techniques in the context of Predictive Process Monitoring; however, they investigated the intelligibility of several explanation plots relying only on 8 participants, and without deploying a concrete explainable framework inside a company. 

This paper not only proposes a framework able to solve some state-of-the-art current limitations, but also evaluates with process analysts (both from the academy and the industry domains) if they actually understand and feel comfortable with the results returned by our explainable predictive monitoring framework; moreover, the framework is also deployed in a real company, providing business stakeholders with online operational support of their processes.

Breuker et al.~\cite{10.1007/978-3-319-15895-2_46} also try to tackle the problem, but their attempt is not independent of the actual technique employed for predictions; furthermore, their explanations are only based on the activity names, while the explanations can generally involve resources, time, and more (cf.\ the case studies reported in Section~\ref{sec:experiments}).
Hsieh et al.~\cite{DBLP:conf/icpm/HsiehMO21} and Huang et al.~\cite{huang2021counterfactual} apply a counterfactual approach, which provides an understanding on what could have been done differently in order to achieve the desired outcome (e.g. having a loan approved), but their approach focuses only on providing explanations for categorical KPIs.
Rizzi et al.~\cite{Rizzi_20_Explainable} propose an approach based on LIME, but they can provide explanations only for nominal KPIs and their main focus is on improving predictive model accuracy rather than highlighting model drivers.

The framework proposed in this paper specializes the use of Shapley values to the problem of providing explanations for predictive analytics. Furthermore, the use of SHAP enabled us to provide explanations independently from the leveraged predictive model (\textit{model agnostic}), while \textit{model specific} approaches are specifically designed for certain model types; in particular, some approaches~\cite{harl_explainable,DBLP:conf/bpm/WeinzierlZBRM020,DBLP:conf/bpm/SindhgattaMOB20,hanga_2020_interpretable_graphs,DBLP:conf/icpm/Pasquadibisceglie21,WICKRAMANAYAKE2022108773} can provide explanations only for neural network models.
Other works~\cite{DBLP:conf/iclr/2015,DBLP:conf/bpm/SindhgattaMOB20} use attention mechanisms, which also have the limitation that is linked to the lack of consensus that attention weights are always correlated to feature importance. Jain et al.~\cite{DBLP:conf/naacl/JainW19} find it \textit{``at best, questionable – especially when a complex
encoder is used, which may entangle inputs in the hidden space''}, while Serrano et al.~\cite{serrano-smith-2019-attention} state that \textit{``attention  weights  often  fail  to  identify the sets of representations most important to the  model’s  final  decision''}.

\section{Algorithms for Process Predictive Analytics}
\label{sec:predictive}

The proposed framework is independent of the machine- or deep-learning technique that is employed to make the predictions; however, we had to instantiate it to prove its effectiveness. 
Here, we based our explainable predictive monitoring framework on two different algorithms, LSTM and Catboost, in order to demonstrate that our approach can be easily generalized and can be applied on top of any predictive technique. 

Process predictive analytics falls into the problem of supervised learning that aims to learn the model from a training set, for which the values of the dependent variables are known, whereas the value of an independent variable needs to be predicted. Training sets are composed by pairs $(x,y) \in \mathcal{X} \times \mathcal{Y}$ where $\mathcal{X}$ represents the independent variables (also known as \textbf{features}) with their values, and $\mathcal{Y}$ is the value observed for the dependent variable (i.e.\ the value to predict).

Process predictive analytics requires a KPI definition $\overline{\MT}$ as input (cf.\ Definition~\ref{def:KPI}).
The dependent variable $\mathcal{Y}$ takes on a value from the domain of possible KPI values, which corresponds to the image of $\overline{\MT}$: namely $\mathcal{Y}={img}(\overline{\MT})$.
Conversely, the characteristics and nature of $\mathcal{X}$ depends on the AI technique that is used for prediction. In abstract terms, each prediction technique requires the definition of the domain $\mathcal{X}$ and, a \textbf{trace-to-instance encoding function} $\rho: \ME^* \rightarrow \mathcal{X}$, which maps each (prefix of a) trace $\sigma$ in an element $\rho(\sigma)\in\mathcal{X}$.

The prediction model is trained off-line via a dataset $\mathcal{D}$ that is created from an event log $\ML$ as follows. Each prefix $\sigma'$ of each each trace $\sigma \in \ML$ generates one distinct item in $\mathcal{D}$ consisting of a pair $(x,y) \in (\mathcal{X} \times \mathcal{Y})$
 where $x=\rho(\sigma')$ and $y=\overline{\MT}(\sigma,|\sigma'|)$.
Once the dataset item of every trace prefix is created, the model is trained. The resulting prediction model (a.k.a.\ predictor) can be abstracted as an oracle function $\Phi_\mathcal{D}:\mathcal{X} \rightarrow \mathcal{Y}$.

The on-line phase aim is to predict the KPI of interest for a set of running cases of the process, identified by a set $\ML'$ of partial traces (i.e.,\ a log). It relies on the process predictor $\Phi_\mathcal{D}$: for each $\sigma' \in \ML'$, the predicted KPI value of the process instance identified by $\sigma'$ is $\Phi_\mathcal{D}(\rho(\sigma'))$.

\subsection{The use of LSTM for Predictive Monitoring}
\label{sec:LSTM}
A first instantiation of our framework has leveraged on LSTM models~\cite{Hochreiter1997}, a specific kind of Recurrent Neural Networks. LSTM models natively support the predictions where the independent variables are sequences of elements, and the literature has shown that they are among the most suitable methods for predictive business monitoring (cf.\ Section~\ref{sec:related-work-prediction}). 

In the case of LSTM, $\mathcal{X}$ consists of sequences of vectors with a certain number $n$ of dimensions, i.e.\ $\mathcal{X}={(\R^n)}^*$\footnote{In literature, LSTMs are often trained on the basis of matrices. However, a sequence of $m$ vectors in $\R^n$ can be seen, in fact, as a matrix in $\R^{n,m}$. We use here the dataset representation as vectors to simplify the formalization.}.
The definition of the trace-to-instance encoding function requires the intermediate concept of event-to-vector function $\zeta: \ME \rightarrow \R^n$, which encodes each event defined over a set $\MV$ of attributes. Each numerical and boolean attribute $a$ becomes one feature, element of vector $\zeta(e)$. Each literal attributes $a$ is instead represented through the so-called \textit{one-hot encoding}: one different dimension exists for each value $v \in \MW_\MV(a)$, and the dimension referring to value $e(a)$ takes on value $1$, with the other dimensions be assigned value $0$.
The function $\rho$ is then defined as: $\rho(\langle e_1, \ldots, e_m \rangle)=[ \zeta(e_1), \ldots, \zeta(e_m) ]$.

\subsection{The use of Catboost for Predictive Monitoring}
\label{sec:Catboost}

Catboost is a high-performance open source framework for gradient boosting on decision trees~\cite{Catboost}, which outperforms and solves limitations of current state-of-the-art implementations of gradient boosted decision trees.
Gradient boosting is essentially a process of constructing an ensemble predictor; the main difference with other ensemble methods, such as Random Forest~\cite{RandomForest}, is that instead of having parallel weak learners trying to predict the result on the given dataset, the models are built sequentially and then gradient descent is used to keep building new and better models.
It is backed by solid theoretical results that explain how strong predictors can be built by iteratively combining weaker models (base predictors) in a greedy manner.
Catboost, in particular, at each iteration $t$ of the algorithm performs a random permutation of the features and a tree is constructed on the basis of it. 
Moreover, for each split of a tree, CatBoost combines (concatenates) all categorical features (and their combinations) already used for previous splits in the current tree with all categorical features in the dataset. 

In the domain of Catboost learning, $\mathcal{X}$ (the independent variables) consists of a vector with a certain number $n$ of dimensions, i.e.\ $\mathcal{X}={(\R^n)}$.
The \textit{event-to-vector encoding function} $\zeta: \ME \rightarrow \R^n$ is defined as $\zeta(e) = \displaystyle\bigoplus_{a \in \MV} e(a)$, where each numeric, categorical and boolean attribute $a$ of event $e$ is encoded as a different dimension of $\zeta(e)$, which takes on value $e(a)$.

Moreover, since the final outcome of the process may be influenced by the previously occurred events, it could be important to take into consideration also the history of the process. Therefore, the function $\rho$ is defined differently depending on the number  of past events considered. Given a trace $\langle e_1, \ldots, e_m \rangle$ and a number $0 \leq k < m$ of past events considered, we define  $\rho^k(\langle e_1, \ldots, e_m \rangle)=\langle \zeta(e_{m-k}) \bigoplus \ldots \bigoplus \zeta(e_{m}) \rangle$, where $1 \leq k < m$ represents the number of historical past events to be considered. The parameter $k$ can be optimized depending on the dataset using techniques of hyper-parameter optimization, of which an example of application is discussed in Section~\ref{sec:experiments}. Note that, if $k=0$, only the last occurred event is used.

Alternatively, an aggregated history of each trace $\sigma$ can also be considered, encoding the number of times that each activity has been performed in $\sigma$. To this aim, we define a function $\rho^{aggr}(\langle e_1, \ldots, e_m \rangle)$; here, for each activity $act \in A$, one dimension exists in $\rho^{aggr}(\sigma): \ME^* \rightarrow (\NN)^{|A|}$ that takes on a value equal to the number of events $e \in \sigma$ that refer to $act$, i.e.\ such that $e(a)=act$.
The function $\rho$ is then defined as: $\rho(\langle e_1, \ldots, e_m \rangle)=\rho^{aggr}(\langle e_1, \ldots, e_m \rangle) \bigoplus \zeta(e_m)$.

\section{Explainable Predictive Process Analytics}
\label{sec:Explainable}

This section reports on the details of the algorithm that was used to explain the predictions, and how Shapley values can be adapted to explain any predictive model.  

Section~\ref{sec:shapTheory} introduces the theory behind Shapley values, while Section~\ref{sec:sv} illustrates its application and adaptation for predictive process monitoring. Then, in Section~\ref{sec:expl_approach} we provide the general picture and the two main types of explanations reported.

\subsection{The Theory of Shapley Values}
\label{sec:shapTheory}

The Shapley Values \cite{shapley1953value} is a game theory approach to fairly distribute the payout among the players that have collaborated in a cooperative game. This theory can be adapted as an approach to explain a predictive model. 
The assumption is that the features from an instance correspond to the players, and the payout is the difference between the prediction made by the predictive model and the average prediction (also called {\em base value}). 
Intuitively, given a predicted instance, the Shapley Value of a feature expresses the amount of contribution of the feature value to the model prediction~\cite{molnar2020interpretable}: 

\begin{definition}[Shapley Value]
Let $F=\{f_1,\ldots,f_m\}$ be the set of features used by the oracle function $\Phi: X_1\times\ldots\times X_m \rightarrow \mathcal{Y} $ to predict a KPI. 
The Shapley value for feature $f_i$ which assumes value $x_i\in X_i$ is defined as:
\begin{equation*}
\resizebox{\columnwidth}{!}{$
   \psi_i=\sum_{F' \subseteq F \setminus\{f_i\}}\frac{|F'|!\left(m-|F'|-1\right)!}{m!}\left(val\left(F'\cup\{f_i\}\right)-val(F')\right)
   $}
\end{equation*}
where $val(F')$ is the so-called payout for only using the set of feature values in $F' \subset F$ in making the prediction.
 \label{def:SV}
\end{definition}

Intuitively, the formula in Definition~\ref{def:SV} evaluates the effect of incorporating the value $x_i\in X_i$ of the feature $f_i$ into any possible subset of the feature values considered for prediction. In the equation, set $F$ runs over all possible subsets of feature values, the term $val\left(F'\cup\{f_i\}\right)-val(F')$
corresponds to the marginal value of adding the feature $f_i$ which assumes value $x_i$ in the prediction using only the set of feature values in $F$,
and the term
$\frac{|F'|!\left(m-|F'|-1\right)!}{m!}$ corresponds to all the possible permutations with subset size $|F'|$, to weight different sets in the formula.
This way, all possible subsets of attributes are considered, and the corresponding effect is used to compute the Shapley Value of $x_i$. 

\subsection{Explainable Predictions through Shapley Values for Catboost}
\label{sec:sv}

This paper only reports on the explanation of predictions for the Catboost method, while the explanations for LSTM model are reported in~\cite{galanti2020explainable}. This choice is motivated by the fact that, as further discussed in Section~\ref{sec:performance_evaluation}, Catboost is preferrable to LSTM for prediction, due to the increased speed to build the model while keeping similar accuracy. 

The starting point is the trace-to-instance encoding function $\rho: \ME^* \rightarrow \mathcal{X}$, which maps each (prefix of a) trace $\sigma$ in an element $\rho(\sigma)\in\mathcal{X}$ (cf.\ Section~\ref{sec:predictive}). 
Let us recall that given a trace $\sigma=\langle e_1, \ldots, e_m \rangle$,  $\rho(\sigma)=[x^1, \ldots,x^n]$, each feature $f^i$ has an associated value $x^i$.

When applied for explainable predictive monitoring, the Shapley values for a trace $\sigma$ are computed over $\rho(\sigma)=[x^1, \ldots,x^n]$, thus resulting in a vector of Shapley values $\Psi=[\psi^1, \ldots ,\psi^n]$, with $\psi^i$ being the Shapley value of feature $f^i$. In accordance with the Shapley values theory, the explanation of $\psi^i$ is as follows: since feature $f^i=x^i$, the KPI prediction deviates $\psi^i$ units from the average KPI value observed in the executions recorded in the event-log.
Please note that any Shapley value $\psi^i$ can be either positive or negative. A positive or negative value indicates that the feature contributes to increasing or decreasing the value, respectively.

The computation is the Shapley value is repeated for each trace of $\ML$. However, if $f^i$ is numerical, several different values can be observed for $f^i$, yielding
a large number of explanations $f^i=x^i_1, \ldots f^i=x^i_k$. Some of these explanations are equivalent from a domain viewpoint: e.g., $amount=10000$, $amount=10050$ might be referring to the same class of amount in a loan application. Therefore, $q$ representative values $w^i_1, \ldots, w^i_q$ are selected out of values $x^i_1, \ldots x^i_k$ (namely, with $q \ll k$) so as to obtain explanations of type $f^i<w^i_1$, $w^i_1 \leq f^i <w^i_2$, $\ldots$, $f^i \geq w^i_q$. Values $w^i_1, \ldots, w^i_q$ can be obtained taking the boundaries of the buckets obtained via discretization techniques. In particular, our implementation operationalizes a discretization of each feature $f^i$ on the basis of decision/regression as follows. The training set consists of tuple with only two features: $f^i$ used as the independent variable, and  the KPI as target/dependent variable. The values observed at the splits of the tree nodes induce the boundaries and, consequently, the buckets.

While an exact computation of the Shapley values requires to consider all combinations of features (hence, the algorithm is exponential on the number of features), efficient estimations can be obtained through polynomial algorithms that use greedy approaches~\cite{molnar2020interpretable}.

\subsection{Overall Approach for Explaining Generic KPI Predictions}
\label{sec:expl_approach}

Explanations can be calculated both offline (i.e. on cases that have been already completed) and online (on cases that have not been completed yet). 
When used offline they are calculated on the test dataset, a part of the dataset not used for training the model (information about the division between train and test sets will be provided in Section~\ref{sec:experiments}). 
Moreover, they can be used with different goals; on one hand, explanations for a set of cases can be aggregated, to provide the stakeholder a global picture of what are the features/factors that the trained model uses to make predictions. This type of explanation is called {\em Global}, and is very important for gaining trust on predictive process monitoring. 
On the other hand, {\em Local} online explanations of a certain prediction can be provided, when there is the need to focus on single running cases with a high risk of failing the desired KPI.

\subsubsection{\textbf{Global Explanations}}
\label{sec:global_explanations}

Our global explanation strategy is to provide a bar chart that overviews the importance of each factor influencing the predictive model. 
The global explanation strategy can be applied both during the offline and the online phase.

In the offline phase, given an event log $L$, we consider each prefix $\sigma'$ of each completed trace in $L$ related to the test set. Then, we compute the explanations as defined in Section~\ref{sec:sv}.
At this point, for each explanation we consider the vector of all the Shapley values $\Psi$ associated to it and we compute the average, obtaining the average influence of a particular explanation over all the instances; we then sort the explanations by descending influence, allowing us to find the most relevant explanations influencing the predictive model.
Note that this is an improvement compared to~\cite{galanti2020explainable}, where a custom threshold selected by the user was applied in order to find the most relevant features and where the real contribution of the features (i.e. how much they were contributing to increase or decrease the prediction) was not taken into account.
In this way, not only do we aim to find the most relevant features, but we also report on their contribution to the output of the predictive model.

\begin{figure}[t!]
    \includegraphics[width=1 \columnwidth]{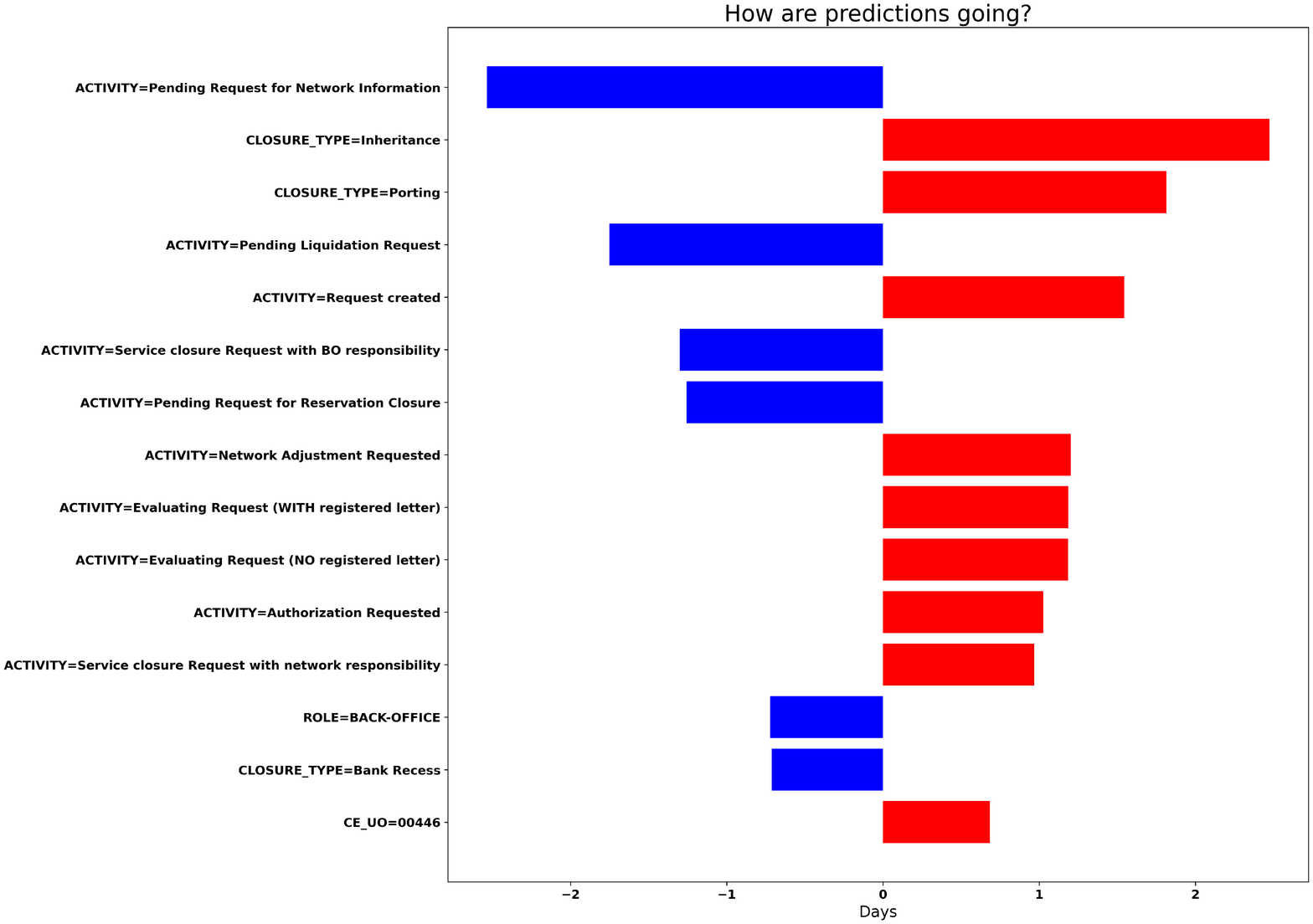}
    \caption{Example of global offline explanations for remaining time prediction}
    \label{fig:explanation_bac_rem_time}
\end{figure}

Figure~\ref{fig:explanation_bac_rem_time} shows an example of our global explanation strategy for the remaining time prediction, represented as a bar chart reporting the most relevant explanations influencing the prediction during the offline phase.
The $y$ axis lists different explanations of types \mbox{$attr=value$}, namely a combination of an attribute with an associated value that was found to be relevant for the predictive model, while the $x$ axis represents the average influence on the predicted time for the considered traces.
For instance, let us consider the explanation \textit{Closure\_type=Inheritance}, which is associated with a red bar with length greater than 2 days. This indicates that when this association attribute-value (explanation) occurs, then it contributes increasing 
the KPI value by more than 2 days (i.e. when an instance of a process with Closure type Inheritance occurs, then more time will be needed on average to finish the process).

A similar reasoning can be repeated for global explanations referring to the online phase. 
For each running case we take the vector of the Shapley values $\Psi$ and associated explanations related to the last observed prefix (which has also the information about previous prefixes); afterwards, for each explanation, we calculate the average value, to obtain the average influence of the explanation over all running cases.

\subsubsection{\textbf{Local Explanations}} 

Our explanation strategy can be also applied in order to generate {\em Local} explanations of a certain prediction. This is particularly useful when there is the need to focus on single running cases that present a high risk of failing the desired KPI.
When we focus on single running cases, we generate a bar chart in the same way we described in the previous section; the only difference is that in order to analyze single running cases, we need to use an interactive tool that enables us to easily perform our analysis.
Note that this is an improvement compared to~\cite{galanti2020explainable}, where local explanations were displayed in a Table instead of a bar chart; after an initial interview phase, business process stakeholders indicated to prefer having an uniform view for both local and global explanations.

Further details on the visualization of local explanations and on the integration of our framework with the IBM Process Mining Suite will be provided in Section~\ref{sec:integration}.

\section{Assessment of the Quality of the Predictions and Explanations}
\label{sec:experiments}
This section reports on the results of the empirical evaluation of our explainable predictive monitoring framework.
We acknowledge that a large share of the novelty lays on providing predictions with explanations of the factors that influence them. 

However, it is also important to report on the quality of the predictions, to illustrate that the explanations are based on high-quality predictions.
This assessment is based on a number of different processes, which are described in Section~\ref{sec:Domain description}.
Section~\ref{sec:performance_evaluation} discusses how predictive models were built, using Catboost and LSTM and leveraging on hyperparameter optimization, and reports on the prediction quality. 

Finally, Section~\ref{sec:explanation_evaluation} shows the nature of the explanations for the different domains. 

\subsection{Domain description}
\label{sec:Domain description}

\begin{table*}[t!]
\caption{Event logs statistics}
\footnotesize
\centering
\resizebox{\textwidth}{!}
{\begin{tabular}{|l|l|l|l|l|l|l|}
\hline
\cellcolor{LightCyan}\textbf{Event Log} &
\cellcolor{LightCyan}{\textbf{\begin{tabular}[c]{@{}c@{}}\#\\ traces\end{tabular}}} &
\cellcolor{LightCyan}{\textbf{\begin{tabular}[c]{@{}c@{}}\#\\ activities\end{tabular}}} &
\cellcolor{LightCyan}{\textbf{\begin{tabular}[c]{@{}c@{}}mean\\ events/trace\end{tabular}}} &
\cellcolor{LightCyan}{\textbf{\begin{tabular}[c]{@{}c@{}}median\\ events/trace\end{tabular}}} &
\cellcolor{LightCyan}{\textbf{\begin{tabular}[c]{@{}c@{}}mean\\ duration\end{tabular}}} &
\cellcolor{LightCyan}{\textbf{\begin{tabular}[c]{@{}c@{}}std deviation\\ duration\end{tabular}}} \\
\hline
 \textbf{Bank Account Closure} & 
 32429 & 
 15 & 
 5.5 & 
 7 & 
 15.5 days & 
 33 days  \\ 
\hline
 \cellcolor{LightGray}\textbf{BPIC 2012}  &
 \cellcolor{LightGray}12369 & 
 \cellcolor{LightGray}23 & 
 \cellcolor{LightGray}14  & 
 \cellcolor{LightGray}8 & 
 \cellcolor{LightGray}7.9 days &
 \cellcolor{LightGray}11.7 days  \\ \hline
\textbf{BPIC 2012 - W}  & 9658  & 6  & 7.5 & 6 & 11.4 days & 12.7 days  \\ \hline
\cellcolor{LightGray}\textbf{BPIC 2013}  & 
\cellcolor{LightGray}7554  & 
\cellcolor{LightGray}13 & 
\cellcolor{LightGray}8.7 & 
\cellcolor{LightGray}6 & 
\cellcolor{LightGray}12.1 days & 
\cellcolor{LightGray}28.6 days  \\  \hline
\textbf{HelpDesk 2017}  & 4580  & 14 & 4.7 & 4 & 40.9 days & 8.4 days \\ \hline
\cellcolor{LightGray}\textbf{Fine Management}  & 
\cellcolor{LightGray}125926  & 
\cellcolor{LightGray}11 & 
\cellcolor{LightGray}4 & 
\cellcolor{LightGray}5 & 
\cellcolor{LightGray}384 days & 
\cellcolor{LightGray}362 days \\  \hline
\end{tabular}}
\label{tab:dataset_statistics}
\end{table*}

For our assessment we used 6 real-life event logs, which are described below, while the event-logs statistics are shown in Table~\ref{tab:dataset_statistics}.

\begin{itemize}
    \item \textbf{\emph{Bank Account Closure}} is a process executed at an Italian banking institution. The process deals with the closure of customer's accounts, which may be requested either by the customer or by the bank, for several reasons. 
    \item \textbf{\emph{BPIC 2012}}\footnote{http://dx.doi.org/10.4121/uuid:3926db30-f712-4394-aebc-75976070e91f} is a real-life log of a Dutch Financial Institute. It represents an application process for a personal loan or overdraft within a global financing organization.
    \item \textbf{\emph{BPIC 2012-W}} is the dataset derived from bpi12 challenge, and it represents the subprocess containing only the states of the work items belonging to the application.
    \item \textbf{\emph{BPIC 2013}}\footnote{http://dx.doi.org/10.4121/uuid:500573e6-accc-4b0c-9576-aa5468b10cee} is a dataset extracted from Volvo IT Belgium incident management system.
    \item \textbf{\emph{HelpDesk 2017}}\footnote{ https://doi.org/10.4121/uuid:0c60edf1-6f83-4e75-9367-4c63b3e9d5bb} is a real-life log of SIAV s.p.a. company in Italy, and represents instances of a ticketing process in the company helpdesk area.
    \item \textbf{\emph{Road Traffic Fine Management Process}}\footnote{http://dx.doi.org/10.4121/uuid:270fd440-1057-4fb9-89a9-b699b47990f5} is a real-life event log of an information system managing road traffic fines.
\end{itemize}

\subsection{Predictive Quality Evaluation}
\label{sec:performance_evaluation}

The framework for explainable predictive monitoring has been implemented in Python, using Pandas to elaborate the data, and the Shap library\footnote{\url{https://shap.readthedocs.io/en/latest}} to explain the prediction.
For the LSTM implementation we rely on the Keras framework.
For the Catboost implementation, instead, we leveraged the open source library available.\footnote{https://catboost.ai/} 

We used two/thirds of the traces as training, and one third as test set.
For Catboost, in order to improve the quality of the trained model, we used hyperparameter optimization, with 20\% of the training data employed for this.
In particular, 3 different hyperparameters were tuned on the validation set: the number of trees used (which varied between 1500, 3000 and 4000), the depth of each single tree (3, 6 and 10) and the number of past events to be considered in a partial trace when predicting. 
In fact, as stated in Section~\ref{sec:sv}, a particular encoding is needed in tree-based models in order to include information about past events (i.e. the history of the partial trace); here, in the last observed event of the partial trace, for every event-level attribute of each past event that previously occurred, a different dimension (feature) is added. 
Therefore, since LSTM naturally leverages information about past events, before proceeding to the comparison between Catboost and LSTM, a preliminary study was necessary in order to understand how many past events we should consider when predicting using the Catboost model. In particular, we focused on the scores obtained on the validation set; ideally, we should take the number of events associated to the point with the lowest MAE (when predicting a numerical KPI) or associated to the point with the highest F1 score (when predicting a boolean KPI).
The number of considered events varies from 0 (no information related to past events is used) to the average number of events per case, which was observed considering all completed process executions. 
Moreover, we also considered in our evaluations a different encoding of the history of each partial trace $\sigma$ called \textit{aggregated history} (cf.\ Section~\ref{sec:Catboost}).

However, training and calculating the scores for each predictive model based on a different configuration of considered past events is a heavy procedure (especially in production environments). Therefore, we elaborated a more lightweight strategy; in particular, if the predictive quality does not improve (i.e. it is worse compared to the previous configuration or it improves by a percentage lower than 1\%) for two consecutive steps, we stop our research and we take the best predictive model that was observed until that moment.
We then compared the two approaches in order to understand if our heuristic was able to find an optimal accuracy without trying all possible combinations, and we reported the results in Table~\ref{tab:hist_vs_no_hist_catboost}, where we also highlight the comparison with the model that was not leveraging historical information.
The first two columns indicate the event log and the considered KPI, while the last three columns show the results obtained by Catboost models without considering history, considering the history according to our heuristic and according to the complete procedure respectively.  
We recall that when the KPI is numerical, then the scores represent the MAE (expressed in days for Remaining Time prediction and in Euros for Total Cost prediction); conversely, when the KPI is Boolean (Activity occurrence prediction), the scores represent the F1.
In the last two columns it is also reported the number of previous events considered by the predictive model that were found to have the best predictive quality on the validation set. 
We can clearly see that in 11 cases out of 17 our heuristic approach found the same number of past events as the complete procedure, with a similar predictive quality in other cases, showing that our approach can be used to find a good configuration but reducing the computational time at the same moment.
Moreover, in half of the cases, leveraging historical information can help to achieve a slightly higher accuracy.

\begin{table*}[t!]
\caption{Catboost predictive quality when leveraging historical information. The scores represent the MAE when the KPI is numerical; conversely, when the KPI is Boolean (Activity occurrence prediction), the scores represent the F1.}
\footnotesize
\centering
\resizebox{\textwidth}{!}
{\begin{tabular}{|l|l|l|l|l|}
\hline
\cellcolor{LightCyan}\textbf{Event Log} &
\cellcolor{LightCyan}\textbf{KPI} &
\cellcolor{LightCyan}{\textbf{\begin{tabular}[c]{@{}c@{}}Catboost\\ (no history)\end{tabular}}} & 
\cellcolor{LightCyan}{\textbf{\begin{tabular}[c]{@{}c@{}}Catboost\\ (heuristic)\end{tabular}}} &
\cellcolor{LightCyan}{\textbf{\begin{tabular}[c]{@{}c@{}}Catboost\\ (complete)\end{tabular}}} \\
\hline
 \textbf{Bank Account Closure} & 
 Remaining Time (MAE) & 
 \textbf{4.18} & 
 4.20 (History: 3) & 
 4.20 (History: 3) \\ 
\hline
 \cellcolor{LightGray}\textbf{Bank Account Closure}  &
 \cellcolor{LightGray}Total Cost (MAE) &
 \cellcolor{LightGray}0.92  & 
 \cellcolor{LightGray}0.92 (History: aggr)& 
 \cellcolor{LightGray}0.92 (History: aggr)\\ 
\hline
 \textbf{Bank Account Closure} & 
 Activity \textit{Authorization Requested} (F1) & 
 0.95 & 
 \textbf{0.96 (History: 1)} & 
 \textbf{0.96 (History: 1)} \\ 
\hline
\cellcolor{LightGray}\textbf{Bank Account Closure}  &
 \cellcolor{LightGray}Activity \textit{Pending Request for acquittance of heirs} (F1) &
 \cellcolor{LightGray}0.88 & 
 \cellcolor{LightGray}0.88 (History: aggr)& 
 \cellcolor{LightGray}0.88 (History: aggr)\\ \hline
 \textbf{Bank Account Closure} & 
 Activity \textit{BO Adjustment Requested} (F1) & 
 0.66 & 
 0.66 (History: 0) & 
 \textbf{0.67 (History: 6)} \\ 
\hline
\cellcolor{LightGray}\textbf{BPIC 2012}  & 
\cellcolor{LightGray}Remaining Time (MAE) & 
\cellcolor{LightGray}6.61 & 
\cellcolor{LightGray}6.56 (History: aggr) & 
\cellcolor{LightGray}\textbf{6.53 (History: 4)}\\  
\hline
\textbf{BPIC 2012} & 
 Activity \textit{A\_ACCEPTED} (F1) & 
 0.69 & 
 0.69 (History: 1) & 
 \textbf{0.70 (History: 3)} \\ 
\hline
\cellcolor{LightGray}\textbf{BPIC 2012}  &
 \cellcolor{LightGray}Activity \textit{A\_CANCELLED} (F1) &
 \cellcolor{LightGray}0.55 & 
 \cellcolor{LightGray}\textbf{0.56 (History: 1)} & 
 \cellcolor{LightGray}\textbf{0.56 (History: 1)}\\ 
\hline
 \textbf{BPIC 2012} & 
 Activity \textit{A\_DECLINED} (F1) & 
 0.56 & 
 0.56 (History: 0) & 
 0.56 (History: 13) \\ 
\hline
\cellcolor{LightGray}\textbf{BPIC 2012 - W}  &
 \cellcolor{LightGray}Remaining Time (MAE) &
 \cellcolor{LightGray}7.38  & 
 \cellcolor{LightGray}\textbf{7.36 (History: 1)}& 
 \cellcolor{LightGray}\textbf{7.36 (History: 1)}\\ 
\hline
\textbf{BPIC 2013} & 
 Remaining Time (MAE) & 
 9.74 & 
 \textbf{9.58 (History: 2)} & 
 \textbf{9.58 (History: 2)} \\ 
\hline
 \cellcolor{LightGray}\textbf{BPIC 2013}  &
 \cellcolor{LightGray}Activity \textit{Push to front (2°/3° line)} (F1) &
 \cellcolor{LightGray}0.87 & 
 \cellcolor{LightGray}0.87 (History: 0)& 
 \cellcolor{LightGray}0.87 (History: 2)\\ 
\hline
 \textbf{BPIC 2013} & 
 Activity \textit{Wait \- User} (F1) & 
 0.69 & 
 0.69 (History: 0) & 
 0.69 (History: 5) \\ 
\hline
\cellcolor{LightGray}\textbf{HelpDesk}  & 
\cellcolor{LightGray}Remaining Time (MAE) & 
\cellcolor{LightGray}5.27 & 
\cellcolor{LightGray}5.27 (History: 0) & 
\cellcolor{LightGray}5.27 (History: 0)\\  
\hline
\textbf{Fine Management} & 
 Remaining Time (MAE) & 
 196.93 & 
 \textbf{185.69 (History: 2)} & 
 \textbf{185.69 (History: 2)} \\ 
\hline
\cellcolor{LightGray}\textbf{Fine Management}  &
 \cellcolor{LightGray}Activity \textit{Send for Credit Collection} (F1) &
 \cellcolor{LightGray}0.81 & 
 \cellcolor{LightGray}0.81 (History: aggr)& 
 \cellcolor{LightGray}0.81 (History: aggr)\\
\hline
\textbf{Fine Management} & 
 Activity \textit{Send Appeal to Prefecture} (F1) & 
 0.43 & 
 0.37 (History: aggr) & 
 0.37 (History: aggr) \\ 
 \hline
\end{tabular}}
\label{tab:hist_vs_no_hist_catboost}
\end{table*}

\begin{table*}[t!]
\caption{Predictive quality comparison between Catboost and LSTM. The scores represent the MAE (expressed in days for Remaining Time prediction and in Euros for Total Cost prediction) when the KPI is numerical; conversely, when the KPI is Boolean (Activity occurrence prediction), the scores represent the F1.}
\footnotesize
\centering
\resizebox{\textwidth}{!}
{\begin{tabular}{|l|l|l|l|}
\hline
\cellcolor{LightCyan}\textbf{Event Log} &
\cellcolor{LightCyan}\textbf{KPI} &
\cellcolor{LightCyan}\textbf{LSTM} &
\cellcolor{LightCyan}\textbf{Catboost} \\
\hline
 \textbf{Bank Account Closure} & 
 Remaining Time (MAE) & 
 4.37 & 
 \textbf{4.18} \\
\hline
 \cellcolor{LightGray}\textbf{Bank Account Closure}  &
 \cellcolor{LightGray}Total Cost (MAE) &
 \cellcolor{LightGray}0.95 & 
 \cellcolor{LightGray}\textbf{0.92} \\
\hline
 \textbf{Bank Account Closure} & 
 Activity \textit{Authorization Requested} (F1) & 
 \textbf{0.99} &
 0.96 \\ 
\hline
\cellcolor{LightGray}\textbf{Bank Account Closure}  &
 \cellcolor{LightGray}Activity \textit{Pending Request for acquittance of heirs} (F1) &
 \cellcolor{LightGray}\textbf{0.90} & 
 \cellcolor{LightGray}0.88 \\ \hline
 \textbf{Bank Account Closure} & 
 Activity \textit{BO Adjustment Requested} (F1) & 
 0.65 & 
 \textbf{0.67} \\ 
\hline
\cellcolor{LightGray}\textbf{BPIC 2012}  & 
\cellcolor{LightGray}Remaining Time (MAE) & 
\cellcolor{LightGray}6.66 &
\cellcolor{LightGray}\textbf{6.53}\\  
\hline
\textbf{BPIC 2012} & 
 Activity \textit{A\_ACCEPTED} (F1) & 
 0.60 & 
 \textbf{0.70} \\ 
\hline
\cellcolor{LightGray}\textbf{BPIC 2012}  &
 \cellcolor{LightGray}Activity \textit{A\_CANCELLED} (F1) &
 \cellcolor{LightGray}0.37 & 
 \cellcolor{LightGray}\textbf{0.56}\\ 
\hline
 \textbf{BPIC 2012} & 
 Activity \textit{A\_DECLINED} (F1) & 
 0.51 & 
 \textbf{0.57} \\ 
\hline
\cellcolor{LightGray}\textbf{BPIC 2012 - W}  &
 \cellcolor{LightGray}Remaining Time (MAE) &
 \cellcolor{LightGray}7.84 & 
 \cellcolor{LightGray}\textbf{7.36} \\ 
\hline
\textbf{BPIC 2013} & 
 Remaining Time (MAE) & 
 11.82 & 
 \textbf{9.58} \\ 
\hline
 \cellcolor{LightGray}\textbf{BPIC 2013}  &
 \cellcolor{LightGray}Activity \textit{Push to front (2°/3° line)} (F1) &
 \cellcolor{LightGray}0.81 & 
 \cellcolor{LightGray}\textbf{0.87} \\
\hline
 \textbf{BPIC 2013} & 
 Activity \textit{Wait \- User} (F1) & 
 0.45 & 
 \textbf{0.69} \\ 
\hline
\cellcolor{LightGray}\textbf{HelpDesk}  & 
\cellcolor{LightGray}Remaining Time (MAE) & 
\cellcolor{LightGray}5.96 & 
\cellcolor{LightGray}\textbf{5.27} \\  
\hline
\textbf{Fine Management} & 
 Remaining Time (MAE) & 
 233 & 
 \textbf{185.69} \\ 
\hline
\cellcolor{LightGray}\textbf{Fine Management}  &
 \cellcolor{LightGray}Activity \textit{Send for Credit Collection} (F1) &
 \cellcolor{LightGray}0.77 & 
 \cellcolor{LightGray}\textbf{0.81} \\
\hline
\textbf{Fine Management} & 
 Activity \textit{Send Appeal to Prefecture} (F1) & 
 \textbf{0.52} & 
 0.43 \\
 \hline
\end{tabular}}
\label{tab:lstm_catboost_comparison}
\end{table*}

After discovering how many past events we should consider when predicting using the Catboost model, we illustrate in Table~\ref{tab:lstm_catboost_comparison} the comparison between the Catboost model (shown in the last column) and the LSTM model (shown in the second last column), which naturally leverages information about past events. 
As it can be seen, Catboost performs better than LSTM when predicting a numerical KPI, while it shows comparable results when predicting a boolean KPI. 
In the light of these results, we opted to integrate Catboost in the software suite of IBM, and to put LSTM models away. Explanations are also reported for our case study using Catboost, only.

The next section will report on the outcome of the global explanation strategy described in Section~\ref{sec:global_explanations} and applied on the new employed Catboost model in several domains, showing that our explanation technique is suitable to be generalized and adapted for any predictive model. 

\subsection{Explanation evaluation}
\label{sec:explanation_evaluation}

Section \ref{sec:global_explanations} showed that the explanation for a learnt prediction model is given as a bar chart; let us recall that the $y$ axis lists the different explanations, while the $x$ axis represents the average influence on the considered KPI. The explanations are ordered by descending absolute influence on the predicted KPI.
However, when predicting a boolean KPI, the values on the x axis represent a probability in the domain $[0:1]$; therefore, in order to have a representation of the influence of the explanations that was the same independently from the considered KPI, we shifted the probability in the domain $[-1:+1]$. As shown in Figure~\ref{fig:explanation_bac_auth_req}, this allowed us to give a more clear interpretation of the influence of the explanations.

This section is intended to describe the outcome of our global explanation strategy during the offline phase, in order to show the validity of the explanations provided in all the datasets that have been described in Section~\ref{sec:Domain description}. 

\subsubsection{Bank Account Closure}

For the bank, which deals with the closure of customer's accounts, it is of interest to obtain an estimate of the remaining time until the end for running cases. This allows the bank to decide which cases require special attention, in order to not postpone them too much further. Also, the bank wants to be informed whether there are high chances that one or more of the following activities will occur: \emph{Authorization Requested}, \emph{Pending Request for Acquittance of heirs}, and \emph{Back-Office Adjustment Requested}. They are linked to contingency actions, which should be avoided because they would cause inefficiencies in terms of time, costs, and resource utilization. Finally, the bank is also interested in obtaining an estimate of the total cost of a running case, in order to detect in advance which cases require particular attention.
Each trace is associated with an attribute \emph{Closure\_Type}, which encodes the type of procedure that is carried out for the specific account holder, and the \emph{Closure\_Reason}, namely the reason triggering the closure's request.

Figure~\ref{fig:explanation_bac_rem_time} was used in Section~\ref{sec:global_explanations} to give an impression on how global explanations are given. Here, we give some more information: it reports on the application of our explainable framework for remaining time prediction. 
The fact that the closure type is Inheritance (\emph{Closure\_Type=Inheritance}) is one of the largest factors that influences the prediction; the color red of the bar and the information that the value is positive (i.e. greater than 2 days) indicates that the influence is towards increasing the remaining time.
From a domain viewpoint, when the type of procedure is Inheritance, the bank-account holder is passed away. A further analysis of the data confirms this finding: if the type is \emph{Inheritance}, the process duration is 29 days, versus 14 days when the type is different. 
Also when the closure type is Porting (\textit{Closure\_type=Porting}), the influence is towards increasing the remaining time, and this is also confirmed analyzing the data, since the process duration related to this closure type is 24 days.
The evidence in the explanations illustrates that Catboost allowed learning a prediction model that leverages on the closure type to estimate the remaining time.

Other important explanations are related to the activities performed and the resources involved.
When activities are performed by a resource director (such as \textit{Authorization Requested}) or when the activity performed is \textit{Network Adjustment Requested} (that only occurs when an error is made in the early stages of the process), the behaviour is considered as exceptional; consequently, the cases usually take longer to complete. 
This is indicated by the two red bars in the rows \emph{ACTIVITY=Authorization Requested} and \emph{ACTIVITY=Network Adjustment Requested}, which indicate that the average influence is towards increasing the remaining time by 1 day. Even in this case our framework was able to learn to correctly identify the influencers of the process. 

Finally, some explanations are not related to an exceptional behaviour, but rather to the structure of the process; for example, the activity \textit{Request Created} is always performed at the beginning of the process, therefore its influence is towards increasing the remaining time, while explanation \emph{Role=Back-office} is a role associated to resources performing back-office activities, which are generally executed in the final part of the process, thus reducing the remaining time.

\begin{figure}[t!]
  \centering
    \includegraphics[width=\columnwidth]{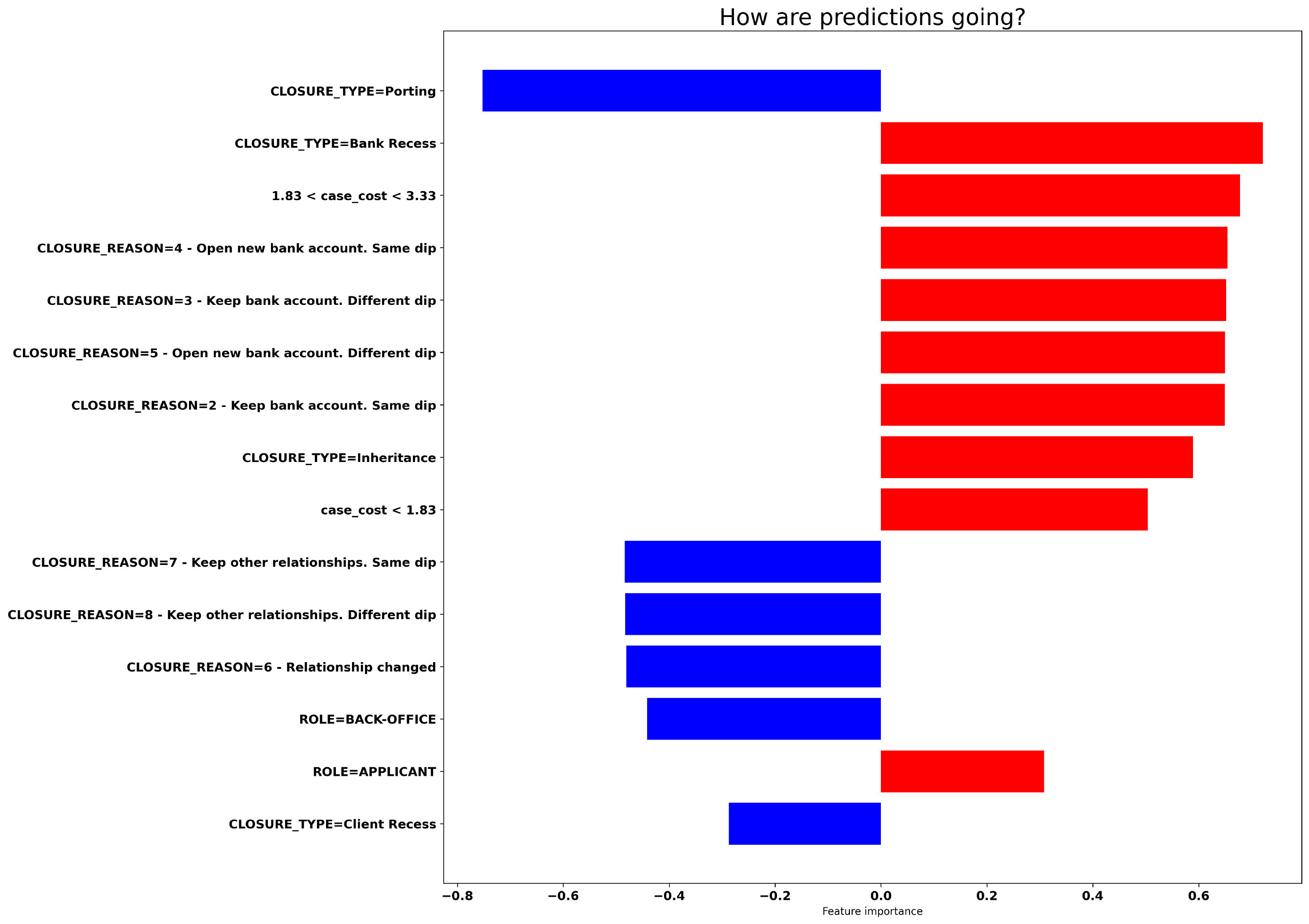}
\caption{Offline explanations for \emph{Authorization Requested} prediction (Bank Account Closure)}
  \label{fig:explanation_bac_auth_req}
\end{figure}

We mentioned that the financial institute aims to avoid activities related to inefficiencies (e.g. rework), such as \textit{Authorization Requested}, \textit{Pending Request for Acquittance of Heirs} and \textit{Back-Office Adjustment Requested}.
The explanations related to the first activity are shown in Figure \ref{fig:explanation_bac_auth_req}.
It can be easily seen that the attributes related to the Closure\_type and Closure\_reason are largely influencing the prediction. 
In particular, when it is the bank that decides to close the bank account (explanation \textit{Closure\_type=Bank Recess}), there is a high probability (0.70) that the director's approval will be necessary; this is also indicated by the red color of the bar, which indicates that the influence is towards predicting that the activity will be performed, and it is confirmed in the data: when the Closure\_type is Bank Recess, then the authorization is requested in the 71\% of cases (5117 out of 7174).
Conversely, when \textit{Closure\_type=Porting}, it is highly unlikely that a director's authorization to proceed further is needed, as it is indicated by the blue bar associated with a very high negative probability (-0.80).

Other important attributes are related to the reason triggering the closure's request (indicated by the attribute \textit{Closure\_reason}).
When there is the need to close the old bank account and open a new one in the same department (\textit{Closure\_reason=4 - Open new bank account. Same dip}) or when there is the need to keep the same bank account, but transfer the control of it to another department (\textit{Closure\_reason=3 - Keep bank account. Different dip}), then there is a high probability that an authorization from the director is needed; these are, in fact, exceptional situations, which need to be managed carefully. A further analysis of the data confirmed that the authorization is needed in the 75\% of cases (383 out of 504 cases with the first closure closure) for the first closure reason, and in the 84\% of cases (177 out of 209 cases) for the second closure reason. 

Finally, when cost of the case is very low, then the probability to predict that the director's authorization will be needed increases considerably (as it is shown in the explanations $1.83<case\_cost<3.33$ and $case\_cost<1.83$); this is due to the fact that the authorization to proceed further is usually requested to a director in the early stages of the process, when the cost of the case is still very low. 

In order to confirm the findings reported here, we conducted additional experiments with different KPIs also on publicly-available event logs. The results are discussed in~\ref{appendix:additional_experiments}.

\section{Evaluation of the Human Understandability of the Predictions and Explanations}
\label{sec:user_evaluation}

Section~\ref{sec:experiments} has shown an evaluation of the quality of predictions and explanations for the framework described in Section~\ref{sec:Explainable}. However, it needs to be ultimately used by process analysts, for whom it must be effective and comprehensible. To this aim, we carried out an empirical user evaluation of the explanations produced by our explainable predictive monitoring framework.
In particular, for the user evaluation we involved real and potential process analysts, in order to assess whether or not they would understand the explanations and, consequently, trust the whole  explainable predictive monitoring framework.
Our framework has been integrated in the IBM Process Mining suite\footnote{https://www.ibm.com/it-it/cloud/cloud-pak-for-business-automation/process-mining}.
This enabled us to perform a complete user evaluation: we did not only analyze the results produced by our framework in isolation, but we also considered the context in which these results were produced.

Section~\ref{sec:integration} describes the integration of the explainable predictive framework inside the IBM Process Mining suite, while Section~\ref{sec:evaluation_methodology} reports on the methodology carried out for the user evaluation.
Finally, Section~\ref{sec:evaluation_results} discusses the feedback obtained by the process analysts who partecipated in the user evaluation.

\subsection{A Software System for Explainable Predictive Process Analytics}
\label{sec:integration}

This section describes the integration of the explainable predictive monitoring framework discussed in Section~\ref{sec:Explainable} within the IBM Process Mining suite. 
This enables us to provide process stakeholders with a ready-to-use module that provisions online operational support for their processes, as well as the influencers driving them, without requiring any specific technical knowledge.  

\begin{figure}[t!]
    \includegraphics[width=1 \columnwidth]{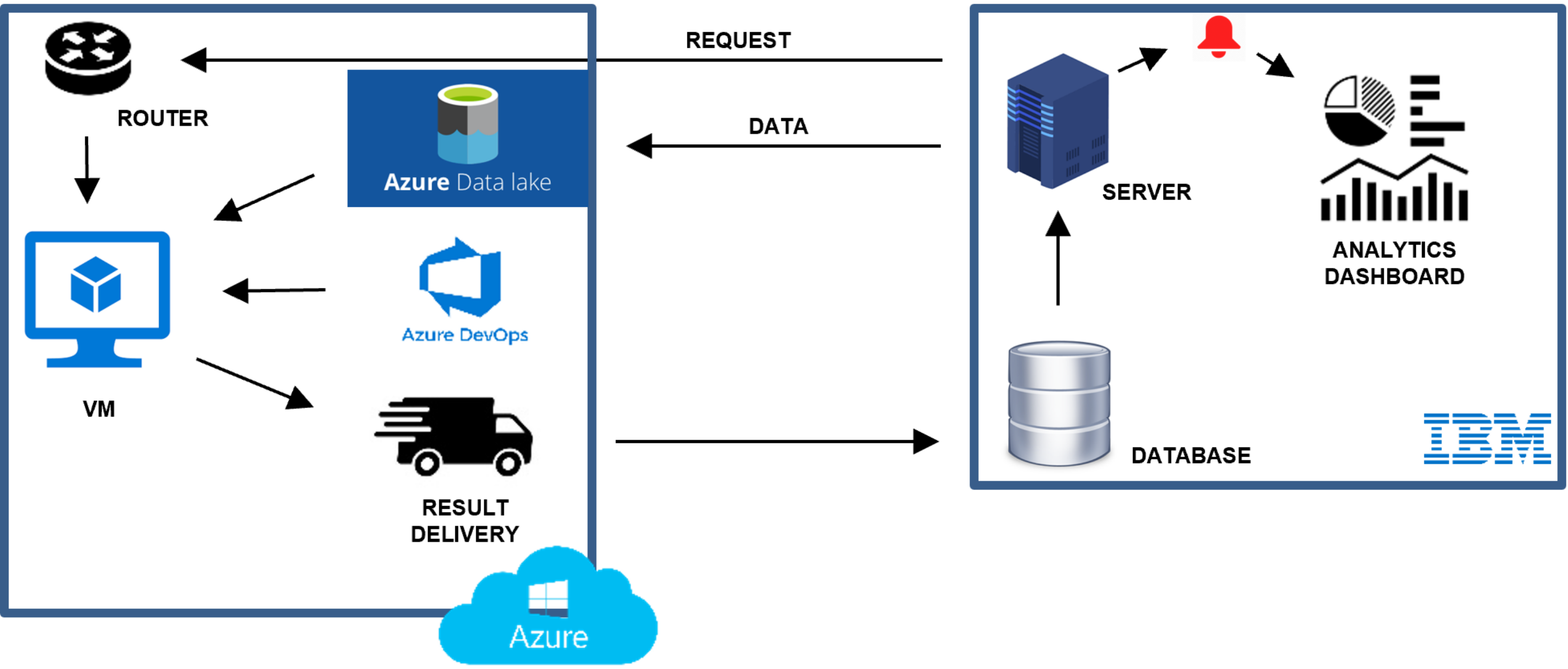}
    \caption{High level overview of the main components involved in the explainable decision support system}
    \label{fig:decision_support_system}
\end{figure}

The back-end of the predictive monitoring is based on the Azure infrastructure, which enables us to deploy the technique in the cloud and develop a whole system around it.
Figure~\ref{fig:decision_support_system} provides an high level overview of the main components involved in the explainable decision support system.
The IBM Process Mining Suite is represented on the right side, while the Azure infrastracture (also called Machine Learning Plarform) hosting our framework is shown on the left.
The latter is in charge of processing the requests coming from the IBM Process Mining suite, preparing a compute instance to execute our framework, and delivering the results back. 
In particular, the Machine Learning Platform has been tested to work with datasets up to 10 million events, it can handle multiple requests coming from different users and, in case a customer requests it, multiple compute instances can be easily provided by allocating new clusters, enabling to scale on demand.

The router is the component responsible to process the incoming requests and verify if the domain and the user related to the request are authorized; if the check is positive, the Process Mining Suite uploads the event log on the Azure Datalake, which is the component responsible for storing and archiving the data. 
Once the data has been fully uploaded, the router forwards the request to another component (here shown as VM), which is responsible to create a virtual machine with the necessary computational power and copy the event log data inside the virtual machine; at this point, the code of our explainable predictive framework (which is located in a repository inside Azure DevOps component) is downloaded inside the virtual machine, where it will be executed.
The output produced by our explainable predictive framework executed in the VM component will be stored inside the Azure Datalake and then sent to the database of the Process Mining suite via the Result Delivery component. Finally, the results will be shown in the Analytics dashboard via the Application server.

\begin{figure*}[ht!]
    \centering
    \includegraphics[width=\textwidth]{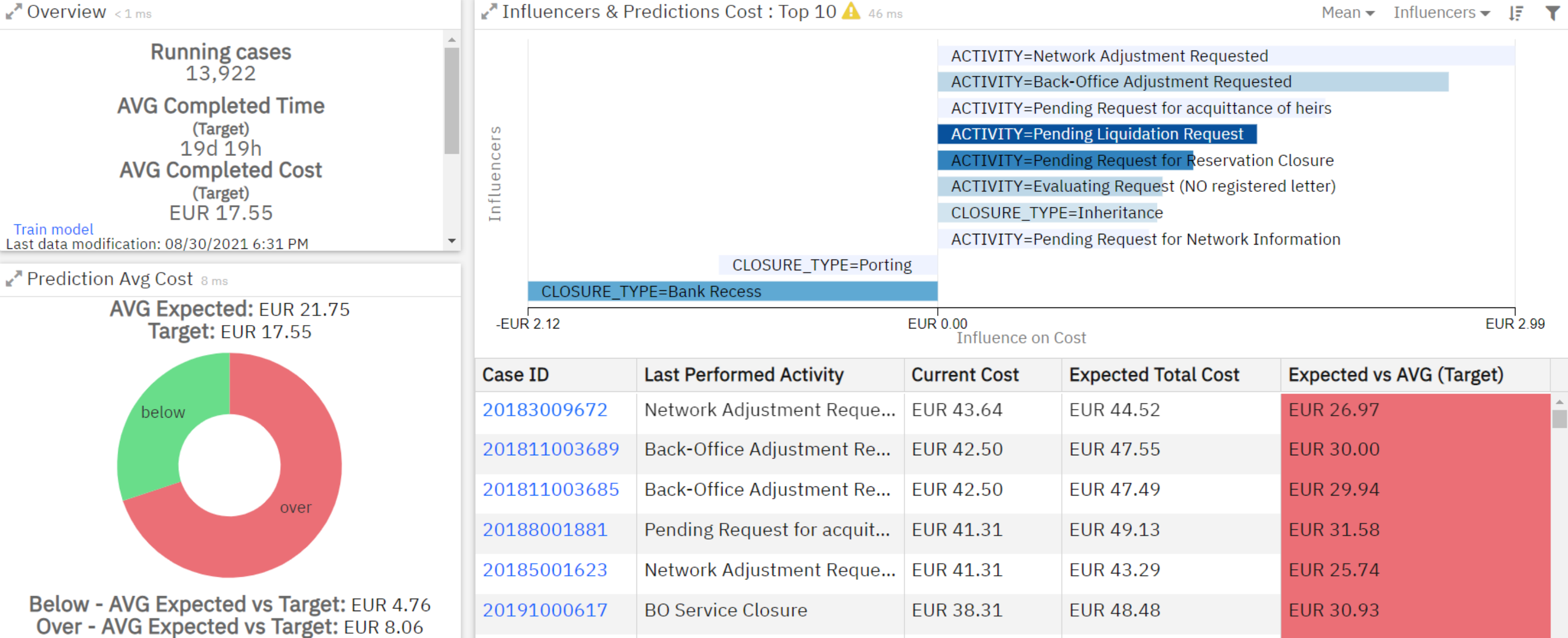}
    \caption{The IBM Analytics dashboard for Explainable Predictive Process Monitoring.}
    \label{fig:ibm_dashboard}
\end{figure*}

Figure~\ref{fig:ibm_dashboard} shows a screenshot of the \textit{Analytics Dashboard} within the IBM Process Mining suite for prediction of the total cost of cases, namely the cost necessary to complete a case. The use case presented here is related to the Bank Account Closure process process, which deals with the closure of customer's accounts, which may be requested by the customer or by the bank, for several reasons. 

\begin{figure}
    \centering
    \includegraphics[width=1\columnwidth]{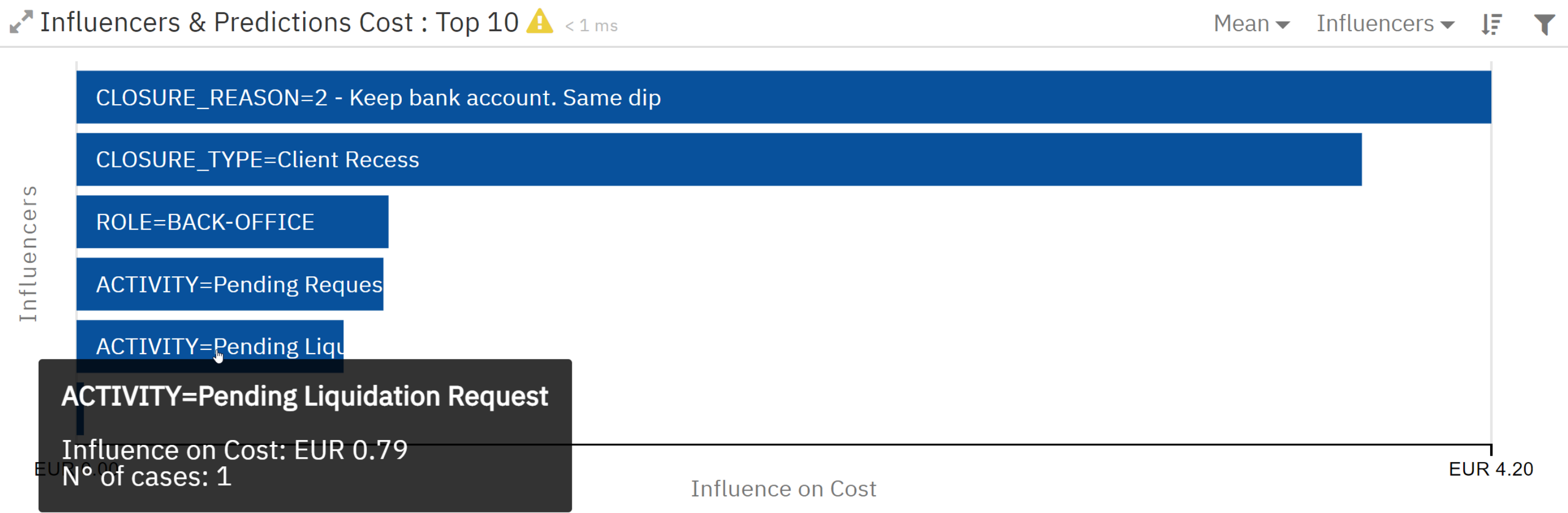}
    \caption{Explanations related to one running case. An explanation is visible by passing over with the mouse when a corresponding bar is too short.}
    \label{fig:one_case}
\end{figure}

The upper-left corner reports on general process statistics, such as the number of running cases and the average case total time (here labeled as \textit{Completed Time}) and cost. 
In the bottom-left corner, the widget shows how many running cases are predicted to cost too much (indicated by the red portion of the pie chart) and what is the average foreseen overcost for those cases.
The bottom-right corner lists the running cases, each associated with the case identifier and the last performed activity; since this dashboard refers to the process total cost, each case is also associated with the current cost, the expected total cost as forecasted by the predictive monitor, and its difference wrt.\ the average completion cost, here also named as target. 
When a user clicks on a specific running case (e.g.\ with id 201811010127), it is possible to see the explanations for that case, named \textit{influencers} in the tool (see Figure~\ref{fig:one_case}).
Let us recall that when \textit{f} is categorical, each explanation is of form \textit{f=x}
and is associated with a \textit{Shapley value} $\psi$ and that the explanation's
interpretation is as follows: since feature $f$ takes on value $x$ for this running case,
the KPI prediction deviates $\psi$ units from the average KPI observed in the executions recorded in the event-log.
Conversely, when \textit{f} is numerical, each explanation takes one form between $f<w_1$, $w_1 \leq f <w_2$, $\ldots$, $f \geq w_q$.
Let us consider the first form. The interpretation is as follows: since feature $f$ takes on a value minor than $w_1$ for this running case, the KPI prediction deviates $\psi$ units from the average KPI observed in the executions recorded in the event-log.

Let us consider again the all-cases dashboard in Figure~\ref{fig:ibm_dashboard}: the bar chart in the top-right corner provides an helicopter view of the explanations. 
In particular, each row of the bar chart represents an explanation, and extends towards left or right, depending whether the average Shapley value for the explanation is negative or positive. The colour indicates the frequency of an explanation, with darker colours indicating a large number of running cases with that explanation. 

As an example, explanation \textit{ACTIVITY=Network Adjustment Requested} has a large bar with a light colour: this means that, for a small number of cases, the fact that the latest activity has been a \emph{Network Adjustment Requested} has contributed to reduce the predicted total case cost by an average value of 2.99 Euros. The explanation \emph{CLOSURE\_TYPE=Bank Recess} is conversely associated with a darker colour, namely with a large number of cases. The average shapley value is equal to -2.12 Euros: when the closure of the bank account is requested directly by the bank, the total cost reduces by 2.12 Euros wrt.\ the average.

\subsection{Methodology of the evaluation}
\label{sec:evaluation_methodology}
A user study was run to test the usability of the Explainable Predictive Analytics module and the users’ experience. 
Section~\ref{sec:participants} reports on the characteristics of the users that participated in the experimental session, while Section~\ref{sec:materials_and_methods} describes the experimental setting and the questionnaires that have been adopted to carry out the evaluation session.

\subsubsection{Participants}
\label{sec:participants}
Twenty users were involved in the case study, equally split between males and females, with an average age of 31.20 and standard deviation (SD) of 8.43. On average, they had an educational level (years of school) of 18.15 years, with a standard deviation of 1.42. Out of the 20 users, five were Master’s or PhD students in Data Science or a related field, 13 were employees in an international IT services company, and two were both students and employees of the same company. Eleven participants were Italian, while nine were not. Experiments were conducted in Italian or English according to the favourite language of each subject. 

All participants had some previous knowledge about Process Mining: students had previously taken a course on Process Mining, while employees had already applied Process Mining in their analysis work. 
Indeed, we asked them about self-evaluating their process mining knowledge on a 5-point Likert Scale (with 1 = superficial knowledge, 2 = basic knowledge, 3 = medium knowledge, 4 = good knowledge, 5 = advanced knowledge), reporting an average score of 3.5 (SD = 1.36).
Eighteen out of 20 users reported a previous experience with process mining tools, with an average use frequency of 3.39 and SD of 1.14 (the scale points were 1 = almost never, 2 = rarely, 3 = sometimes, 4 = often, 5 = always). 
However, none of them had previous experience with the specific module being analyzed.

\subsubsection{Materials and Methods}
\label{sec:materials_and_methods}
The user study was designed in accordance with the Declaration of Helsinki and approved by the ethics committee for psychological research at the University of Padova (protocol number 4259).
All participants were required to read and provide informed consent before starting the experiment.
After providing demographic data and some information related to their previous knowledge of process mining, users were presented with an 8-minute video explaining the main structure and functionalities of the Explainable Predictive Analytics module. Then, participants were asked to complete 18 tasks within the module. The tasks were aimed at testing the intelligibility of the information provided by the Explainable Predictive Analytics module regarding the process related to the closure of the customers’ bank accounts. Example of tasks are: “According to the prediction of the algorithm, how many cases will cost more than the average?”, “What is the most frequent influencer affecting cost prediction?”. 
The complete list of the tasks is reported in Table~\ref{tab:annex_1}.

\begin{table*}[t!]
\caption{List of the tasks completed by the users within the Explainable Predictive Analytics module. The third and fourth columns report respectively the average (and Standard Deviation) accuracy obtained by participants in completing the task and the evaluation of the task difficulty according to the 1-item questionnaire by Tedesco and Tullis~\cite{tedesco2006}}.
\footnotesize
\centering
\resizebox{\textwidth}{!}
{\begin{tabular}{|l|l|c|c|}
\hline
\cellcolor{LightCyan}\textbf{N.} &
\cellcolor{LightCyan}\textbf{Task} &
\cellcolor{LightCyan}{\begin{tabular}[c]{@{}c@{}}\textbf{Accuracy}\\\textbf{(Average, SD)}\end{tabular}} &
\cellcolor{LightCyan}{\begin{tabular}[c]{@{}c@{}}\textbf{Difficulty}\\\textbf{(Average, SD)}\end{tabular}} 
\\
\hline
 \textbf{1} & 
 In the Overview widget, what is the meaning of AVG Completed Cost? & 
 0.83 (0.37) & 
 2.10 (0.79) \\
\hline
 \cellcolor{LightGray}\textbf{2}  &
 \cellcolor{LightGray}According to the prediction of the algorithm, how many cases will cost more than the average? &
 \cellcolor{LightGray}0.95 (0.22) & 
 \cellcolor{LightGray}2.25 (0.85) \\
\hline
 \textbf{3} & 
 Focus now on case 20183009672. What is the current case cost? & 
 1.00 (0.00) &
 1.60 (0.68) \\ 
\hline
\cellcolor{LightGray}\textbf{4}  &
 \cellcolor{LightGray}Always consider the case 20183009672. According to the algorithm, what will be the total cost of the case?  &
 \cellcolor{LightGray}1.00 (0.00) & 
 \cellcolor{LightGray}1.80 (0.77) \\ \hline
 \textbf{5} & 
 In “Influencers \& Predictions Cost” widget, what does the column “Expected vs AVG (Target)” represent? & 
 0.70 (0.41) & 
 2.75 (1.02) \\ 
\hline
\cellcolor{LightGray}\textbf{6}  & 
\cellcolor{LightGray}\begin{tabular}{@{}l@{}}Click on the practice 20183009672 and write below which is the influencer, relative to the costs, most significant\\ for this case.  \end{tabular}& 
\cellcolor{LightGray}0.95 (0.22) &
\cellcolor{LightGray}2.70 (0.57)\\  
\hline
\textbf{7} & 
\begin{tabular}{@{}l@{}}Still in relation to the 20183009672 case, does the most significant influencer raise or lower the prediction of the \\cost of the case? \end{tabular}& 
 0.90 (0.31) & 
 2.25 (0.79) \\ 
\hline
\cellcolor{LightGray}\textbf{8}  &
\cellcolor{LightGray}\begin{tabular}{@{}l@{}}Still in relation to the 20183009672 case, how much does the most significant influencer raise the prediction of \\the cost of the case?\end{tabular} &
 \cellcolor{LightGray}0.90 (0.31) & 
 \cellcolor{LightGray}2.10 (0.97)\\ 
\hline
 \textbf{9} & 
 How much, on average, is the cost of the process influenced by the Closure type “Bank Recess”? & 
 0.88 (0.28) & 
 2.60 (0.88) \\ 
\hline
\cellcolor{LightGray}\textbf{10}  &
 \cellcolor{LightGray}How many cases are influenced by the "Bank Recess" influencer? &
 \cellcolor{LightGray}1.00 (0.00) & 
 \cellcolor{LightGray}1.95 (1.00) \\ 
\hline
\textbf{11} & 
\begin{tabular}{@{}l@{}}Does the fact that a Network Adjustment Requested (an activity that consists of reworking the file following an \\error) has been carried out in the process influence the prediction of the total cost?\end{tabular} &
 0.95 (0.22) & 
 2.10 (0.72) \\ 
\hline
 \cellcolor{LightGray}\textbf{12}  &
 \cellcolor{LightGray}How much is the total cost prediction affected by a "Network Adjustment Requested"? &
 \cellcolor{LightGray}0.85 (0.37) & 
 \cellcolor{LightGray}1.85 (0.88) \\
\hline
 \textbf{13} & 
 What is the most frequent influencer affecting cost prediction? & 
 0.70 (0.47) & 
 2.45 (1.00) \\ 
\hline
\cellcolor{LightGray}\textbf{14}  & 
\cellcolor{LightGray}\begin{tabular}{@{}l@{}}Order Influencers bar chart for median influence instead of mean influence. Look at the graph sorted by the median.\\ What is the explanation that most influences cost prediction? \end{tabular} & 
\cellcolor{LightGray}0.90 (0.31) & 
\cellcolor{LightGray}2.80 (1.15) \\  
\hline
\textbf{15} & 
 \begin{tabular}{@{}l@{}}Order Influencers bar chart again for mean influence. What information provides the ACTIVITY = Back-Office \\Adjustment Requested histogram? \end{tabular}& 
 0.75 (0.30) & 
 2.50 (0.76) \\ 
\hline
\cellcolor{LightGray}\textbf{16}  &
\cellcolor{LightGray}\begin{tabular}{@{}l@{}} In the "Influencers \& Predictions Cost" widget, filter for the influencer relating to the CLOSURE\_TYPE Bank Recess.\\ What are the two influencers that best explain cost prediction when the CLOSURE\_TYPE is Bank Recess?\end{tabular} &
 \cellcolor{LightGray}0.80 (0.38) & 
 \cellcolor{LightGray}3.10 (0.91) \\
\hline
\textbf{17} & 
 \begin{tabular}{@{}l@{}}View influencers related to case 20183009672. Focus on the influencers related to the case 20183009672. Explain \\what information the first histogram provides (Current Cost $>$ 18.82). \end{tabular}& 
 0.58 (0.37) & 
 3.55 (1.15) \\
 \hline
 \cellcolor{LightGray}\textbf{18}  &
 \cellcolor{LightGray}\begin{tabular}{@{}l@{}}Change the settings, setting the display of "Completed cases". In this new view of completed cases, who is the most\\ frequent influencer? \end{tabular}&
 \cellcolor{LightGray}0.80 (0.41) & 
 \cellcolor{LightGray}2.55 (1.10) \\
\hline
\end{tabular}}
\label{tab:annex_1}
\end{table*}

After each task, participants were asked about the difficulty of the task, which was measured according to the 1-item questionnaire proposed by Tedesco and Tullis (“Overall this task was”, 1 = very easy, 2 = easy, 3 = neither easy nor difficult, 4 = difficult, 5 = very difficult).
Previous literature reports this questionnaire to be the most reliable on small subject's sample in terms of its correlation with performance measures, if compared with other questionnaires on task difficulties~\cite{tedesco2006}.
Finally, at the end of the evaluation session, the users filled out two questionnaires for the assessment of usability and user experience~\cite{salford60928,Sauro2016}:

\begin{itemize}
\item Post-Study System Usability Questionnaire (PSSUQ)~\cite{James1992}: it is a 19-items questionnaire that assesses user satisfaction with system usability. In addition to the overall satisfaction score, the PSSUQ provides sub-scores for three specific dimensions: System Usefulness (SysUse), Information Quality (InfoQual), Interface Quality (InterQual). Responses are given on a 7-point Likert Scale from 1 = strongly agree to 7 = strongly disagree.  

\item User Experience Questionnaire (UEQ)~\cite{10.1007/978-3-540-89350-9_6}: it includes 26 items testing the users’ experience with the product they interacted with. Items are made up of pairs of opposite adjectives (e.g., “impractical – practical”, “boring – exciting”), with a response scale on seven points. The UEQ measures both pragmatic and hedonic components of the user experience, providing a score for six different dimensions: Attractiveness (overall impression of the product: do users like or dislike the product?), Perspicuity (is it easy to get familiar with the product? Is it easy to learn how to use the product?), Efficiency (can users solve their tasks without unnecessary effort?), Dependability (does the user feel in control of the interaction), Stimulation (is it exciting and motivating to use the product?), Novelty (is the product innovative and creative? Does the product catch the interest of users?)~\cite{10.1007/978-3-319-07668-3_37}. 
\end{itemize}

The literature proposes normative data for these questionnaires~\cite{doi:10.1080/10447318.2002.9669130}. This has allowed us to derive sound conclusions on the goodness of the numerical results obtained. For PSSUQ, we considered satisfactory the scores that are in or above the normative range reported by Lewis~\cite{doi:10.1080/10447318.2002.9669130}: the mean of the overall satisfaction rating is 2.82 with a 99\% confidence interval (CI) ranging from 2.62 to 3.02. More specifically going into the three dimensions, the means for the subscales are 2.80
(99\% CI 2.57–3.02) for System Usefulness, 3.02 (99\% CI 2.79–3.24) for Information Quality, and 2.49 (99\% CI 2.28–2.71) for Interface Quality.

For UEQ, an interpretation of the level of satisfaction of the outcome can be achieved by comparing the obtained scores with those of a benchmark data set~\cite{Schrepp_ueq_questionnaire}, which contains data from 20190 users from 452 studies concerning different products (business software, web pages, web shops, social networks). 
In particular, the comparison with the benchmark dataset allows us to qualitatively classify the product for each analyzed dimension as: 

\begin{itemize}
\item Excellent: the score obtained for the evaluated product is in the range of the 10\% best results.
\item Good: 10\% of the products in the benchmark data set have a better score, while 75\% of the products are worse.
\item Above average: 25\% of the products in the benchmark have a better score than the score obtained for the evaluated product, while 50\% of the products are worse.
\item Below average: 50\% of the products in the benchmark have a better score than the score obtained for the evaluated product, while 25\% of the products are worse.
\item Bad: the score obtained for the evaluated product is in the range of the 25\% worst results.
\end{itemize}

\subsection{Evaluation results}
\label{sec:evaluation_results}

This section discusses the results of the usability study, which has focused on four dimensions: accuracy to carry out the tasks, perceived task difficulty, usability of the module and user experience.

\subsubsection{Task Accuracy}
\label{sec:task-accuracy}

The task accuracy was calculated as the percentage of tasks correctly fulfilled by the users. For each task, one point was assigned if the task was completed correctly, 0 when the user failed, and 0.5 points if the user’s response was partially correct; as an example, 0.5 points were assigned when the user had to find the two main influencers for a particular case, but only one was indicated.

On average, users obtain an accuracy of 0.86 (with SD of 0.11), with the 17 being the most failed task (0.58).
The accuracy for each task is reported in the third column of Table~\ref{tab:annex_1}. 
This level of accuracy can be reasonably considered good; since most of the tasks requires one to leverage on the prediction's explanations, we can conclude that \textbf{explanations are generally comprehensible to correctly carry out analysis' tasks}. One should also take into account that users were confronted for the first time with the idea of explainable process predictive analytics and with its operationalization in the IBM Process Mining software suite. 

As shown in Table~\ref{tab:annex_1}, tasks 5, 13, 17 have the lowest scores, respectively 0.70, 0.70, 0.58.

Task 5 investigated whether the users could understand the meaning of the column \textit{Expected vs AVG (Target)}, i.e.\ the last column of the Table in Figure~\ref{fig:ibm_dashboard}. In particular, the value for a certain process instance (i.e., case) shows the difference between the predicted value for that process instance (shown in the same Table under the column \textit{Expected Total Cost}) and the actual observed average cost (which is shown in the Overview widget in the top-left corner).
The low accuracy for task 5 is likely related to the fact that this value is calculated using the information contained in two different widgets. 
We plan to show a tooltip explaining the meaning of this column. However, it is worthwhile observing that this task is not related to using the explanations, but is rather related to the visualization of the prediction.

Conversely, task 13 focuses on whether or not the users could correctly indicate the most frequent explanation affecting the cost prediction. 
An accurate answer to this task should be obtained by inspecting the colours of the bars of the different explanations and focusing on those with darker colours, which mean greater frequencies. Note that the meaning of coloring is consistent with other modules of the  IBM Process Mining suite (e.g., in a process model, darker colours are given to the activities that occur more frequently). Indeed, breaking the users down in two groups, composed respectively by users familiar and unfamiliar with other parts of the IBM Process Mining suite, we observed that task 13 was generally answered incorrectly by subjects that never used IBM Process Mining before (among the 6 users that answered incorrectly, 5 users out of 6 never used IBM Process Mining before).

Task 17 had the lowest accuracy, whose perceived difficulty was certainly the highest (cf.\ last column of Table~\ref{tab:annex_1}). This was reported in some of the user's feedback: "\textit{It was not clear the meaning of the influencer related to the Current Cost, if this is the cost then the influence on the cost is...}". 
The Current Cost is an attribute that encodes the cost of a case after each event occurred in a process; it is an attribute that is not part of the original event log, but it has been added in order to increase the predictive accuracy of the model.

To address the issue with this type of task, we plan to show in separated views explanations related to process attributes already present in the original event log and explanations related to encoded attributes not present in the original event log (such as the cost). This also came as a comment from one of the users in the feedback: "\textit{For me it was difficult to understand the relations between the influencers related to process attributes and those related to the cost; I would rather prefer to analyze them separately}"). It is also rather important to facilitate users to interpret the semantics of the encoded attributes: a tooltip could then be useful to explain their semantics. 

\subsubsection{Task Difficulty}
Last column of Table~\ref{tab:annex_1} shows the perceived task difficulty for each of the tasks. The average task difficulty is 2.39, with standard deviation of 0.41. 
This means that most of the responses given by participants fall in the range between "easy" and "neither easy nor difficult", leading us to conclude that \textbf{users have found most of the tasks reasonably easy}.

\begin{figure}[t!]
    \centering
    \includegraphics[width=1 \columnwidth]{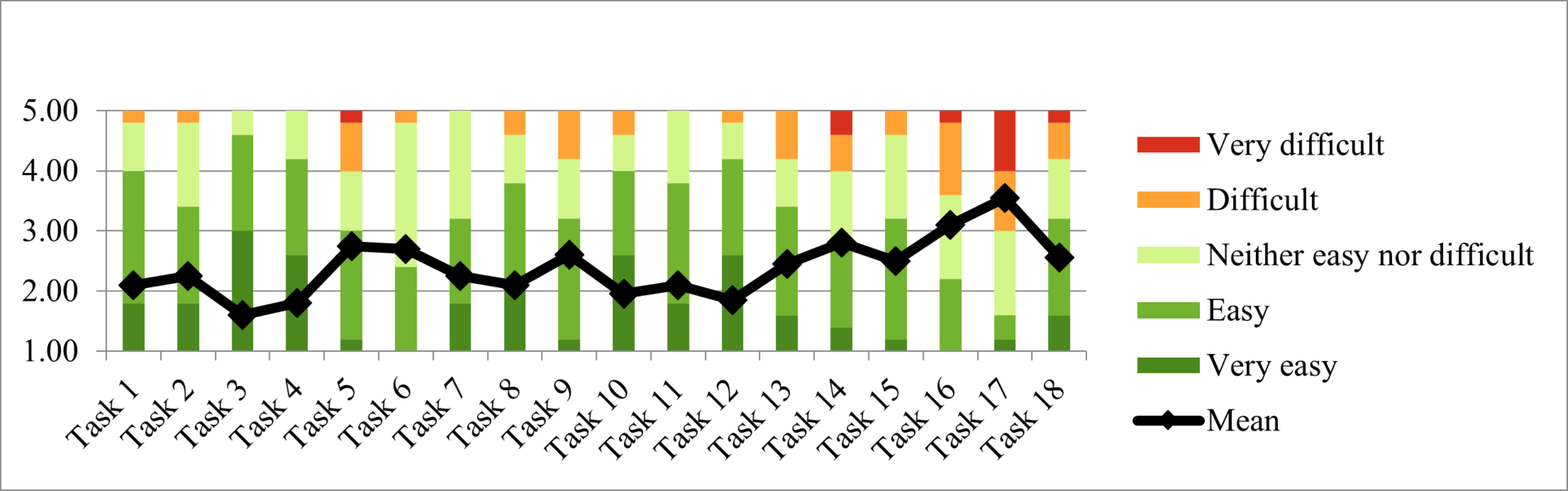}
    \caption{Distribution of the responses in the 1-item questionnaire on task difficulty for each task. 
    For each task, the height of the several boxplots varies depending on the percentage of users reporting the corresponding task difficulty.}
    \label{fig:responses_distribution}
\end{figure}

We also illustrate a graphical distribution of the responses for each task in Figure~\ref{fig:responses_distribution}. 
Consistently with what was observed in the accuracy assessment described in Section~\ref{sec:task-accuracy}, the task perceived as the most difficult was the task 17.

\subsubsection{Usability}

\begin{figure}[t!]
    \centering
    \includegraphics[width=1 \columnwidth]{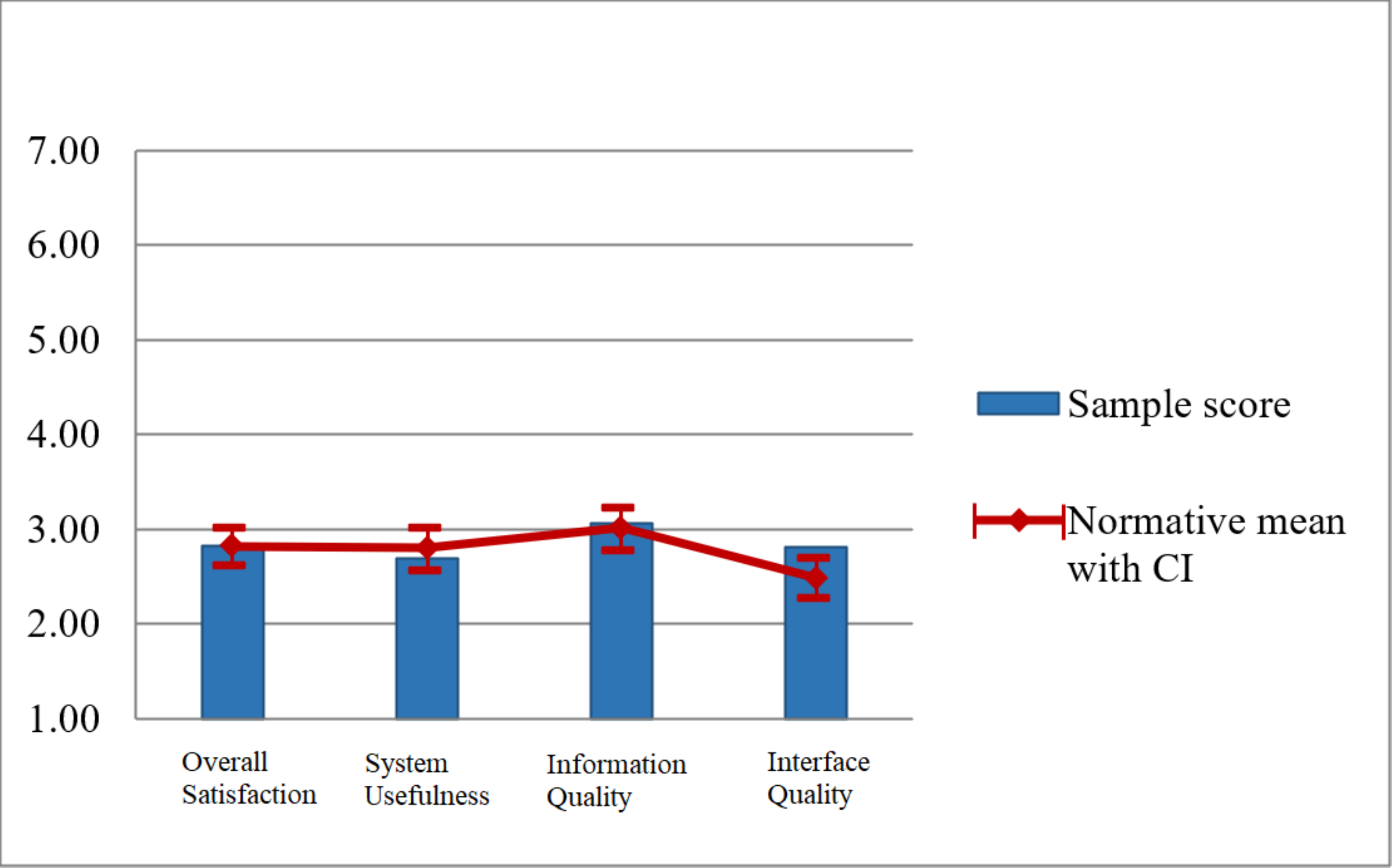}
    \caption{Average scores in PSSUQ subscales and their collocation compared to the normative range (mean; CI). Responses are given on a 7-point Likert Scale from 1 = strongly agree to 7 = strongly disagree. Ideally, the scores should be the lowest possible.}
    \label{fig:average_pssuq_scores}
\end{figure}

The analysis of the results of the Post-Study System Usability Questionnaire (PSSUQ) is summarized in 
Figure~\ref{fig:average_pssuq_scores}: the bars are the values obtained in our evaluation, while the depicted ranges show the normative values with confidence interval (cf.\  Section~\ref{sec:materials_and_methods}).
The overall satisfaction score of 2.83 (with SD of 1.24) is within the normative range, thus testifying a good level of user satisfaction of the interface.
In the three subscales, participants obtain the following scores: System Usefulness = 2.70 (SD = 1.23), Information Quality = 3.07 (SD = 1.61), Interface Quality = 2.82 (SD = 1.38); these scores fall within the normative range as well.

In other words, \textbf{no critical issues emerged regarding the system usefulness, the quality of the information provided by the system, and the quality of the interface (i.e. no major issues regarding the intelligibility of the explanations were found).}

\subsubsection{User Experience}

\begin{figure}[t!]
    \centering
    \includegraphics[width=1 \columnwidth]{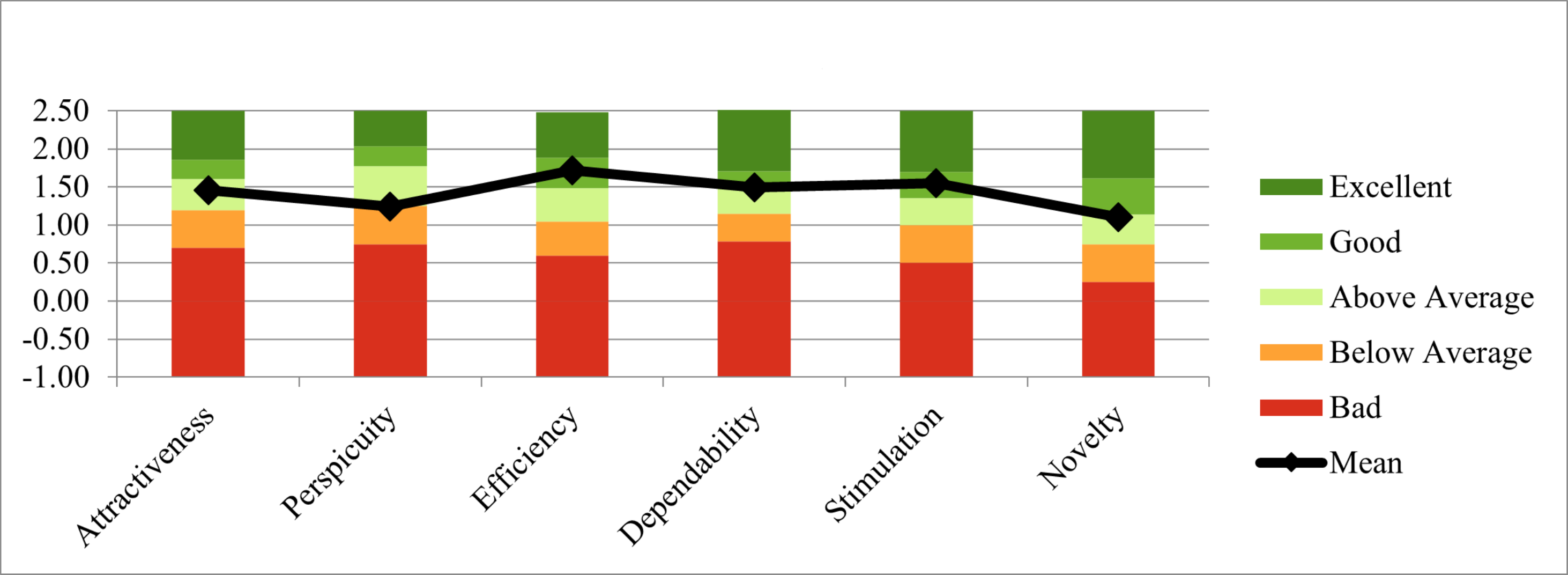}
    \caption{Average scores obtained by the users in the UEQ subscales (represented by the black line), and their collocation with respect to other products of the benchmark data set (represented by the coloured blocks).
    Note that “good” means that 75\% of products of the benchmark data set perform worse; “above average” means that 50\% of products of the benchmark data set perform worse; “below average” indicates that 25\% of the products of the benchmark data set perform worse.}
    \label{fig:average_scores_ueq}
\end{figure}

As concerns the user experience (UX), the six dimensions measured by the UEQ have obtained the following scores: Attractiveness = 1.45 (SD = 1.06), Perspicuity = 1.24 (SD = 1.20), Efficiency = 1.71 (SD = 1.16), Dependability = 1.49 (SD = 0.94), Stimulation = 1.55 (SD = 0.99), Novelty = 1.10 (SD = 1.26). Comparing these scores with those of the benchmark data set (cf. Section~\ref{sec:materials_and_methods}), it should be noted that the Explainable Predictive Analytics module is evaluated as good for Efficiency, Dependability, and Stimulation (see Figure~\ref{fig:average_scores_ueq}). It is above average compared to other products for Attractiveness and Novelty. Instead, Perspicuity is just below average; this could be due to the fact that most of the users were not particularly familiar with the adopted process mining tool. Therefore, in order to improve the Perspicuity of the Explainable Predictive Analytics module, it could be important in the future to plan a training phase beforehand, in which users get familiar with the process mining tool.  
In~\ref{appendix:response_distribution}, the distribution of the responses for each item of the questionnaire is reported.

Summarizing, we conducted a rigorous user evaluation to understand if process analysts (both from the academy and the industry domains) were comfortable with the results returned by our explainable predictive monitoring framework. Several points emerged from our study; we summarize the main findings here.

\subsection{Concluding Remarks}
The user study has showed that overall the users achieved a good score and could find the answer for most of the questions without particular problems: cf.\ an average accuracy of 0.86 out of 1, and an average, perceived task difficulty of 2.39 out of 5. Considering that most of the tasks focus on the explanations, this shows that \textbf{predictions are actually explained in a form that is effective and efficient for process analysts}.
However, the lower accuracy observed in some tasks has highlighted points for improvement, but these improvements are not related to prediction's explanations, except for task 17 where users did not well understand explanations that relate to features that are not directly mapped to process' attributes. However, the problem could likely be overcome by adding a tooltip that indicates the semantics of these additional features.

The Post-Study Usability Questionnaire pointed out that the usability of the explainable process predictive analytics framework and of its operationalization was good, beyond the minimum usability level that literature indicates as suitable. This confirmed  
that \textbf{the proposed explainable framework was considered intelligible by the users}. 

Last but not least, the User Experience Questionnaire also showed levels of user satisfaction with the framework that literature consider as appropriate: this indicates that \textbf{users were satisfied with the explainable process predictive framework}. A slight lower value, still within the boundaries, was observed about perspicuity: this is likely related to the fact that users were confronted for the first time with the framework, and a short training phase beforehand would be beneficial.

\section{Conclusions}
\label{sec:conclusions}

A lot of research has been devoted towards increasingly accurate frameworks for predictive process monitoring. Moreover, in the last years, growing attention has been paid to ensure that that the resulting predictive-monitoring system is workable in practice. With practical workability, here we intend that the process analysts and stakeholders need to trust the system and its predictions.
Previous studies have already shown that a necessary condition to build trust is to explain the reason of the provided predictions; proposals that do not put explanation as a core feature are not going to be adopted in practice. 

Several works have recently put forward methods to explain the predictions, using a plethora of different techniques (see Section~\ref{sec:Related_works}). However, as also stated by Stierle et al., none of these works verified whether prediction's explanations are provided to process analysts in a form that is intelligible and that can give them actionable insights easily, effectively and efficiently. Furthermore, most of the frameworks have stopped at a prototype phase and have not been deployed in real software suites. 

This paper attempts to address the issues above: assessing the quality of prediction explanations with users,
developing the framework into a commercial software. 
In particular, the paper contributions are the following: 

\begin{enumerate}
    \item Gradient boosting (Catboost) and LSTM models were compared on several real-life case studies to assess which model is most suitable for the task of process predictive analytics. In fact, they guarantee similar quality of predictions, but Catboost has shown a drastically reduced model's training time. Section~\ref{sec:experiments} provides an extensive discussion on this.
    \item The explainable process predictive framework has been implemented as a module of a commercial software, the IBM Process-Mining software suite  (cf.\ Section~\ref{sec:integration}).
        \item A user study has been conducted on the IBM Process-Mining suite to assess the efficiency and effectiveness of the proposed explainable process predictive analytics module. Results have shown that, indeed, the module and the form in which explanations are provided are efficient, effective, usable, and satisfactory from a final-user viewpoint. See Section~\ref{sec:user_evaluation}. 
\end{enumerate}
Note that these points are all novel contribution, compared with our previous work reported in~\cite{galanti2020explainable}. 
As a matter of fact, Rizzi et al.~\cite{rizzi_explainable_22} are the first that previously attempted to conduct some user studies on explainable process predictive analytics, but this attempt was limited to 8 subjects, and without using a commercial, fully-fledged operationalization. 

As future work, we aim to perform an assessment against the quality evaluation criteria for Explainable AI introduced in literature (see, e.g.\,\cite{10.1007/978-3-031-07481-3_7}). Furthermore, we aim to extend and adjust the explainable framework from predictive to prescriptive analytics. In the latter case, the framework needs to suggest which activities to perform as next in order to recover those cases that, otherwise, are predicted to not meet satisfactory KPI values.
This extension is not trivial, because explanations should concretely describe the process aspects that influenced the choice of the recommended activities.

\section*{Acknowledgement}
This work was supported by the Department of Mathematics, University of Padua, with HPC resources.
This research did not receive any specific grant from funding agencies in the public, commercial, or not-for-profit sectors.

\bibliography{references.bib}

\begin{thebibliography}{10}
\expandafter\ifx\csname url\endcsname\relax
  \def\url#1{\texttt{#1}}\fi
\expandafter\ifx\csname urlprefix\endcsname\relax\def\urlprefix{URL }\fi
\expandafter\ifx\csname href\endcsname\relax
  \def\href#1#2{#2} \def\path#1{#1}\fi

\bibitem{Marquez-Chamorro18}
A.~E. M{\'{a}}rquez{-}Chamorro, M.~Resinas, A.~Ruiz{-}Cort{\'{e}}s, Predictive
  monitoring of business processes: {A} survey, {IEEE} Transaction on Services
  Computing 11~(6) (2018) 962--977.

\bibitem{10.1007/s11257-017-9195-0}
I.~Nunes, D.~Jannach, A systematic review and taxonomy of explanations in
  decision support and recommender systems, User Modeling and User-Adapted
  Interaction 27~(3–5) (2017) 393–444.

\bibitem{doshivelez2017rigorous}
F.~Doshi-Velez, B.~Kim, Towards a rigorous science of interpretable machine
  learning (2017).
\newblock \href {http://arxiv.org/abs/1702.08608} {\path{arXiv:1702.08608}}.

\bibitem{DBLP:conf/icsoc/SindhgattaOM20}
R.~Sindhgatta, C.~Ouyang, C.~Moreira,
  \href{https://doi.org/10.1007/978-3-030-65310-1\_31}{Exploring
  interpretability for predictive process analytics}, in: E.~Kafeza,
  B.~Benatallah, F.~Martinelli, H.~Hacid, A.~Bouguettaya, H.~Motahari (Eds.),
  Service-Oriented Computing - 18th International Conference, {ICSOC} 2020,
  Dubai, United Arab Emirates, December 14-17, 2020, Proceedings, Vol. 12571 of
  Lecture Notes in Computer Science, Springer, 2020, pp. 439--447.
\newblock \href {https://doi.org/10.1007/978-3-030-65310-1\_31}
  {\path{doi:10.1007/978-3-030-65310-1\_31}}.
\newline\urlprefix\url{https://doi.org/10.1007/978-3-030-65310-1\_31}

\bibitem{DBLP:conf/ecis/StierleBWZM021}
M.~Stierle, J.~Brunk, S.~Weinzierl, S.~Zilker, M.~Matzner, J.~Becker, Bringing
  light into the darkness - {A} systematic literature review on explainable
  predictive business process monitoring techniques, in: F.~Rowe, R.~E. Amrani,
  M.~Limayem, S.~Matook, C.~Rosenkranz, E.~A. Whitley, A.~E. Quammah (Eds.),
  28th European Conference on Information Systems - Liberty, Equality, and
  Fraternity in a Digitizing World , {ECIS} 2020, Marrakech, Morocco, June
  15-17, 2020, 2021.

\bibitem{DBLP:conf/sac/TerragniH19}
A.~Terragni, M.~Hassani, Optimizing customer journey using process mining and
  sequence-aware recommendation, in: Proceedings of the 34th {ACM/SIGAPP}
  Symposium on Applied Computing, {SAC} 2019,, {ACM}, 2019, pp. 57--65.

\bibitem{Park19}
G.~{Park}, M.~{Song}, Prediction-based resource allocation using lstm and
  minimum cost and maximum flow algorithm, in: International Conference on
  Process Mining (ICPM), 2019, pp. 121--128.

\bibitem{TaxVRD17}
N.~Tax, I.~Verenich, M.~La~Rosa, M.~Dumas, Predictive business process
  monitoring with {LSTM} neural networks, in: Proceedings of 29th International
  Conference on Advanced Information Systems Engineering (CAiSE 2017), 2017,
  pp. 477--492.

\bibitem{LSTM_time}
N.~Navarin, B.~Vincenzi, M.~Polato, A.~Sperduti, {LSTM} networks for data-aware
  remaining time prediction of business process instances, in: Proceedings of
  the {IEEE} Symposium Series on Computational Intelligence {(SSCI 2017)},
  2017.

\bibitem{Catboost}
A.~V. Dorogush, V.~Ershov, A.~Gulin,
  \href{http://learningsys.org/nips17/assets/papers/paper\_11.pdf}{Catboost:
  gradient boosting with categorical features support}, in: Proceedings of the
  Workshop on ML Systems at NIPS 2017, 2017.
\newline\urlprefix\url{http://learningsys.org/nips17/assets/papers/paper\_11.pdf}

\bibitem{lime}
M.~T. Ribeiro, S.~Singh, C.~Guestrin, ``why should {I} trust you'': Explaining
  the predictions of any classifier, in: Proceedings of the 22nd {ACM} {SIGKDD}
  International Conference on Knowledge Discovery and Data Mining, San
  Francisco, 2016, pp. 1135--1144.

\bibitem{shrikumar2017learning}
A.~Shrikumar, P.~Greenside, A.~Kundaje, Learning important features through
  propagating activation differences, in: Proceedings of the 34th International
  Conference on Machine Learning-Volume 70, JMLR. org, 2017, pp. 3145--3153.

\bibitem{shap}
S.~M. Lundberg, S.-I. Lee, A unified approach to interpreting model
  predictions, in: Advances in neural information processing systems, 2017, pp.
  4765--4774.

\bibitem{Shu:2019:DEF:3292500.3330935}
K.~Shu, L.~Cui, S.~Wang, D.~Lee, H.~Liu, Defend: Explainable fake news
  detection, in: International Conference on Knowledge Discovery \& Data
  Mining, SIGKDD, ACM, 2019, pp. 395--405.

\bibitem{lundberg2018explainable}
S.~M. Lundberg, B.~Nair, M.~S. Vavilala, M.~Horibe, M.~J. Eisses, T.~Adams,
  D.~E. Liston, D.~K.-W. Low, S.-F. Newman, J.~Kim, et~al., Explainable
  machine-learning predictions for the prevention of hypoxaemia during surgery,
  Nature biomedical engineering 2~(10) (2018) 749.

\bibitem{DBLP:conf/iclr/2015}
D.~Bahdanau, K.~Cho, Y.~Bengio, Neural machine translation by jointly learning
  to align and translate, in: The 3rd International Conference on Learning
  Representations, {ICLR} 2015, San Diego, CA, USA, May 7-9, 2015, Conference
  Track Proceedings, 2015.

\bibitem{Meacham2019Explainable}
S.~Meacham, G.~Isaac, D.~Nauck, B.~Virginas, Towards Explainable AI: Design and
  Development for Explanation of Machine Learning Predictions for a Patient
  Readmittance Medical Application, 2019, pp. 939--955.

\bibitem{gradients}
M.~Sundararajan, A.~Taly, Q.~Yan, Axiomatic attribution for deep networks, in:
  Proceedings of the 34th International Conference on Machine Learning-Volume
  70, JMLR. org, 2017, pp. 3319--3328.

\bibitem{verenich_19_white_box}
I.~Verenich, M.~Dumas, M.~La~Rosa, H.~Nguyen, Predicting process performance: A
  white‐box approach based on process models, Journal of Software: Evolution
  and Process 31 (2019) e2170.
\newblock \href {https://doi.org/10.1002/smr.2170}
  {\path{doi:10.1002/smr.2170}}.

\bibitem{bohmer_2020_logo}
K.~B{\"o}hmer, S.~Rinderle-Ma, Logo: Combining local and global techniques for
  predictive business process monitoring, in: S.~Dustdar, E.~Yu, C.~Salinesi,
  D.~Rieu, V.~Pant (Eds.), Advanced Information Systems Engineering, Springer
  International Publishing, Cham, 2020, pp. 283--298.

\bibitem{brunk_2020_cause_vs_effect}
J.~Brunk, M.~Stierle, L.~Papke, K.~Revoredo, M.~Matzner, J.~Becker, Cause vs.
  effect in context-sensitive prediction of business process instances,
  Information Systems 95 (09 2020).
\newblock \href {https://doi.org/10.1016/j.is.2020.101635}
  {\path{doi:10.1016/j.is.2020.101635}}.

\bibitem{Jagatheesaperumal_21}
S.~K. Jagatheesaperumal, M.~Rahouti, K.~Ahmad, A.~Al-Fuqaha, M.~Guizani, The
  duo of artificial intelligence and big data for industry 4.0: Review of
  applications, techniques, challenges, and future research directions (2021).
\newblock \href {https://doi.org/10.48550/ARXIV.2104.02425}
  {\path{doi:10.48550/ARXIV.2104.02425}}.

\bibitem{Rehse2019}
J.-R. Rehse, N.~Mehdiyev, P.~Fettke, Towards explainable process predictions
  for industry 4.0 in the dfki-smart-lego-factory, KI - K{\"u}nstliche
  Intelligenz 33~(2) (2019) 181--187.

\bibitem{sachan20_explainable}
S.~Sachan, J.-B. Yang, D.-L. Xu, D.~Benavides, Y.~Li, An explainable ai
  decision-support-system to automate loan underwriting, Expert Systems with
  Applications 144 (Apr. 2020).
\newblock \href {https://doi.org/10.1016/j.eswa.2019.113100}
  {\path{doi:10.1016/j.eswa.2019.113100}}.

\bibitem{inbook}
S.~Meacham, G.~Isaac, D.~Nauck, B.~Virginas, Towards Explainable AI: Design and
  Development for Explanation of Machine Learning Predictions for a Patient
  Readmittance Medical Application, 2019, pp. 939--955.

\bibitem{DBLP:journals/ml/Coma-PuigC22}
B.~Coma{-}Puig, J.~Carmona, Non-technical losses detection in energy
  consumption focusing on energy recovery and explainability, Machine Learning
  111~(2) (2022) 487--517.
\newblock \href {https://doi.org/10.1007/s10994-021-06051-1}
  {\path{doi:10.1007/s10994-021-06051-1}}.

\bibitem{alvarez2018robustness}
D.~Alvarez-Melis, T.~S. Jaakkola, On the robustness of interpretability
  methods, arXiv preprint arXiv:1806.08049 (2018).

\bibitem{Elkhawaga_22_XAI}
G.~Elkhawaga, M.~Abuelkheir, M.~Reichert, Xai in the context of predictive
  process monitoring: Too much to reveal (2022).
\newblock \href {https://doi.org/10.48550/ARXIV.2202.08265}
  {\path{doi:10.48550/ARXIV.2202.08265}}.

\bibitem{Elkhawaga_22_issues}
G.~Elkhawaga, M.~Abuelkheir, M.~Reichert, Explainability of predictive process
  monitoring results: Can you see my data issues? (2022).
\newblock \href {https://doi.org/10.48550/ARXIV.2202.08041}
  {\path{doi:10.48550/ARXIV.2202.08041}}.

\bibitem{Stevens_22}
A.~Stevens, J.~De~Smedt, Explainable artificial intelligence in process mining:
  Assessing the explainability-performance trade-off in outcome-oriented
  predictive process monitoring (2022).
\newblock \href {https://doi.org/10.48550/ARXIV.2203.16073}
  {\path{doi:10.48550/ARXIV.2203.16073}}.

\bibitem{10.1007/978-3-030-91431-8_4}
M.~Velmurugan, C.~Ouyang, C.~Moreira, R.~Sindhgatta, Evaluating stability of
  post-hoc explanations for business process predictions, in: H.~Hacid, O.~Kao,
  M.~Mecella, N.~Moha, H.-y. Paik (Eds.), Service-Oriented Computing, Springer
  International Publishing, Cham, 2021, pp. 49--64.

\bibitem{10.1007/978-3-030-79108-7_8}
M.~Velmurugan, C.~Ouyang, C.~Moreira, R.~Sindhgatta, Evaluating fidelity of
  explainable methods for predictive process analytics, in: S.~Nurcan,
  A.~Korthaus (Eds.), Intelligent Information Systems, Springer International
  Publishing, Cham, 2021, pp. 64--72.

\bibitem{rizzi_explainable_22}
W.~Rizzi, M.~Comuzzi, C.~Di~Francescomarino, C.~Ghidini, S.~Lee, F.~M. Maggi,
  A.~Nolte, Explainable predictive process monitoring: A user evaluation
  (2022).
\newblock \href {https://doi.org/10.48550/arxiv.2202.07760}
  {\path{doi:10.48550/arxiv.2202.07760}}.

\bibitem{10.1007/978-3-319-15895-2_46}
D.~Breuker, P.~Delfmann, M.~Matzner, J.~Becker, Designing and evaluating an
  interpretable predictive modeling technique for business processes, in:
  Business Process Management Workshops, Springer, 2015, pp. 541--553.

\bibitem{DBLP:conf/icpm/HsiehMO21}
C.~Hsieh, C.~Moreira, C.~Ouyang, Dice4el: Interpreting process predictions
  using a milestone-aware counterfactual approach, in: C.~D. Ciccio, C.~D.
  Francescomarino, P.~Soffer (Eds.), 3rd International Conference on Process
  Mining, {ICPM} 2021, Eindhoven, Netherlands, October 31 - Nov. 4, 2021,
  {IEEE}, 2021, pp. 88--95.
\newblock \href {https://doi.org/10.1109/ICPM53251.2021.9576881}
  {\path{doi:10.1109/ICPM53251.2021.9576881}}.

\bibitem{huang2021counterfactual}
T.-H. Huang, A.~Metzger, K.~Pohl, Counterfactual explanations for predictive
  business process monitoring, in: European, Mediterranean, and Middle Eastern
  Conference on Information Systems, Springer, 2021, pp. 399--413.

\bibitem{Rizzi_20_Explainable}
W.~Rizzi, C.~Di~Francescomarino, F.~Maggi, Explainability in Predictive Process
  Monitoring: When Understanding Helps Improving, 2020, pp. 141--158.
\newblock \href {https://doi.org/10.1007/978-3-030-58638-6_9}
  {\path{doi:10.1007/978-3-030-58638-6_9}}.

\bibitem{harl_explainable}
M.~Harl, S.~Weinzierl, M.~Stierle, M.~Matzner, Explainable predictive business
  process monitoring using gated graph neural networks, Journal of Decision
  Systems 0~(0) (2020) 1--16.
\newblock \href {https://doi.org/10.1080/12460125.2020.1780780}
  {\path{doi:10.1080/12460125.2020.1780780}}.

\bibitem{DBLP:conf/bpm/WeinzierlZBRM020}
S.~Weinzierl, S.~Zilker, J.~Brunk, K.~Revoredo, M.~Matzner, J.~Becker, {XNAP:}
  making lstm-based next activity predictions explainable by using {LRP}, in:
  A.~del{-}R{\'{\i}}o{-}Ortega, H.~Leopold, F.~M. Santoro (Eds.), Business
  Process Management Workshops - {BPM} 2020 International Workshops, Seville,
  Spain, September 13-18, 2020, Revised Selected Papers, Vol. 397 of Lecture
  Notes in Business Information Processing, Springer, 2020, pp. 129--141.
\newblock \href {https://doi.org/10.1007/978-3-030-66498-5\_10}
  {\path{doi:10.1007/978-3-030-66498-5\_10}}.

\bibitem{DBLP:conf/bpm/SindhgattaMOB20}
R.~Sindhgatta, C.~Moreira, C.~Ouyang, A.~Barros, Exploring interpretable
  predictive models for business processes, in: D.~Fahland, C.~Ghidini,
  J.~Becker, M.~Dumas (Eds.), Business Process Management - 18th International
  Conference, {BPM} 2020, Seville, Spain, September 13-18, 2020, Proceedings,
  Vol. 12168 of Lecture Notes in Computer Science, Springer, 2020, pp.
  257--272.
\newblock \href {https://doi.org/10.1007/978-3-030-58666-9\_15}
  {\path{doi:10.1007/978-3-030-58666-9\_15}}.

\bibitem{hanga_2020_interpretable_graphs}
K.~M. Hanga, Y.~Kovalchuk, M.~M. Gaber, A graph-based approach to interpreting
  recurrent neural networks in process mining, IEEE Access 8 (2020)
  172923--172938.
\newblock \href {https://doi.org/10.1109/ACCESS.2020.3025999}
  {\path{doi:10.1109/ACCESS.2020.3025999}}.

\bibitem{DBLP:conf/icpm/Pasquadibisceglie21}
V.~Pasquadibisceglie, G.~Castellano, A.~Appice, D.~Malerba, {FOX:} a
  neuro-fuzzy model for process outcome prediction and explanation, in: C.~D.
  Ciccio, C.~D. Francescomarino, P.~Soffer (Eds.), 3rd International Conference
  on Process Mining, {ICPM} 2021, Eindhoven, Netherlands, October 31 - Nov. 4,
  2021, {IEEE}, 2021, pp. 112--119.
\newblock \href {https://doi.org/10.1109/ICPM53251.2021.9576678}
  {\path{doi:10.1109/ICPM53251.2021.9576678}}.

\bibitem{WICKRAMANAYAKE2022108773}
B.~Wickramanayake, Z.~He, C.~Ouyang, C.~Moreira, Y.~Xu, R.~Sindhgatta, Building
  interpretable models for business process prediction using shared and
  specialised attention mechanisms, Knowledge-Based Systems 248 (2022) 108773.
\newblock \href {https://doi.org/https://doi.org/10.1016/j.knosys.2022.108773}
  {\path{doi:https://doi.org/10.1016/j.knosys.2022.108773}}.

\bibitem{DBLP:conf/naacl/JainW19}
J.~Sarthak, B.~C. Wallace, Attention is not explanation, in: Proceedings of the
  2019 Conference of the North American Chapter of the Association for
  Computational Linguistics: Human Language Technologies, Association for
  Computational Linguistics, 2019, pp. 3543--3556.

\bibitem{serrano-smith-2019-attention}
S.~Serrano, N.~A. Smith, Is attention interpretable?, in: Proceedings of the
  57th Annual Meeting of the Association for Computational Linguistics,
  Association for Computational Linguistics, Florence, Italy, 2019, pp.
  2931--2951.

\bibitem{Hochreiter1997}
S.~Hochreiter, J.~{Urgen Schmidhuber}, {Long Short-Term Memory}, Neural
  Computation 9~(8) (1997) 1735--1780.
\newblock \href {http://arxiv.org/abs/1206.2944} {\path{arXiv:1206.2944}}.

\bibitem{RandomForest}
L.~Breiman, \href{https://doi.org/10.1023/A:1010933404324}{Random forests},
  Machine Learning (2001).
\newline\urlprefix\url{https://doi.org/10.1023/A:1010933404324}

\bibitem{shapley1953value}
L.~S. Shapley, A value for n-person games, Contributions to the Theory of Games
  2~(28) (1953) 307--317.

\bibitem{molnar2020interpretable}
C.~Molnar, Interpretable machine learning, Lulu. com, 2020.

\bibitem{galanti2020explainable}
R.~Galanti, B.~Coma-Puig, M.~de~Leoni, J.~Carmona, N.~Navarin, Explainable
  predictive process monitoring, in: Proceedings of the 2nd International
  Conference on Process Mining (ICPM 2020), IEEE, 2020, pp. 1--8.

\bibitem{tedesco2006}
D.~Tedesco, T.~Tullis, A comparison of methods for eliciting post-task
  subjective ratings in usability testing, in: Usability Professionals
  Association Conference (UPA), 2006, pp. 1--9.

\bibitem{salford60928}
A.~Hodrien, T.~Fernando, \href{http://usir.salford.ac.uk/id/eprint/60928/}{A
  review of post-study and post-task subjective questionnaires to guide
  assessment of system usability}, Journal of Usability Studies 16~(3) (2021)
  203--232.
\newline\urlprefix\url{http://usir.salford.ac.uk/id/eprint/60928/}

\bibitem{Sauro2016}
J.~Sauro, J.~Lewis, Standardized usability questionnaires, in: Quantifying the
  User Experience, Elsevier, 2016, pp. 185--248.
\newblock \href {https://doi.org/10.1016/B978-0-12-802308-2.00008-4}
  {\path{doi:10.1016/B978-0-12-802308-2.00008-4}}.

\bibitem{James1992}
J.~Lewis, Psychometric evaluation of the post-study system usability
  questionnaire: The {PSSUQ}, Proceedings of the Human Factors Society Annual
  Meeting 36 (1992) 1259--1260.
\newblock \href {https://doi.org/10.1177/154193129203601617}
  {\path{doi:10.1177/154193129203601617}}.

\bibitem{10.1007/978-3-540-89350-9_6}
B.~Laugwitz, T.~Held, M.~Schrepp, Construction and evaluation of a user
  experience questionnaire, in: A.~Holzinger (Ed.), HCI and Usability for
  Education and Work, Springer Berlin Heidelberg, Berlin, Heidelberg, 2008, pp.
  63--76.

\bibitem{10.1007/978-3-319-07668-3_37}
M.~Schrepp, A.~Hinderks, J.~Thomaschewski, Applying the user experience
  questionnaire ({UEQ}) in different evaluation scenarios, in: A.~Marcus (Ed.),
  Design, User Experience, and Usability. Theories, Methods, and Tools for
  Designing the User Experience, Springer International Publishing, Cham, 2014,
  pp. 383--392.

\bibitem{doi:10.1080/10447318.2002.9669130}
J.~R. Lewis, Psychometric evaluation of the {PSSUQ} using data from five years
  of usability studies, International Journal of Human–Computer Interaction
  14~(3-4) (2002) 463--488.
\newblock \href {https://doi.org/10.1080/10447318.2002.9669130}
  {\path{doi:10.1080/10447318.2002.9669130}}.

\bibitem{Schrepp_ueq_questionnaire}
M.~Schrepp, User experience questionnaire handbook. all you need to know to
  apply the ueq successfully in your project. (09 2015).
\newblock \href {https://doi.org/10.13140/RG.2.1.2815.0245}
  {\path{doi:10.13140/RG.2.1.2815.0245}}.

\bibitem{10.1007/978-3-031-07481-3_7}
H.~L{\"o}fstr{\"o}m, K.~Hammar, U.~Johansson, A meta survey of quality
  evaluation criteria in explanation methods, in: Proceedings of CAiSE Forum
  2022, Vol. 452 of LNBIP, Springer, 2022, pp. 55--63.

\end{thebibliography}

\appendix
\counterwithin{figure}{section}

\section{}
\label{appendix:additional_experiments}

\subsection{Bank Account Closure}

\begin{figure}[h!]
    \includegraphics[width=1 \columnwidth]{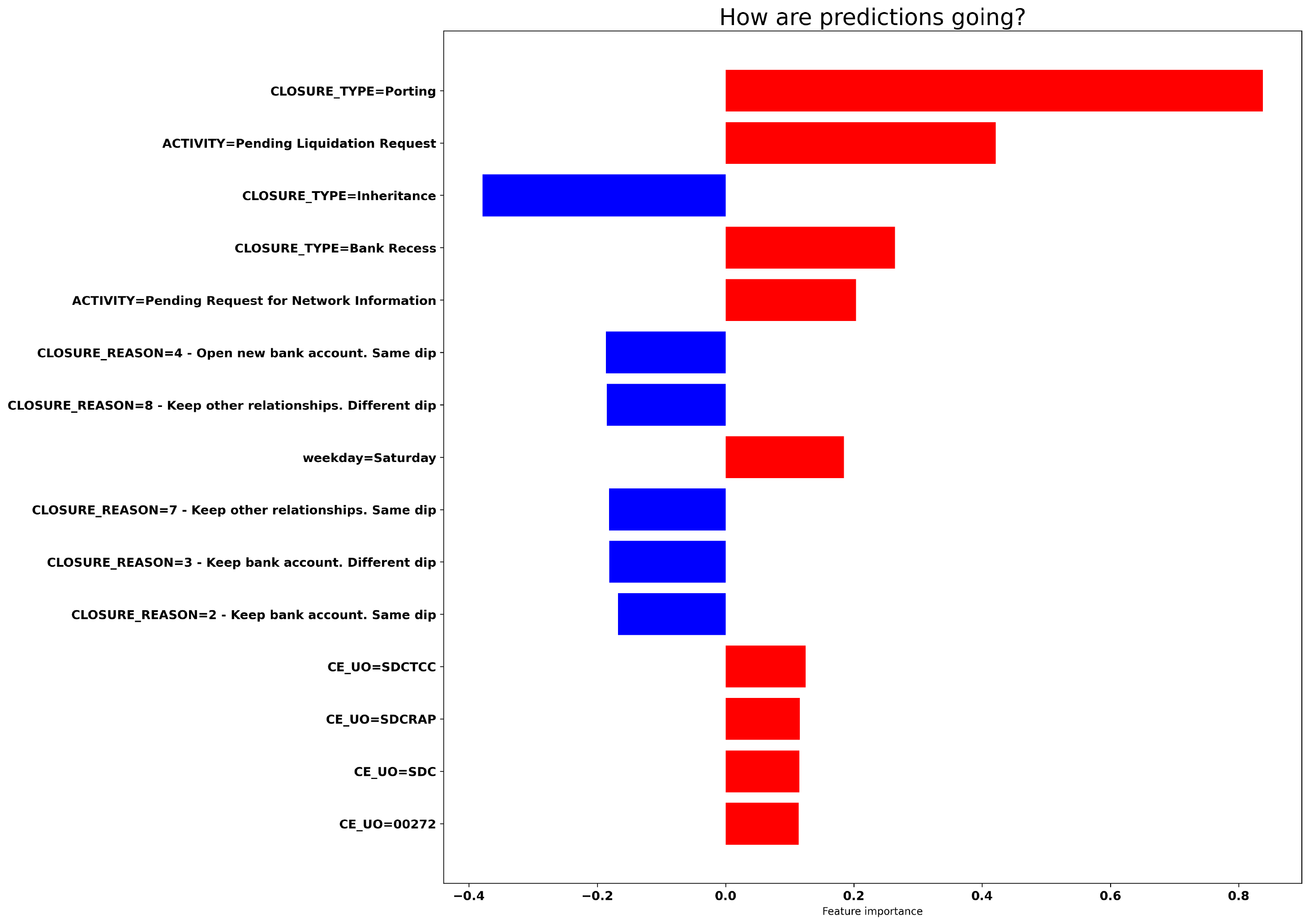}
    \caption{Offline explanations for \emph{Back-Office Adjustment Requested} prediction (Bank Account Closure)}
    \label{fig:explanation_bac_bo_adj}
\end{figure}

The bar chart related to \emph{Back-Office Adjustment Requested} prediction (Figure \ref{fig:explanation_bac_bo_adj}) shows that the attributes related to the type and the reason of bank account closure are influencing the most.

When the closure type is Porting, it indicates that the customer has decided to move one or more services and the current-account balance from one bank account to another. Here, it is the most influential factor and its influence is towards strongly increasing the probability that a Back-Office Adjustment will be requested (as it can be seen by the red bar associated with a value of 0.85). Also the influence of \textit{Closure\_type=Bank Recess} is towards increasing the probability; a further analysis of the data confirmed that more than 2/3 of the total back-office adjustments is performed when the process is characterized by one of these two closure types.
When a customer decides to open a new bank account in the same department of the previous bank account and decides to close the old one, it is less likely that a Back-office Adjustment will be performed; this can be seen by looking at the explanation \textit{Closure\_reason=4 - Open new bank account. Same dip}, which is associated with a negative probability of -0.20. Finally, an interesting thing to be noticed is that when one or more of the activities previously performed in the process were performed on Saturday, then it's more likely that a Back-Office Adjustment will be needed in the future, as highlighted by the explanation \textit{weekday=Saturday} associated with the positive value 0.20. 

\begin{figure}[t!]
    \includegraphics[width=1 \columnwidth]{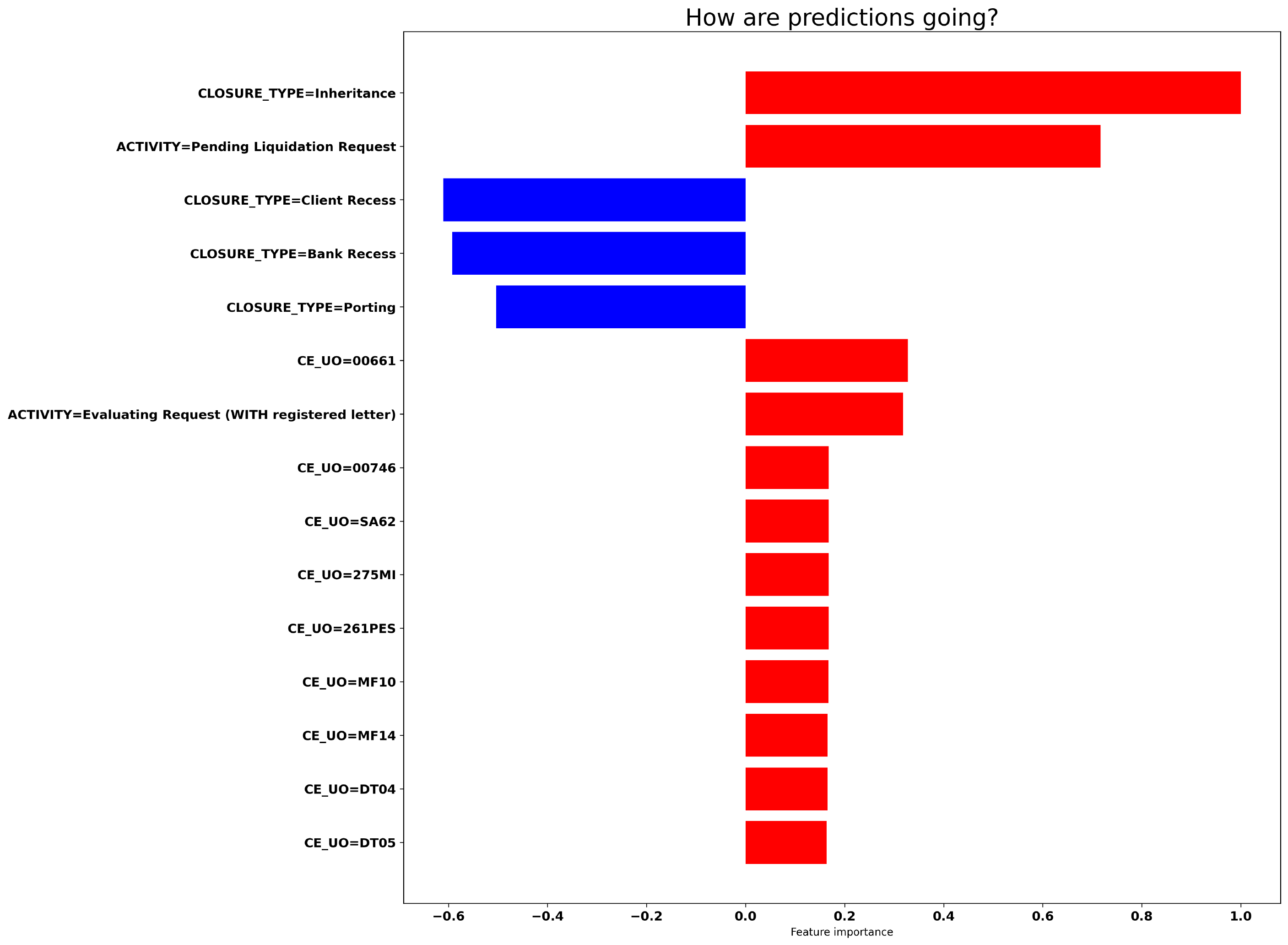}
    \caption{Offline explanations for \emph{Pending Request for Acquittance of Heirs} prediction (Bank Account Closure)}
    \label{fig:explanation_bac_pen_req}
\end{figure}

The results for the third activity on which the bank decided to focus (\textit{Pending Request for Acquittance of Heirs}) are reported in Figure~\ref{fig:explanation_bac_pen_req}. 
Here, the closure type is one of the most influential factors. In particular, the fact that the closure type is Inheritance (\emph{Closure\_Type=Inheritance}) strongly influences towards predicting that a \textit{Pending Request for Acquittance of Heirs} will occur; this can be seen by the large red bar associated with a value that is near 1. Analyzing the data, it has been seen that a \textit{Pending Request for Acquittance of Heirs} is performed most of the times when the closure type is Inheritance, while it is very unlikely to be performed when other closure types are performed. This also reflects in the blue bars related to the other three types of closure type (\emph{Closure\_Type=Client Recess}, \emph{Closure\_Type=Bank Recess} and \emph{Closure\_Type=Porting}), which are associated with a value -0.60.
Moreover, in the lower part of the plot, it can be seen that other attributes that are influencing the prediction by slightly increasing the probability are related to the resource (shown in the plot as \textit{Ce\_Uo}) that has performed one or more of the activities preceding \textit{Pending Request for Acquittance of Heirs}.

\begin{figure}[t!]
    \includegraphics[width=1 \columnwidth]{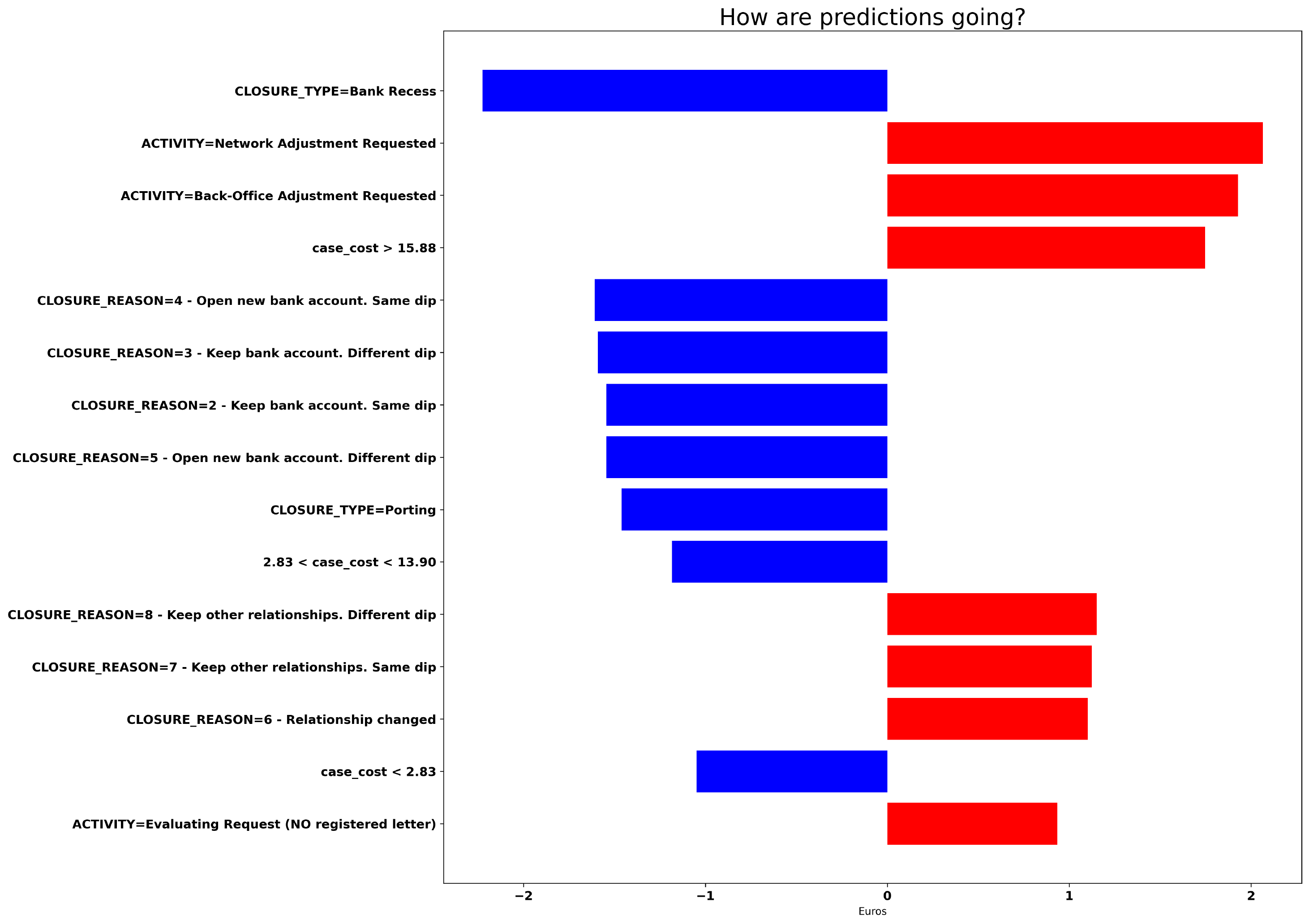}
    \caption{Offline explanations for Cost prediction (Bank Account Closure)}
    \label{fig:explanation_bac_cost}
\end{figure}

Figure~\ref{fig:explanation_bac_cost} shows the application of our framework for the case cost prediction. 
The main factor that contributes to decrease the cost of a case is represented by \emph{Closure\_type=Bank Recess}, which is indicated when the closure of the bank account is requested by the bank.
The information that the value is negative (i.e. -2 Euros) indicates that the influence is towards decreasing the cost. This is mainly caused by the fact that most of the times here the director does not need to carefully evaluate the request before proceeding and, since the hourly director's wage is certainly higher than that of other bank employees, the predicted case cost will be smaller. 
The director is similarly not involved when customers, for different reasons, decide to close only one of their bank accounts (labeled respectively as \emph{Closure\_Reason=4 - Open new bank account. Same dip}, \emph{Closure\_Reason=3 - Keep bank account. Different dip} and \emph{Closure\_Reason=2 - Keep bank account. Same dip}), which is a factor that yields lower costs.
Another reason is that when only one between different bank accounts of a customer is closed, then the process carried out in the bank is simpler and less Back-office adjustment activities need to be performed compared to when all bank accounts need to be closed, leading to minor costs.
The other main factors that contribute to increase the cost of the case are represented by \textit{Activity=Network Adjustment Requested} and \textit{Activity=Back-Office Adjustment Requested}. The information that the values are positive (i.e. 2 Euros) indicates that the influence is towards increasing the cost. This is mainly caused by the fact that these activities should be avoided, since they are only performed when some problem has occurred during the processing of the customer's request and a rework needs to be done, leading to inefficiencies in terms of time, costs, and resource utilization.

\subsection{BPIC 2012}

\begin{figure}[t!]
    \includegraphics[width=1 \columnwidth]{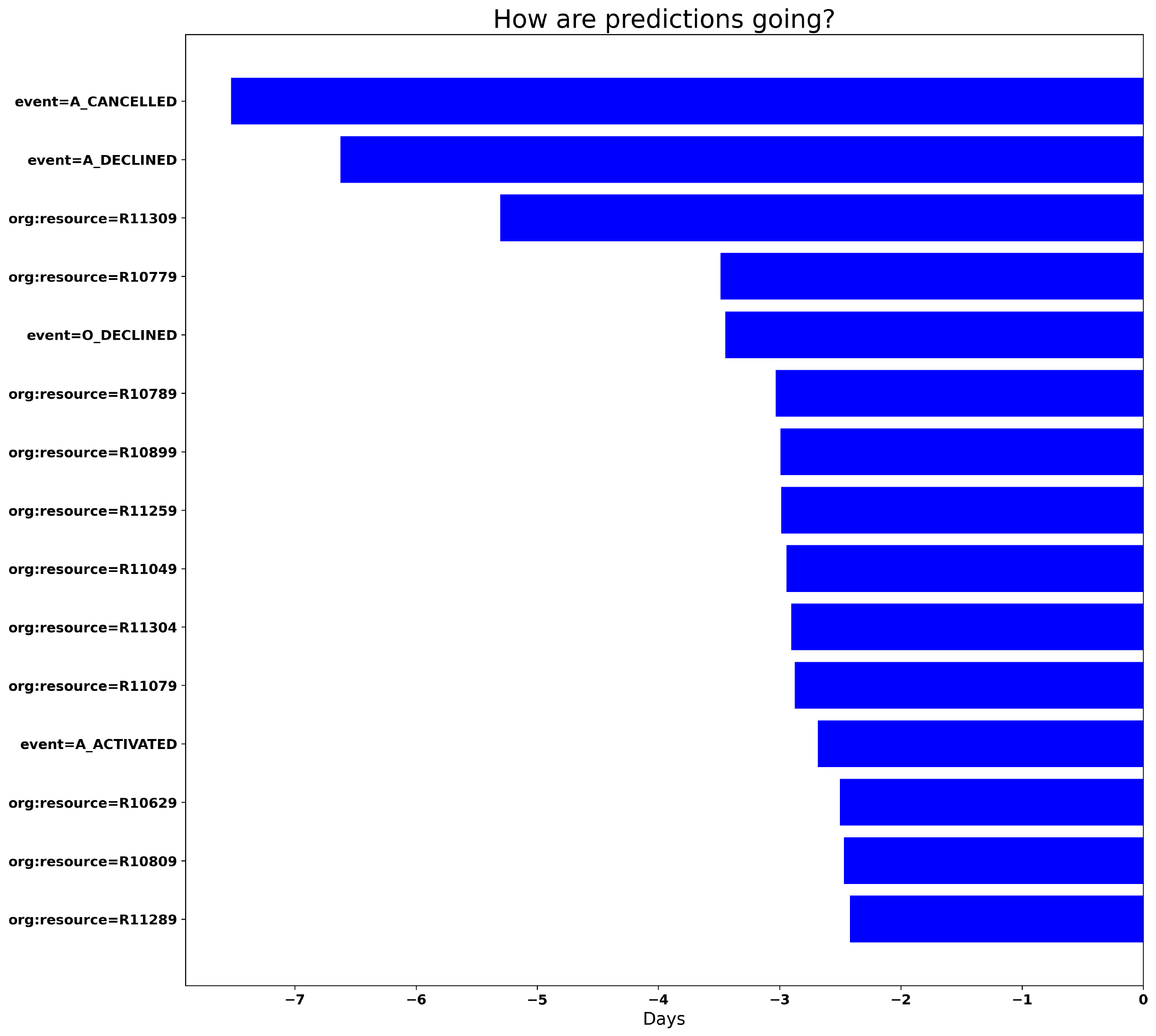}
    \caption{Offline explanations for Remaining time prediction (BPIC 2012)}
    \label{fig:explanation_bpi12_rem_time}
\end{figure}

This is the dataset from BPI 2012 challenge, and it represents an application process for a personal loan or overdraft within a global financing organization.
The event log is a composition of three merged intertwined sub processes. The first letter of each task name identifies from which sub process (source) it originated from.
Figure \ref{fig:explanation_bpi12_rem_time} shows the application for the remaining time prediction in the bpi12 dataset.

It can be clearly seen by the color blue of the bars and the negative values that the influence of the principal factors is towards decreasing the remaining time prediction. 
In particular, the fact that a particular activity has been previously performed in the process, is one of the largest factors that influences the prediction. This can be seen in the explanations \textit{event=A\_CANCELLED}, \textit{event=A\_DECLINED}, and \textit{event=O\_DECLINED}, which are associated to the values -7.5, -6.5, and -3.5 respectively. 
From a domain viewpoint, the application for a personal loan can be rejected by the financial organization (action represented by the task A\_DECLINED), or it can be directly closed because the user decides to cancel the request (represented by the task A\_CANCELLED); therefore, when these actions are performed, the loan application process will be suddenly interrupted and it will finish earlier compared to the average duration of the application process, motivating the very high negative impact on the remaining time prediction. 
This also happens when the offer that has been sent by the organization is rejected by the user (action represented by the task O\_DECLINED); the negative impact on the remaining time is motivated by the fact that the organization can still make a counter-proposal to the customer, but it can also lead to the closure of the application process. However, since this action is generally performed later in the process compared to the two previous actions (which occur in the early stages of the process), it has a lower impact on the remaining time (-3.5 compared to -6.5 and -7.5).   
The other attributes that are influencing the prediction are related to the resources that have performed one or more activities in the process.

\begin{figure}[t!]
    \includegraphics[width=1 \columnwidth]{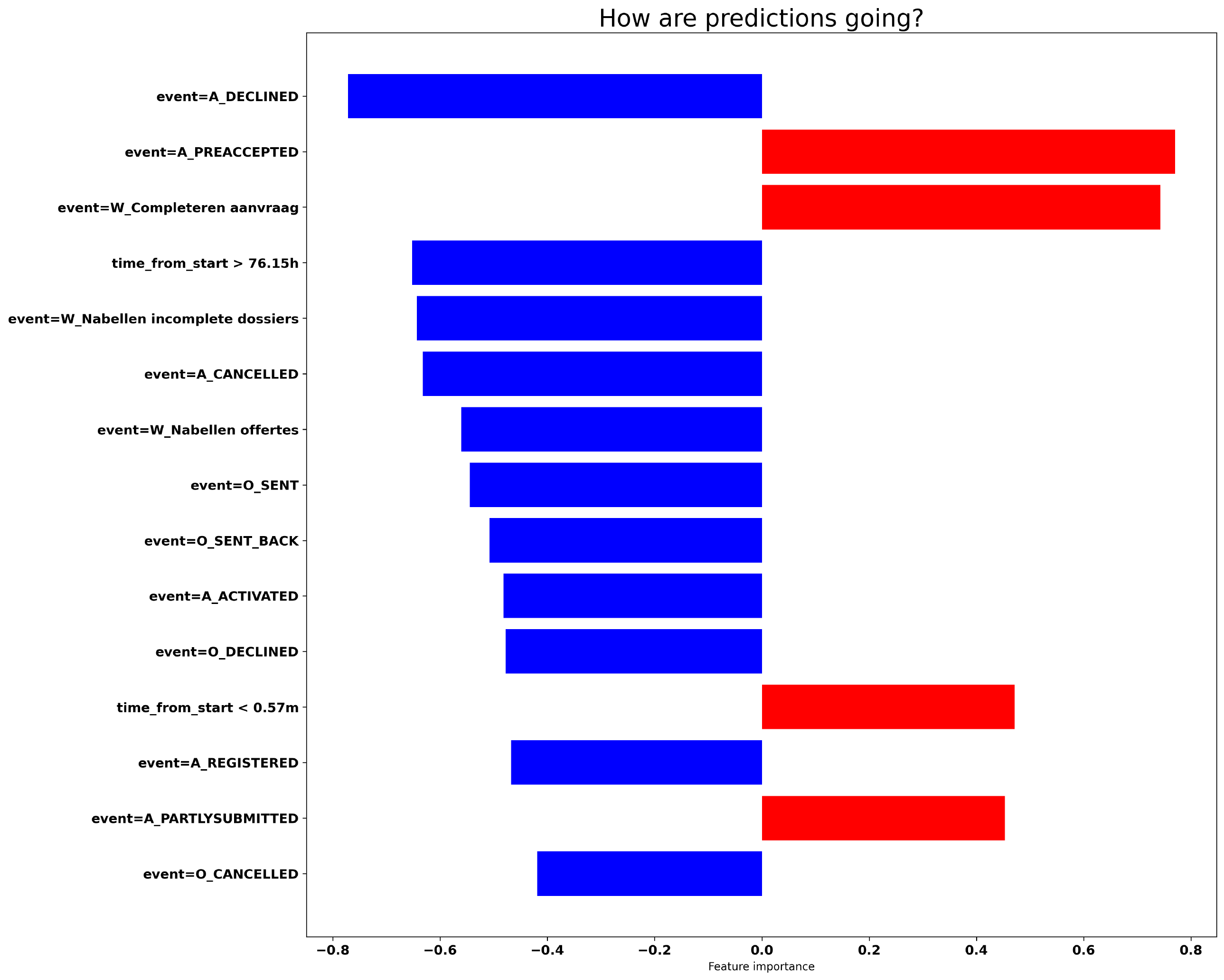}
    \caption{Offline explanations for \textit{A\_ACCEPTED} prediction (BPIC 2012)}
    \label{fig:explanation_bpi12_A_ACCEPTED}
\end{figure}

In this process we also focused on three activities, namely \textit{A\_ACCEPTED}, \linebreak\textit{A\_DECLINED}, and  \textit{A\_CANCELLED}, whose occurrence would be interesting to be predicted beforehand since they indicate if the application has been accepted, declined or cancelled respectively.

Figure~\ref{fig:explanation_bpi12_A_ACCEPTED} reports on the outcome of the application of our explainable framework for activity \textit{A\_ACCEPTED} occurrence prediction. 
It can be easily seen that the attributes related to the activity performed are largely influencing the prediction. 
The most influential factor here is \textit{event=A\_DECLINED}. When this activity is performed, there is a very high probability (-0.80) that the loan application is not going to be accepted. Conversely, as it can be seen by the red bars, the influence of \textit{event=A\_PREACCEPTED} and \textit{event=W\_Completeren aanvraag} is towards highly increasing the probability to have the application accepted; these are in fact activities which are usually performed immediately before proceeding with the acceptance of the application.
Moreover, other activities similarly influence the prediction but, since in the process they are usually performed after the acceptance of the application, their influence is towards lowering the probability that the application will be accepted afterwards, as it can be seen by the blue bars related to the explanations \textit{event=W\_Nabellen incomplete dossiers},\textit{event=W\_Nabellen offertes}, \textit{event=O\_SENT}, \textit{event=O\_SENT\_BACK}, \textit{event=A\_ACTIVATED} and \textit{event=O\_DECLINED} and associated with values -0.50 and -0.60.
Finally, since the application is usually accepted in the initial part of the process, the time elapsed since the beginning of the process is a crucial feature. As it can be seen, when the time elapsed is above a certain threshold (\textit{$time\_from\_start > 76.15h$}), then the influence is towards lowering the probability that the application will be accepted by -0.65; instead, when the time elapsed is below a certain threshold (\textit{$time\_from\_start < 0.57m$}), then the influence is towards increasing the probability by 0.50.

\begin{figure}[t!]
    \includegraphics[width=1 \columnwidth]{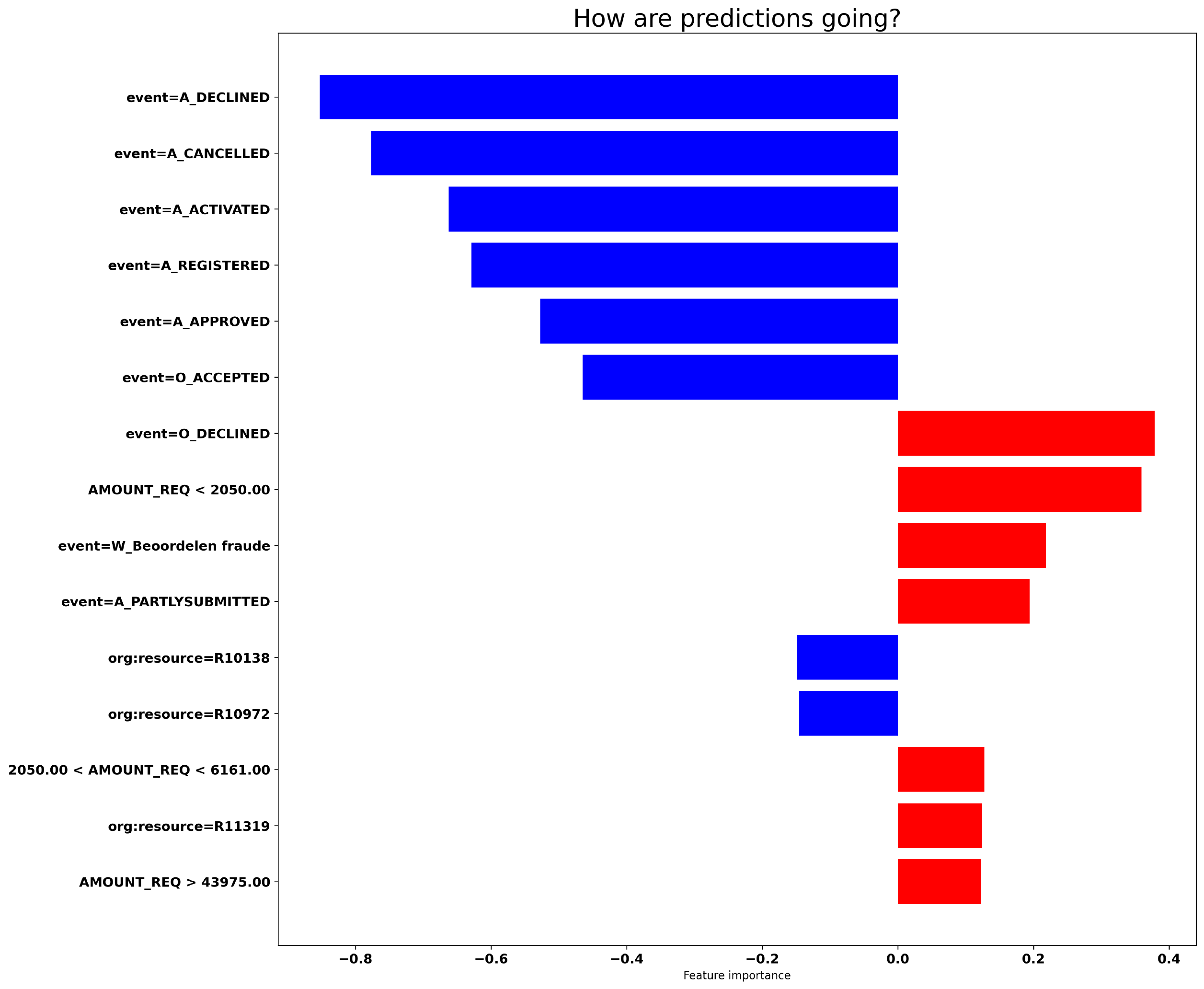}
    \caption{Offline explanations for \textit{A\_DECLINED} prediction (BPIC 2012)}
    \label{fig:explanation_bpi12_A_DECLINED}
\end{figure}

The explanations related to the activity \textit{A\_DECLINED} are shown in Figure \ref{fig:explanation_bpi12_A_DECLINED}. 
As before, attributes related to the activity performed are the most important influencers for the prediction. 
When the application is declined (\textit{event=A\_DECLINED}), there is still a chance that the organization makes a counter-proposal to the customer; however, since it is rare that an application is declined twice, if \textit{A\_DECLINED} is performed once, than the influence is strongly towards not predicting that this activity will occur again in the future (as it can be seen by the strong negative value -0.85).
The same reasoning applies also for \textit{event=A\_CANCELLED}, which is associated with value -0.80. 
Moreover, also when the acceptance process for the loan application has started, the probability to predict that the loan will be declined decreases; this is highlighted by the blue bars associated with the explanations \textit{event=A\_ACTIVATED}, \textit{event=A\_REGISTERED}, \textit{event=A\_APPROVED}, \textit{event=O\_ACCEPTED}, with values -0.65, -0.60, -0.50, -0.45, respectively.
Conversely, activity \textit{O\_DECLINED} can be directly followed by \textit{A\_DECLINED}; therefore when activity \textit{O\_DECLINED} is performed (\textit{event=O\_DECLINED}), the influence is towards predicting that the application will be declined (as underlined by the red bar with value 0.40).

\begin{figure}[t!]
    \includegraphics[width=1 \columnwidth]{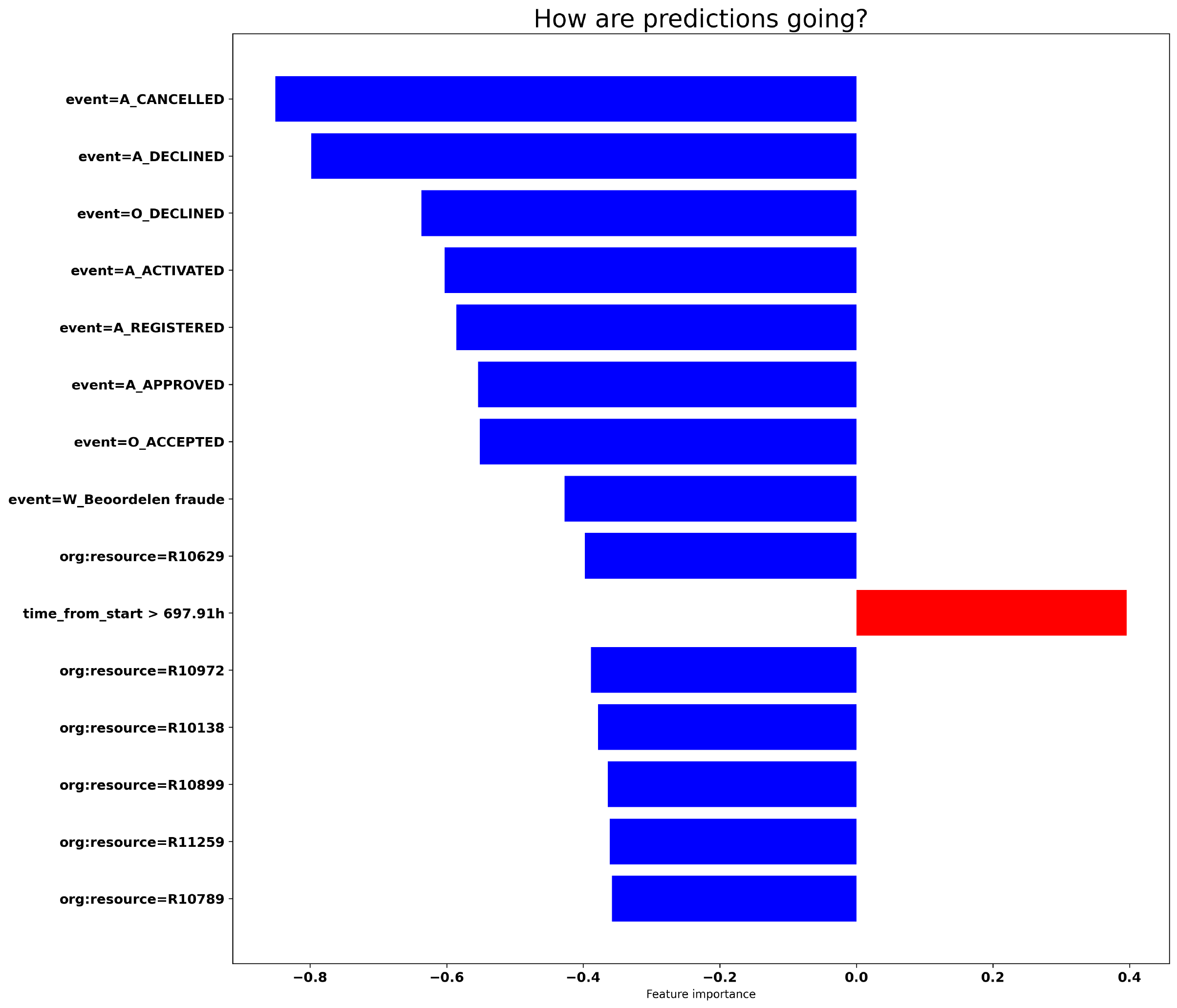}
    \caption{Offline explanations for \textit{A\_CANCELLED} prediction (BPIC 2012)}
    \label{fig:explanation_bpi12_A_CANCELLED}
\end{figure}

Finally, Figure \ref{fig:explanation_bpi12_A_CANCELLED} reports on the explanations for \textit{A\_CANCELLED} occurrence prediction, which are very similar to the explanations shown in Figure \ref{fig:explanation_bpi12_A_DECLINED}.
Since it is rare that an application is cancelled twice, if \textit{A\_CANCELLED} is performed once, than the influence is strongly towards not predicting that this activity will occur again (as it can be seen by the strong negative value -0.85).
The same reasoning applies also for \textit{event=A\_DECLINED} and \textit{event=O\_DECLINED}, which is associated with value -0.80. 
Additionally, as it was previously mentioned, when the acceptance process for the loan application has started, the probability to predict that the loan will be cancelled decreases; this is underlined by the blue bars associated with the explanations \textit{event=A\_ACTIVATED}, \textit{event=A\_REGISTERED}, \textit{event=A\_APPROVED}, and \textit{event=}\linebreak\textit{O\_ACCEPTED}, with values -0.60, -0.60, -0.55, -0.55, respectively.
Conversely, activity \textit{O\_DECLINED} can be directly followed by \textit{A\_DECLINED}; therefore when activity \textit{O\_DECLINED} is performed (\textit{event=O\_DECLINED}), the influence is towards predicting that the application will be declined (as underlined by the red bar with value 0.40).

\subsection{BPIC 2012 - W}

\begin{figure}[t!]
    \includegraphics[width=1 \columnwidth]{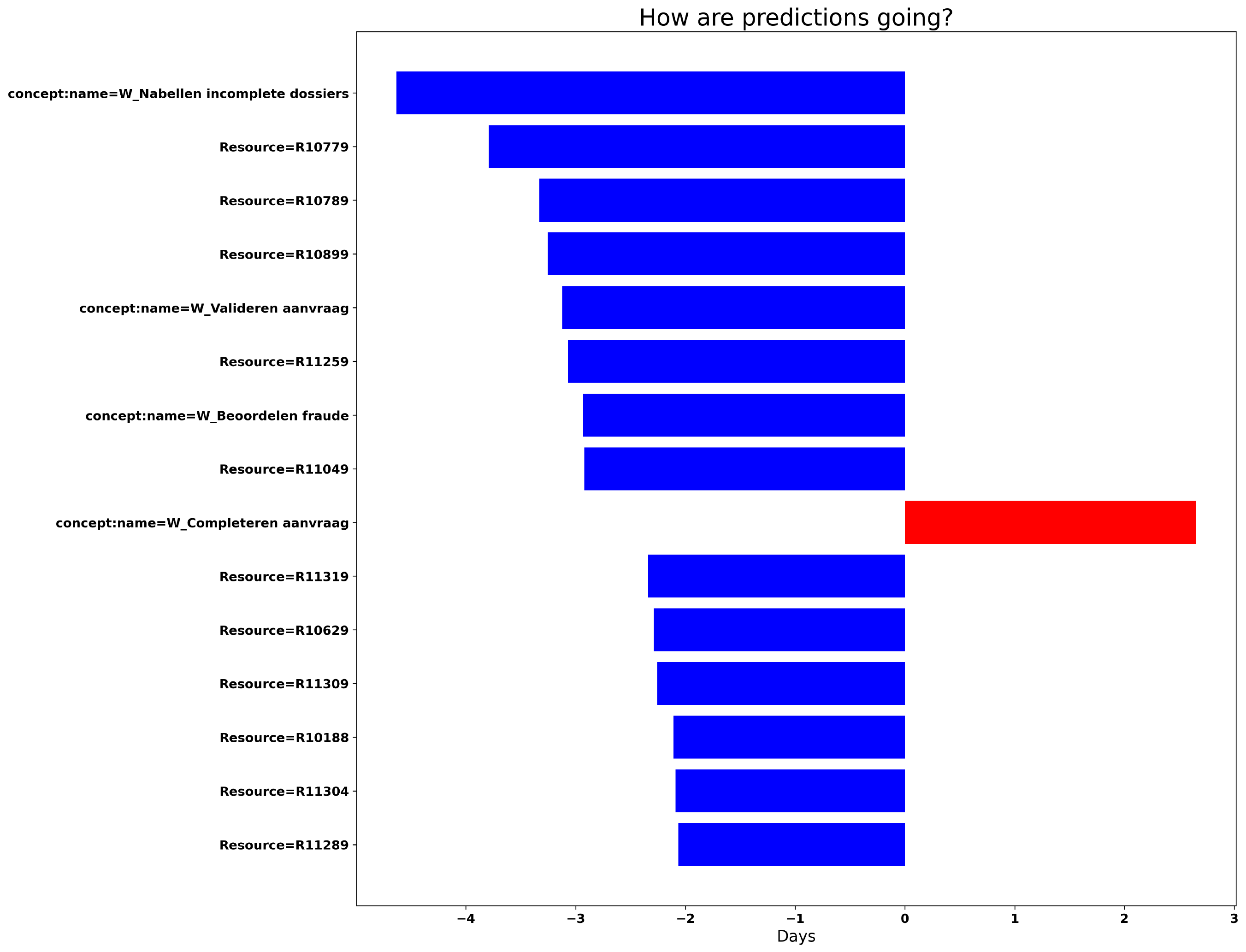}
    \caption{Offline explanations for Remaining Time prediction (BPIC 2012 - W)}
    \label{fig:explanation_bpi12_w_rem_time}
\end{figure}

Figure \ref{fig:explanation_bpi12_w_rem_time} shows the application for the remaining time prediction in the Bpi12 - W dataset, which is the dataset derived from bpi12 challenge, representing the subprocess containing only the states of the work items belonging to the application.
Here, the most important influencers are the attributes related to the activity and the resource performing the activity.
The main factor that contributes to decrease the remaining time of a case is represented by \textit{concept:name=W\_Nabellen incomplete dossiers}.
The information that the value is negative (i.e. -4.5 days) indicates that the influence of performing the activity \textit{W\_Nabellen incomplete dossiers} is towards decreasing the remaining time. This is mainly caused by the fact that this activity is usually performed in the final part of cases, therefore the remaining time to finish the case will be smaller. 
This also regards other activities performed in the final part of cases (\textit{concept:name=W\_Valideren aanvraag} and \textit{concept:name=W\_Beoordelen fraude}) and the related resources performing these activities, which influence is towards decreasing the remaining time. 
Conversely, \textit{concept:name=W\_Completeren aanvraag} is one of the first activities that are performed in this process, therefore the contribution is towards increasing the remaining time prediction, as it can be seen by the red bar associated with a value of 3 days.

\subsection{BPIC 2013}

This is a dataset from BPI 2013 challenge, extracted from Volvo incident management system.
Here, we focused on obtaining an estimate of the remaining time until the end for running cases, to detect the cases requiring special attention. 

\begin{figure}[t!]
    \includegraphics[width=1 \columnwidth]{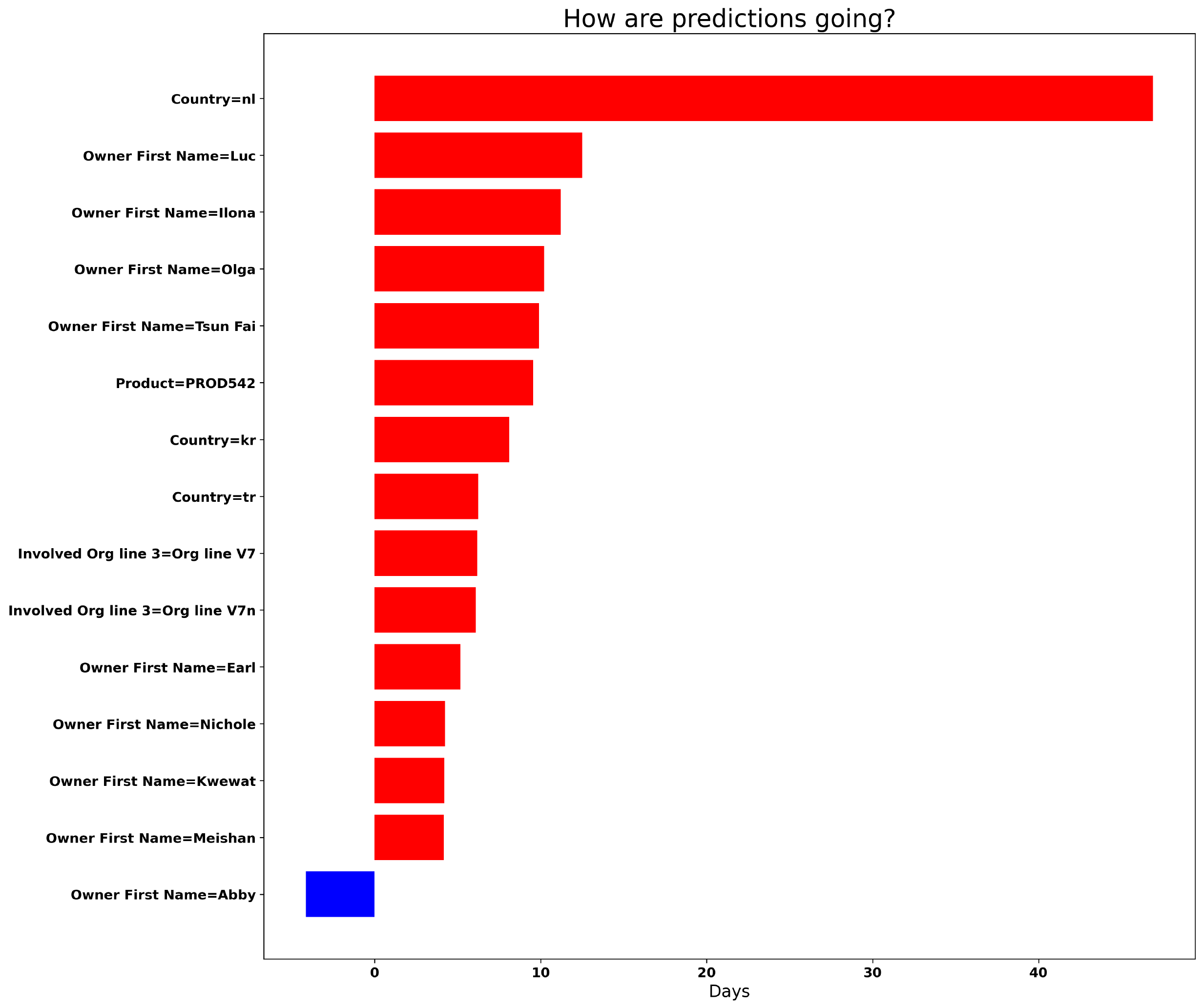}
    \caption{Offline explanations for Remaining Time prediction (BPIC 2013)}
    \label{fig:explanation_bpi13_rem_time}
\end{figure}

Figure \ref{fig:explanation_bpi13_rem_time} reports on the outcome of the application of our explainable framework for remaining time prediction in Bpi13 dataset. 
The fact that the country in which the incident is managed is Netherlands (\textit{Country=nl}) is one of the largest factors that influences the prediction; the color red of the bar and the information that the value is positive (~50 days) indicates that the influence is towards increasing the remaining time. 
A further analysis of the data confirms this finding: if the country is Netherlands, the process duration is 197 days, versus 12.1 days when the type is different. 
This also applies to the situation where the incident is managed in South Korea (\textit{Country=kr}) or in Turkey (\textit{Country=tr}), where the influence is towards increasing the remaining time by ~8 and ~6 days respectively; again, this finding was supported by the data, since it was observed a longer duration of the process (16d 4h and 19d 9h respectively). 
Other important attributes are related to the resource of the support team (indicated by \textit{Owner First Name}) in charge of working on the incident. Here, for example, the fact that one or multiple activities were performed by Luc, Ilona or Olga, contributed to increase the remaining time prediction by ~10 days. 
Also the support team that is in charge of solving the problem can influence the remaining time. Here, the fact that a 3rd level support team has been involved (indicated by \textit{Involved Org line 3=Org line V7}, \textit{Involved Org line 3=Org line V7n}) increased the remaining time by ~7 days; this is probably due to the fact that only the problems of a certain severity level are passed to the 3rd level support team, hence requiring more time to be solved.

This leads us to analyze one of the KPIs in which the company was interested in, the push to front strategy.
In particular, the company wanted to resolve most of the incidents with the first line support teams without involving 2nd or 3rd support line teams (strategy called push to front), in order to have a more efficient process and avoid having too much work concentrated on 2nd and 3rd line support teams. Therefore, we also focused on predicting if at least one activity is going to be performed by a resource not belonging to the first line support team. 

\begin{figure}[t!]
    \includegraphics[width=1 \columnwidth]{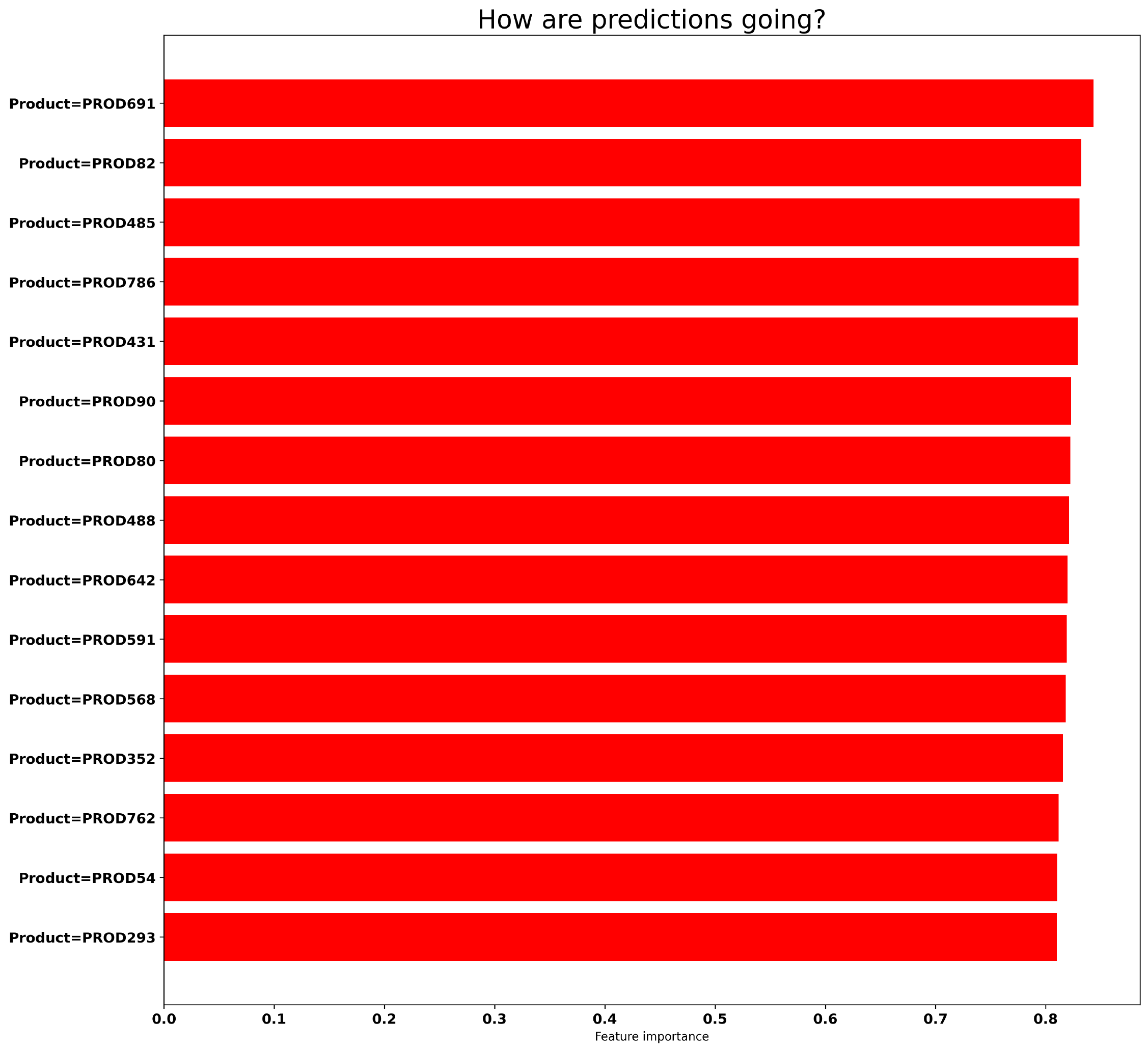}
    \caption{Offline explanations for \textit{Push to Front} prediction (BPIC 2013)}
    \label{fig:explanation_bpi13_push_front}
\end{figure}

The bar chart related to the push to front prediction is shown in Figure \ref{fig:explanation_bpi13_push_front};
It can be clearly seen that the attributes related to the product that presented a problem or that was involved in an incident are the most important ones.
The influence for all of them is towards strongly predicting that a resource from the 2° or 3° line support team will be involved in the management of the incident, as it can be seen by the color red of the bars and the very high positive values (0.80). 
The data confirmed that the products reported in these explanations (such as (\textit{Product=PROD691})) were always associated to the involvement of a resource from the 2° or 3° level support team, thus confirming the findings reported here.  

Finally, the company had also an interest in understanding if people working in the company were abusing the \emph{Wait - User} substatus to hide inefficiencies in the process, which would otherwise being detected by KPIs measuring the total resolution time of an incident.
Therefore, the main objective here is to try to detect as soon as possible if this status will be set.

\begin{figure}[t!]
    \includegraphics[width=1 \columnwidth]{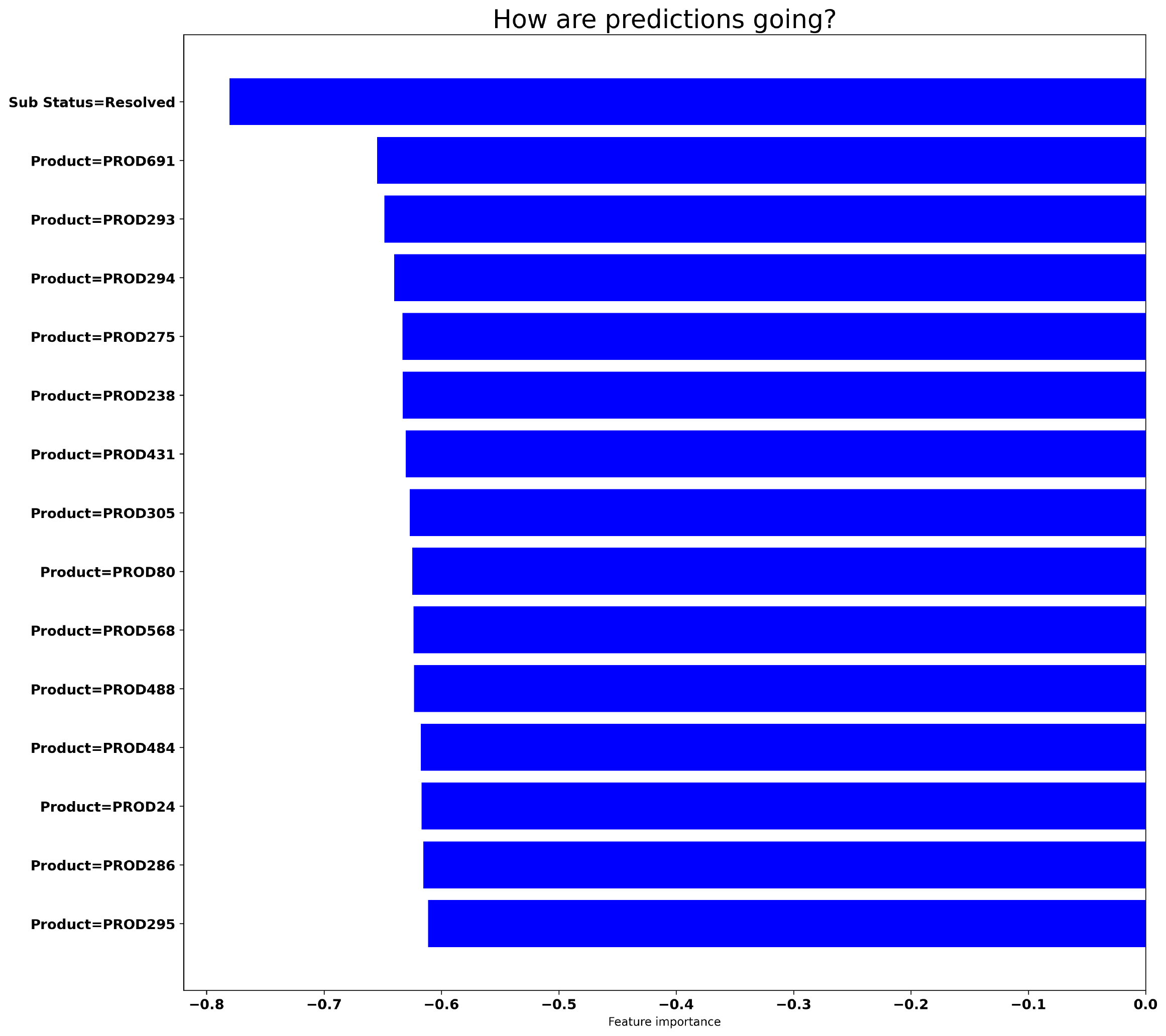}
    \caption{Offline explanations for \textit{Wait - User} prediction (BPIC 2013)}
    \label{fig:explanation_bpi13_wait_user}
\end{figure}

Figure \ref{fig:explanation_bpi13_wait_user} shows the results for the \emph{Wait - User} substatus prediction.
Here, the fact that the \textit{Resolved} substatus has been set (labeled as \textit{Sub Status=Resolved}) is the most important factor, and its contribute is towards strongly decreasing the probability that a \emph{Wait - User} substatus will be set (as it can be seen by the blue bar and the associated value -0.80). The analysis of the process confirmed that it is unlikely that a \emph{Wait - User} status is set after the \textit{Resolved} status, which is indeed the status indicating that the incident has been already solved.
Other important attributes are related to the product that presented a problem or that was involved in an incident; the influence for all of them is towards strongly predicting that the \emph{Wait - User} substatus will not be set during the resolution of the problem.
The data also confirmed that the products reported in these explanations (such as \textit{Product=PROD691}) were never associated to the setting of the status \emph{Wait - User}, thus confirming the findings reported here.

\subsection{HelpDesk 2017}

This dataset is a real-life log of SIAV s.p.a. company in Italy, and represents instances of a ticketing process in the company helpdesk area.

\begin{figure}[t!]
    \includegraphics[width=1 \columnwidth]{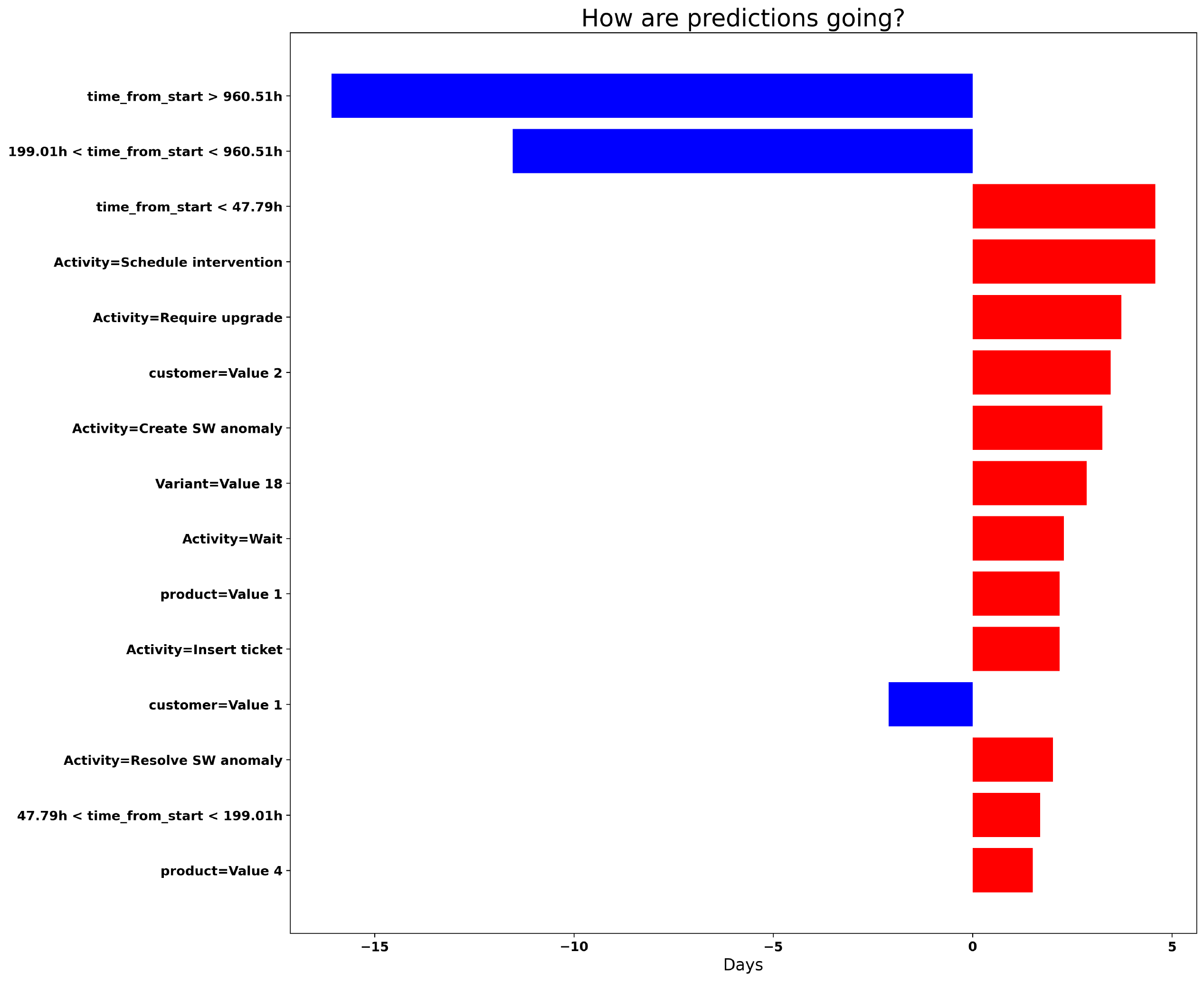}
    \caption{Offline explanations for Remaining Time prediction (HelpDesk 2017)}
    \label{fig:explanation_helpdesk_rem_time}
\end{figure}

Figure \ref{fig:explanation_helpdesk_rem_time} refers to the application for the remaining time prediction.
It can be clearly seen by the color blue of the bars and the negative values that the influence of the principal factors is towards decreasing the remaining time prediction. 
In particular, the time elapsed since the beginning of the process is a crucial feature. As it can be seen, when the time elapsed is between certain thresholds (\textit{$199.01h < time\_from\_start < 960.51h$}), then the influence is towards lowering the predicted remaining time by -12 days. Moreover, when the time elapsed is larger than the average process duration (which was 40 days, the same duration highlighted in \textit{$time\_from\_start > 960.51h$}), the influence towards lowering the predicted remaining time becomes even stronger (-16 days); this is mainly caused by the fact that when the elapsed time is above the average process duration, it is highly probable that the case is going to be closed soon, thus lowering the remaining time prediction.
Conversely, when the process is still in the initial phases (\textit{$time\_from\_start < 47.79h$}), the influence is towards predicting a higher remaining time (+5 days).
Other important attributes are related to the performed activity. In this process a ticket, after having been opened and after assigning a seriousness level, is usually taken in charge by a resource and resolved; however, after having been taken in charge, some activities can occur, such as \textit{Schedule intervention}, \textit{Require upgrade}, \textit{Create SW anomaly} and \textit{Wait}. These are all exceptional activities that are rarely performed, thus increasing the remaining time prediction, as it can be seen by the red bars associated to \textit{Activity=Schedule intervention}, \textit{Activity=Require upgrade}, \textit{Activity=Create SW anomaly} and \textit{Activity=Wait}.

\subsection{Road Traffic Fine Management}

This dataset is a real-life event log of an information system managing road traffic fines.

\begin{figure}[t!]
    \includegraphics[width=1 \columnwidth]{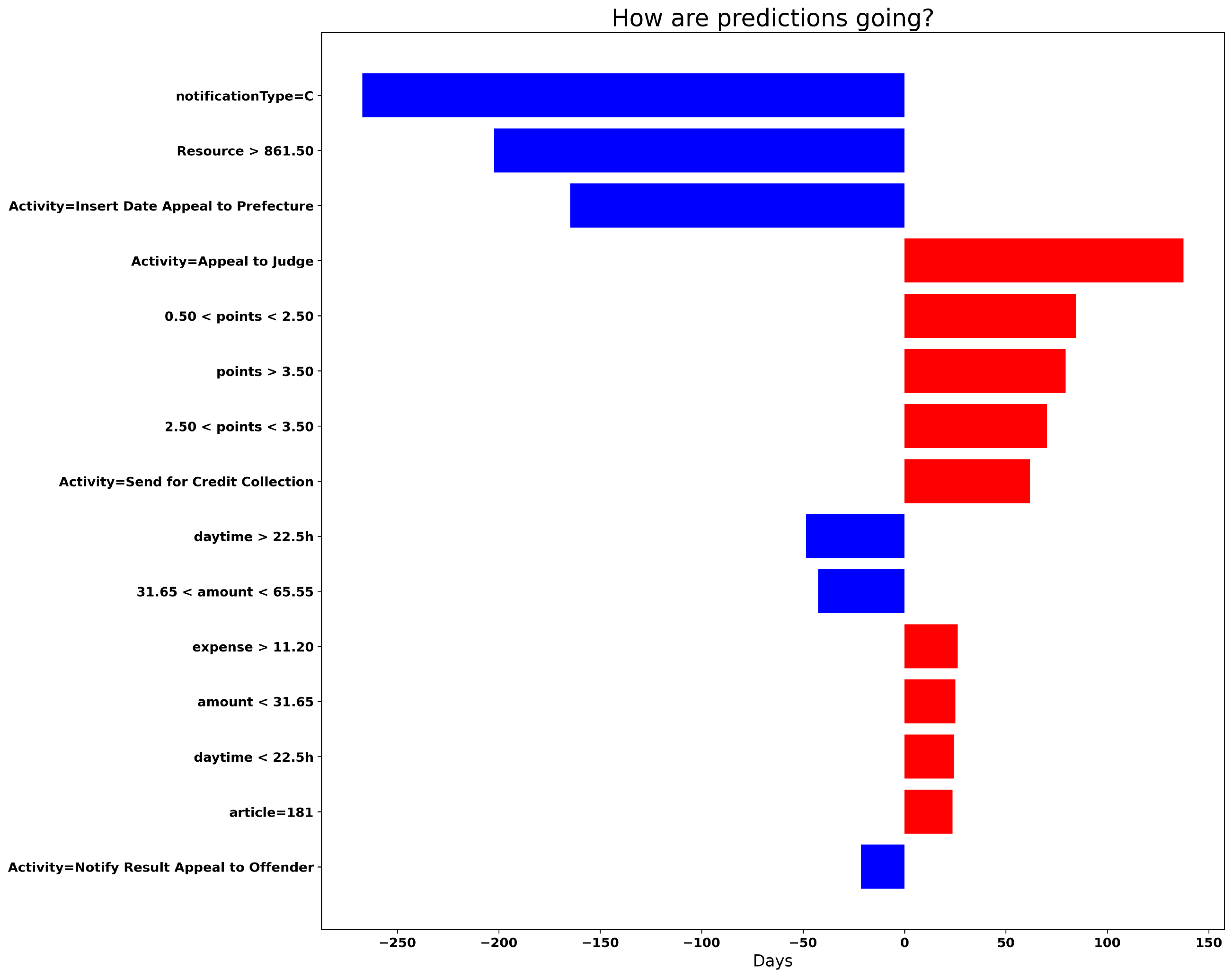}
    \caption{Offline explanations for Remaining Time prediction (Fine Management)}
    \label{fig:explanation_fines_rem_time}
\end{figure}

Figure \ref{fig:explanation_fines_rem_time} reports on the outcome of the application of our explainable framework for remaining time prediction in Fine Management dataset. 
The fact that the fine is related to a parking ticket (\textit{notificationType=C}) is one of the largest factors that influences the prediction; the color blue of the bar and the information that the value is negative (-270 days) indicates that the influence is towards strongly decreasing the remaining time. 
This is principally due to the fact that parking tickets can be paid immediately after the ticket creation, in which case the paperwork is bypassed and unnecessary administration time and costs are avoided. 
A similar explanation can be given for the second largest factor (\textit{$Resource > 861.50$}), which is decreasing the remaining time prediction by -200 days. The resource here is the person responsible to notify the fine; a high identification code indicates that the resource belongs to a local police department, which is usually responsible to notify the fines related to parking tickets.
Other important attributes that are influencing the prediction are related to the performed activity. In particular,  
on the other side of the spectrum, explanation \emph{Activity=Appeal to Judge} has the largest positive shapley value (145 days). This can also be justified: when the offender appeals against the payment of the fine, the execution takes longer due to the involvement of the judge.


\begin{figure}[t!]
    \includegraphics[width=1 \columnwidth]{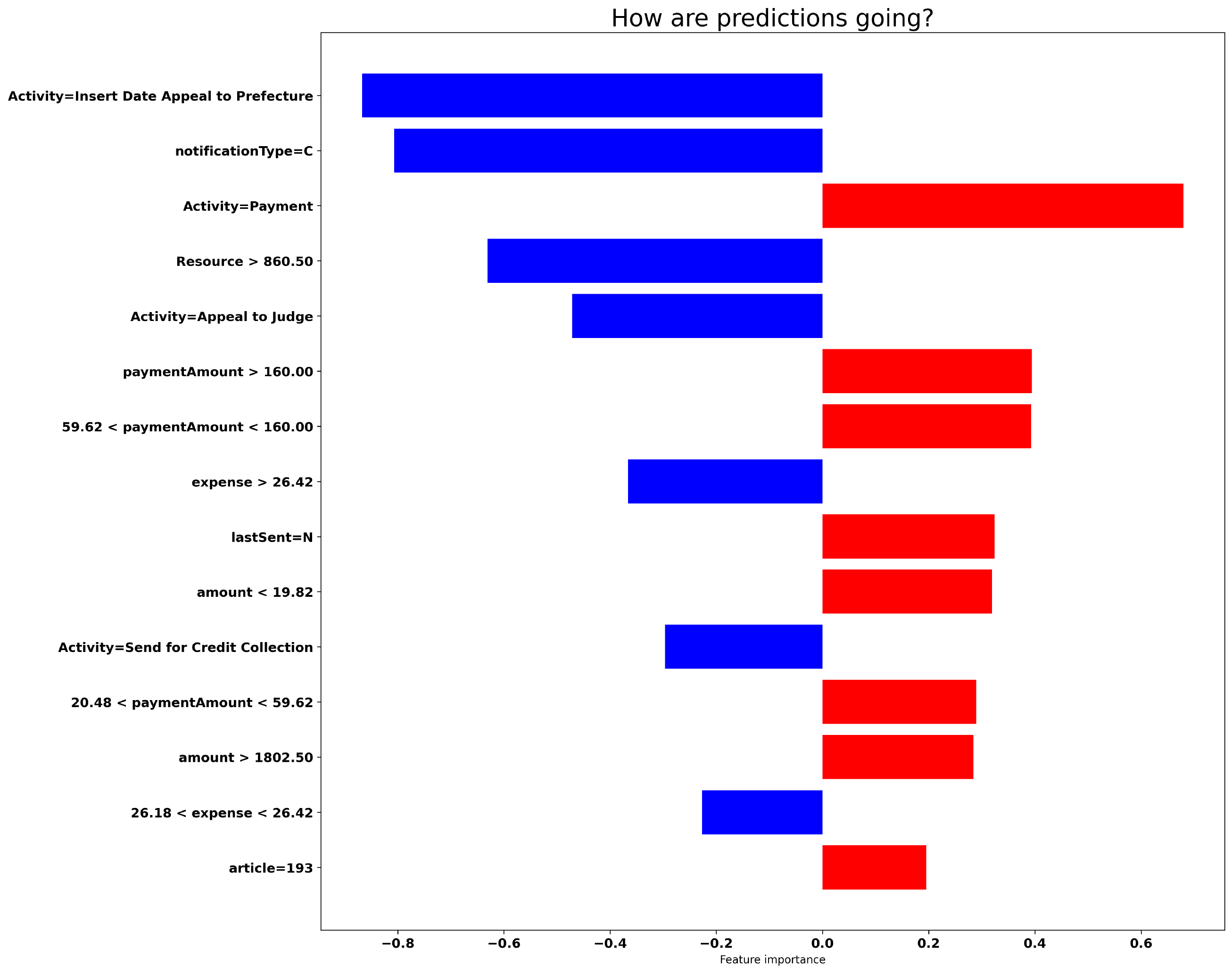}
    \caption{Offline explanations for \textit{Send for Credit Collection} prediction (Fine Management)}
    \label{fig:explanation_send_credit_prefecture}
\end{figure}

Finally, Figure \ref{fig:explanation_send_credit_prefecture} reports on the explanations for \textit{Send for Credit Collection} occurrence prediction, which is the activity performed when the case is sent to an external credit collection agency that will contact the offender to collect the payment.
Here, when an appeal against the fine is initiated within 60 days and it is correctly registered by the corresponding prefecture (\textit{Activity=Insert Date Appeal to Prefecture}) is the most influential factor and its influence is towards strongly decreasing the probability that a \textit{Send for Credit Collection} will be performed (as it can be seen by the blue bar and the associated probability of -0.85). A further analysis of the data confirmed this finding: when an appeal is sent to the prefecture, the external credit collection agency is alerted only in the 7\% of cases (282 over 4043 cases). 
A similar situation happens when, after receiving the results of the appeal, the offender appeals against the result (\textit{Activity=Appeal to Judge}): the influence is again towards decreasing the probability that the external credit collection agency will be alerted (-0.50). The data confirmed that when the judge is appealed, activity \textit{Send for Credit Collection} will be performed only in the 27\% of cases (148 over 541 cases).
Other attributes contributing to decrease the probability that a \textit{Send for Credit Collection} will be performed are \textit{notificationType=C} and \textit{$Resource > 860.50$}, which are decreasing the probability respectively by -0.80 and -0.65. As it was previously said, these factors indicate that the fine is related to a parking ticket and that the resource belongs to a local police department, which represents a simpler case in which tickets can be paid immediately after the ticket creation, thus not usually requiring the involvement of an external agency responsible for the credit collection.
Conversely, when the payment has been already performed (\textit{Activity=Payment}) the influence is towards increasing the probability that a \textit{Send for Credit Collection} will be performed (0.65); this was actually surprising, but a further analysis of the data confirmed that there is a small chance that after performing the payment, a \textit{Send for Credit Collection} action is initiated in the 3\% of cases (2126 out of 65800 cases). This means that there could be cases where an action is initiated by the external credit collector agency even if the payment has been already made, indicating a possible lack of synchronization between the accounting department and the external credit collection agency.

\pagebreak
\section{}
\label{appendix:response_distribution}

\begin{figure}[!htb]
    \centering
    \includegraphics[width=1 \columnwidth]{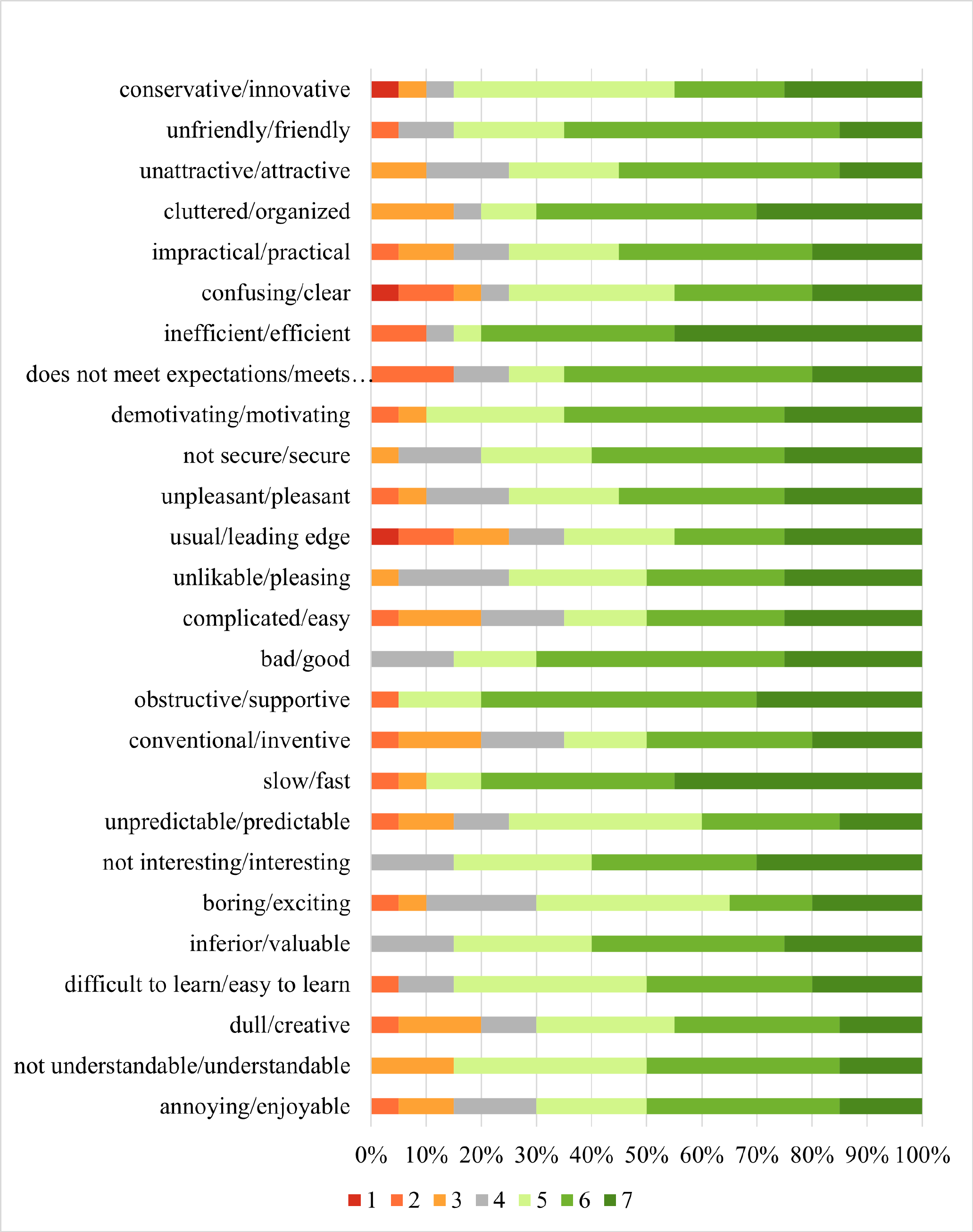}
    \caption{Distribution of the responses for each item of the User Experience Questionnaire (UEQ). 
Each row is an item and reports pairs of opposite adjectives, which are mapped to a response scale on seven points.
For each row, the width of the several boxplots varies depending on the percentage of users reporting the corresponding score.}
\end{figure}

\end{document}